\documentclass{book}
\usepackage{authblk}
\usepackage{roboto}

\usepackage[english]{babel}
\usepackage[letterpaper,top=2cm,bottom=2cm,left=3cm,right=3cm,marginparwidth=1.75cm]{geometry}

\usepackage{amsfonts}
\UseRawInputEncoding

\usepackage{pdflscape}
\usepackage{rotating}
\usepackage{longtable}
\usepackage{array}
\usepackage{float}
\usepackage{amsmath}
\usepackage{graphicx}
\usepackage{inconsolata}
\usepackage[colorlinks=true, allcolors=blue]{hyperref}
\usepackage{enumitem}
\usepackage{tikz}  
\usepackage{listings}
\usepackage{xcolor}
\usepackage[T1]{fontenc}
\usepackage{IEEEtrantools}  
\usepackage{epigraph}  

\usepackage{tikz}
\usetikzlibrary{positioning} 

\definecolor{bgcolor}{rgb}{0.97,0.97,0.97}
\definecolor{codeblue}{rgb}{0.1,0.1,0.8}
\definecolor{codegreen}{rgb}{0,0.4,0}
\definecolor{codegray}{rgb}{0.4,0.4,0.4}
\definecolor{codepurple}{rgb}{0.5,0,0.5}
\definecolor{codered}{rgb}{0.6,0.2,0.2}
\definecolor{lightgray}{rgb}{0.9,0.9,0.9}
\definecolor{darkgray}{rgb}{0.6,0.6,0.6}  

\makeatletter
\renewcommand{\paragraph}{
  \@startsection{paragraph}{4}{\z@}{1ex}{-1em}{\normalfont\normalsize\bfseries\color{gray}}}
\makeatother

\lstdefinestyle{python}{
    language=Python,
    basicstyle=\ttfamily\small\color{black}\usefont{T1}{zi4}{m}{n},  
    keywordstyle=\bfseries\color{codeblue},  
    stringstyle=\color{codegreen},  
    commentstyle=\slshape\color{codegray},  
    showstringspaces=false,
    numbers=left,
    numberstyle=\tiny\color{codegray},  
    stepnumber=1,
    numbersep=8pt,
    frame=single,
    rulecolor=\color{darkgray},  
    breaklines=true,
    backgroundcolor=\color{bgcolor},
    tabsize=4,
    captionpos=b,
    morekeywords={self},  
}

\lstdefinestyle{cmd}{
    language=bash,
    basicstyle=\ttfamily\small\color{black}\usefont{T1}{zi4}{m}{n},  
    keywordstyle=\bfseries\color{blue},
    stringstyle=\color{codegreen},
    commentstyle=\itshape\color{gray},
    showstringspaces=false,
    numbers=none,
    frame=single,
    rulecolor=\color{darkgray},  
    breaklines=true,
    backgroundcolor=\color{bgcolor},
    tabsize=4,
    captionpos=b,
}

\title{Deep Learning, Machine Learning, Advancing Big Data Analytics and Management}
\footnotesize
\author{
    Weiche Hsieh{*}\thanks{National Tsing Hua University, s112033645@m112.nthu.edu.tw}, 
    Ziqian Bi\textsuperscript{*$\dagger$}\thanks{Indiana University, bizi@iu.edu}, 
    Keyu Chen\thanks{Georgia Institute of Technology, kchen637@gatech.edu},
    Benji Peng\thanks{AppCubic, benji@appcubic.com}, 
    Sen Zhang\thanks{Rutgers University, sen.z@rutgers.edu}, 
    Jiawei Xu\thanks{Purdue University, xu1644@purdue.edu}, 
    Jinlang Wang\thanks{University of Wisconsin-Madison, jinlang.wang@wisc.edu}, 
    Caitlyn Heqi Yin\thanks{University of Wisconsin-Madison, hyin66@wisc.edu}, 
    Yichao Zhang\thanks{The University of Texas at Dallas, yichao.zhang.us@gmail.com},
    Pohsun Feng\thanks{National Taiwan Normal University, 41075018h@ntnu.edu.tw},  
    Yizhu Wen\thanks{University of Hawaii, yizhuw@hawaii.edu}, 
    Tianyang Wang\thanks{Xi'an Jiaotong-Liverpool University, Tianyang.Wang21@student.xjtlu.edu.cn},
    Ming Li\thanks{Georgia Institute of Technology, mli694@gatech.edu},  
    Chia Xin Liang\thanks{JTB Technology Corp., cxldun@gmail.com},
    Jintao Ren\thanks{Aarhus University, jintaoren@clin.au.dk}, 
    Qian Niu\thanks{Kyoto University, niu.qian.f44@kyoto-u.ac.jp}, 
    Silin Chen\thanks{Zhejiang University, A1033439225@gmail.com}, 
    Lawrence K.Q. Yan\thanks{Hong Kong University of Science and Technology, kqyan@connect.ust.hk}, 
    Han Xu\thanks{University of Illinois Urbana-Champaign, hanxu8@illinois.edu}, 
    Hong-Ming Tseng\thanks{School of Visual Arts, htseng@sva.edu}, 
    Xinyuan Song\thanks{Emory University, songxinyuan@pku.edu.cn}, 
    Bowen Jing\thanks{University of Manchester, bowen.jing@postgrad.manchester.ac.uk},
    Junjie Yang\textsuperscript\thanks{Pingtan Research Institute of Xiamen University, youngboy@xmu.edu.cn}, 
    Junhao Song\textsuperscript\thanks{Imperial College London, junhao.song23@imperial.ac.uk},
    Junyu Liu\thanks{Kyoto University, liu.junyu.82w@kyoto-u.ac.jp},
    Ming Liu{*$\dagger$}\thanks{Purdue University, liu3183@purdue.edu}
}

\date{}  

\begin{document}

\maketitle

\begingroup
\renewcommand\thefootnote{}\footnote{
    \textsuperscript{*} Equal contribution \\
    \textsuperscript{$\dagger$} Corresponding author
}
\addtocounter{footnote}{0}
\endgroup

\epigraph{"Information is the oil of the 21st century, and analytics is the combustion engine." }{\textit{Peter Sondergaard}}

\epigraph{"Big Data will spell the death of customer segmentation and force the marketer to understand each customer as an individual within 18 months or risk being left in the dust."}{\textit{Ginni Rometty}}

\epigraph{"The world is one big data problem."}{\textit{Andrew McAfee}}

\epigraph{"The most valuable commodity I know of is information." }{\textit{Gordon Gekko}}

\tableofcontents  

\setcounter{part}{2} 

\chapter{Introduction to Big Data Analytics}

\section{What is Big Data?}
    Big Data refers to large sets of data that are characterized by high volume, velocity, variety, value, and veracity, often referred to as the "5Vs". These characteristics make traditional data processing tools inadequate for handling such data efficiently \cite{mayer2013big}.
\begin{figure}[ht]
    \centering
    \includegraphics[width=1.0\textwidth]{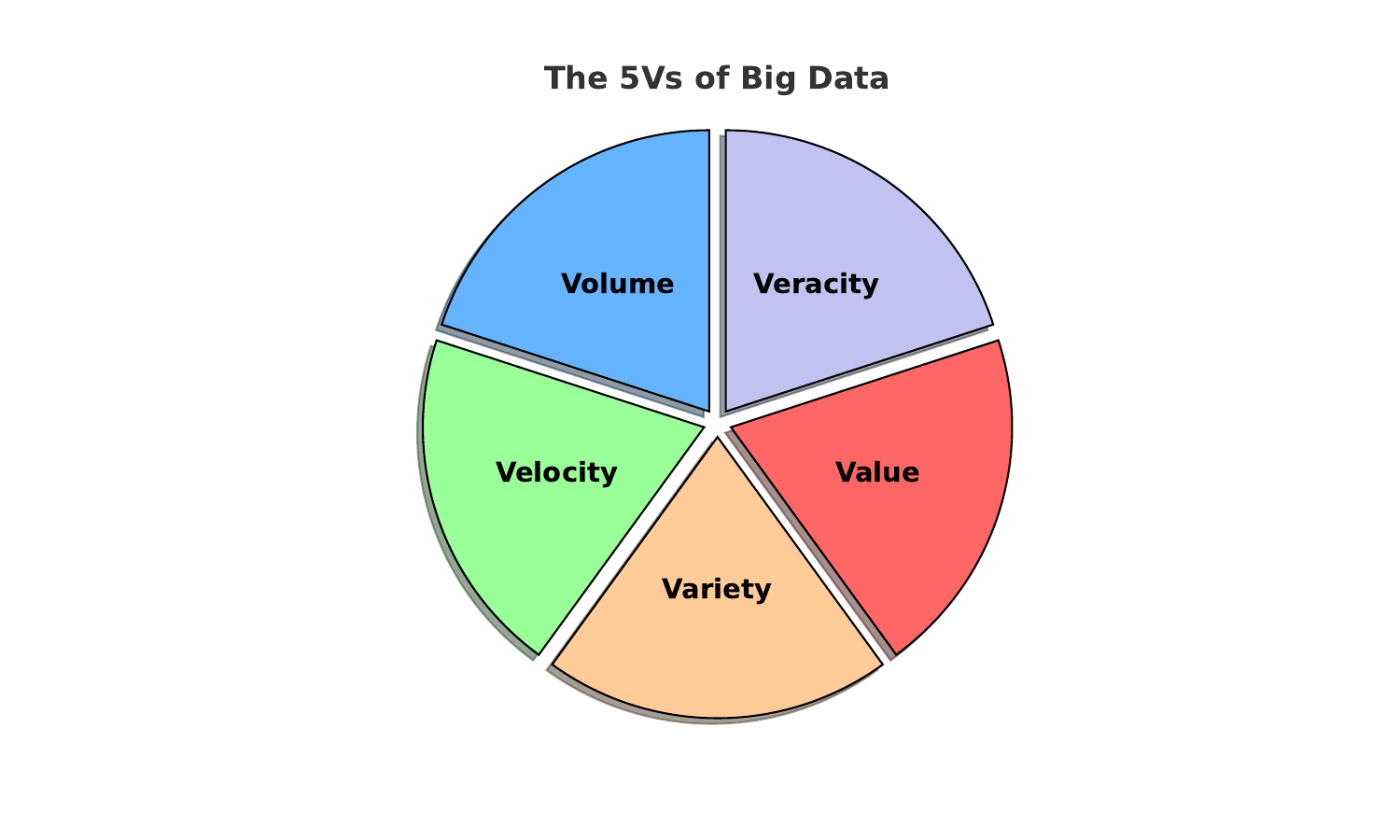}
    \caption{The 5Vs of Big Data}
    \label{fig:5vs}
\end{figure}
    For example, social media platforms generate huge amounts of user data daily, and e-commerce platforms process thousands of transactions every second. These are classic examples of big data \cite{provost2013data}.
\section{The Importance of Big Data}
    The importance of big data lies in its ability to help businesses and organizations make more informed decisions. By analyzing large datasets, companies can uncover market trends, consumer behavior patterns, and optimize their operations. Big data has applications across many sectors, including finance, healthcare, retail, and manufacturing \cite{sagiroglu2013big}.
    For instance, a retailer can analyze customer purchase history to predict future shopping needs, allowing for personalized promotions or recommendations \cite{kitchens2018advanced}.
\section{Big Data vs. Traditional Data}
    Compared to traditional data, big data requires more complex processing methods. Traditional data is often well-structured, relatively small in volume, and can be managed using simple database tools. Big data, on the other hand, necessitates advanced technologies and tools like distributed storage and parallel processing to handle its scale and complexity \cite{mayer2013big}.

    For example, traditional data might be a company's internal sales report, while big data includes massive amounts of unstructured data from sources like social media, sensors, and mobile devices.

\section{Big Data Use Cases and Applications}
    Big data is applied across various fields. Some typical examples include:
    \begin{itemize}
        \item In healthcare, big data helps doctors provide personalized treatments by analyzing patient records and genetic information \cite{dash2019big}.
        \item In finance, big data analysis is used to predict market trends and manage risks \cite{provost2013data}.
        \item In smart transportation systems, big data helps optimize traffic routes, reducing congestion and carbon emissions \cite{jan2019designing}.
    \end{itemize}

\section{Challenges in Big Data Analytics}
    Despite the opportunities, big data analytics faces many challenges, such as data storage and management, data privacy and security issues, and how to extract valuable insights from massive datasets. Additionally, processing big data requires efficient computing resources and sophisticated algorithms \cite{sagiroglu2013big, price2019privacy}.

    For example, big data storage demands distributed databases, while privacy concerns require adherence to strict regulations like GDPR during data collection and usage \cite{van2019does}.

\chapter{The Data Analytics Process}

\section{Survey and Questionnaire-Based Data Collection}
    Surveys and questionnaires remain one of the most structured methods for collecting data, particularly when researchers seek to gather specific information directly from individuals \cite{rea2014designing}. This method is often employed in market research, customer feedback systems, and employee satisfaction studies. The primary advantage of surveys is that they allow the researcher to design specific questions targeted at a particular population, ensuring the collection of relevant data \cite{groves2011survey}.
        
    \textbf{Example}: An online retailer might use a customer satisfaction survey to gather data on shopping experiences. The survey may ask users to rate their satisfaction with product variety, website navigation, and customer support. Such data is then used to improve the user experience and make data-driven decisions about future offerings.

\section{Sensors and IoT Devices}
    In the age of the Internet of Things (IoT), devices and sensors have become a prevalent source of real-time data collection. These devices continuously monitor and report data, providing granular insights into the environment, operations, or performance. Sensors are widely used in industries such as manufacturing, agriculture, healthcare, and smart cities \cite{bahga2014internet}.
        
    \textbf{Example}: In agriculture, smart farming relies on IoT sensors to collect data on soil moisture, temperature, and nutrient levels. This data helps farmers make informed decisions about irrigation, fertilization, and harvesting, ultimately increasing yield and reducing waste. The collected data is often voluminous and highly varied, but through big data analytics, patterns can be identified that enhance productivity \cite{xu2022review}.

\section{Web Scraping}
    Web scraping refers to the automated process of extracting data from websites. It is particularly useful when dealing with unstructured data from sources like social media, news portals, or online product listings. The data collected through web scraping can provide valuable insights into customer sentiment, competitor analysis, and market trends \cite{zhao2022web}.
        
    \textbf{Example}: A company may scrape user reviews from e-commerce websites to analyze sentiment regarding a particular product. By using web scraping, the company can collect thousands of reviews and then apply sentiment analysis techniques to assess the overall public perception of the product. This method enables the collection of large volumes of data in real time, providing businesses with a competitive edge.
        
\subsection{Transaction Data}
    Transaction data is generated whenever a transaction is made, whether it is financial, retail, or service-based. This data is typically structured and includes information about the buyer, seller, transaction amount, time, and place. Transaction data is highly valuable for industries like banking, retail, and e-commerce, where customer behavior, purchasing trends, and financial performance need to be analyzed \cite{harris1986transaction}.
        
    \textbf{Example}: A supermarket chain collects transaction data every time a customer makes a purchase. This data includes the items bought, the quantities, prices, and the method of payment. By aggregating this data, the supermarket can analyze purchasing patterns, determine which products are most popular at specific times, and make decisions on inventory management and promotions \cite{ruben2007vegetables}.
        
\subsection{Social Media and Online Interaction Data}
    Social media platforms generate vast amounts of data from user interactions, including posts, comments, likes, shares, and follows. This unstructured data offers insights into consumer behavior, opinions, and trends. Social media data collection is essential for brands looking to track their online presence and engage with customers \cite{batrinca2015social}.
        
    \textbf{Example}: A fashion brand collects data from platforms such as Instagram and Twitter to track how its latest clothing line is being received by the public. By analyzing hashtags, mentions, and user-generated content, the brand can gauge consumer interest, respond to feedback, and tailor future marketing campaigns \cite{bossetta2018digital}.
        
\subsection{Logs and Machine-Generated Data}
    Logs and machine-generated data are commonly produced by servers, applications, and various systems. These logs capture detailed information about system activities, including user actions, errors, and system performance. Analyzing log data is essential for ensuring system security, monitoring performance, and optimizing operations \cite{schmidt2012logging}.
        
    \textbf{Example}: A cybersecurity company may collect and analyze logs from firewalls and security systems to detect potential threats or breaches. These logs can reveal patterns of unusual activity, such as repeated failed login attempts, which may indicate a brute force attack. Using big data analytics, the company can identify vulnerabilities and take corrective actions \cite{landauer2020system}. 
        
\subsection{Public Data and Open Data}
    Governments, organizations, and institutions often release open data for public use. This data is typically structured and covers areas such as demographics, health, environment, and transportation. Publicly available data serves as a valuable resource for researchers and businesses alike \cite{gurin2014open}.
        
    \textbf{Example}: A health organization may collect public health data from government databases to track the spread of a disease. By integrating this data with proprietary health records, the organization can create predictive models to anticipate outbreaks and allocate resources accordingly \cite{chen2014review}. 
        
    \bigskip
        
    In summary, effective data collection is the foundation of successful big data analytics. The choice of method depends on the type of data needed, its source, and the context of its application. Whether through sensors, surveys, or social media, collecting relevant, accurate, and sufficient data enables organizations to draw meaningful insights and make informed decisions.

\chapter{Data Warehouse}

\section{Introduction to Data Warehousing}

\subsection{Definition and Importance of Data Warehousing}
A data warehouse is a centralized repository that stores large volumes of data from various sources. It is structured in a way that facilitates analysis and reporting, enabling organizations to derive valuable insights. Unlike traditional databases that focus on current transactions, a data warehouse is optimized for querying and analyzing historical data \cite{inmon2005building}. 

\textbf{Importance of Data Warehousing:}
\begin{itemize}
    \item \textbf{Historical Data Analysis:} A data warehouse stores large volumes of historical data that can be used to track and analyze trends over time. For instance, a retail company may use the data warehouse to analyze customer buying patterns across different time periods.
    \item \textbf{Decision Support:} By consolidating data from multiple sources, data warehouses enable better decision-making. Executives can access reports and dashboards to inform business strategy. 
    \item \textbf{Data Consistency:} Data warehouses ensure data is transformed into a consistent format, even if it comes from different sources. This helps in accurate analysis and reporting.
    \item \textbf{Performance Optimization:} Data warehouses are optimized for complex queries and reporting, improving the performance of analytics tasks compared to traditional databases.
\end{itemize}

\subsection{Evolution of Data Warehousing}
The concept of data warehousing has evolved significantly over time to address the increasing volume, variety, and velocity of data \cite{chandra2018comprehensive}.
\begin{itemize}
    \item \textbf{Early Stage:} Initially, data warehousing was designed to handle structured data from limited sources such as ERP and CRM systems.
    \item \textbf{Modern Data Warehousing:} With the rise of big data, data warehouses have evolved to integrate unstructured and semi-structured data, such as social media feeds, sensor data, and logs.
    \item \textbf{Cloud Data Warehousing:} The move to cloud-based architectures further revolutionized data warehouses, enabling organizations to scale their storage and compute power elastically.
\end{itemize}

\subsection{Data Warehousing in the Big Data Ecosystem}
In the context of big data, data warehouses are part of a broader ecosystem that includes data lakes, real-time data processing, and machine learning pipelines. While data warehouses are traditionally used for structured data and reporting, they now coexist with data lakes, which store raw and unstructured data.

\begin{lstlisting}[style=python]
# Python example to show the connection between a data warehouse and big data processing
import pandas as pd
import pyodbc

# Connecting to a data warehouse using pyodbc
conn = pyodbc.connect('DRIVER={SQL Server};'
                      'SERVER=server_name;'
                      'DATABASE=data_warehouse_db;'
                      'UID=user;PWD=password')

# Query to retrieve data for analysis
query = "SELECT * FROM sales_data WHERE year = 2023"
sales_data = pd.read_sql(query, conn)

# Performing basic data analysis
total_sales = sales_data['amount'].sum()
print(f'Total Sales in 2023: {total_sales}')
\end{lstlisting}

\section{Data Warehouse Architecture}

\subsection{Basic Components of a Data Warehouse}
A data warehouse typically consists of several key components \cite{adamson1998data}:
\begin{itemize}
    \item \textbf{Data Sources:} These are external systems like databases, flat files, and web services that provide the raw data.
    \item \textbf{ETL (Extract, Transform, Load) Process:} The ETL layer extracts data from various sources, transforms it into a usable format, and loads it into the warehouse.
    \item \textbf{Storage Layer:} This is where transformed data is stored for long-term analysis. It may include fact and dimension tables, which are organized to support multi-dimensional analysis.
    \item \textbf{Access Layer:} Users access data via query tools, reporting tools, or dashboards.
\end{itemize}

\begin{tikzpicture}[sibling distance=5cm,
  every node/.style = {shape=rectangle, rounded corners,
    draw, align=center,
    top color=white, bottom color=blue!20}]]
  \node {Data Sources}
    child {node {ETL Process} 
        child {node {Extract}}
        child {node {Transform}}
        child {node {Load}}}
    child {node {Storage Layer} 
        child {node {Fact Tables}}
        child {node {Dimension Tables}}}
    child {node {Access Layer}};
\end{tikzpicture}

\subsection{Three-Tier Architecture: ETL, Storage, and Access Layers}
The data warehouse architecture is typically divided into three layers:
\begin{itemize}
    \item \textbf{ETL Layer:} Responsible for extracting, transforming, and loading data from various sources into the warehouse.
    \item \textbf{Storage Layer:} Stores the data in a structured way for efficient querying.
    \item \textbf{Access Layer:} Provides interfaces for users to interact with the data, such as SQL queries, BI tools, and dashboards.
\end{itemize}

\subsection{Data Warehouse vs Data Lake}
Data warehouses and data lakes serve different purposes within the data ecosystem. 
\begin{itemize}
    \item \textbf{Data Warehouse:} Designed for structured, processed data that is used for reporting and analysis. The data is often cleaned and aggregated before being loaded into the warehouse.
    \item \textbf{Data Lake:} A data lake stores raw, unprocessed data, including structured, semi-structured, and unstructured formats. It is often used for exploratory data analysis, machine learning, and big data processing.
\end{itemize}

\textbf{Example:}
A financial institution may use a data warehouse to generate regular financial reports, while the data lake is used for storing logs and raw customer transaction data that can be analyzed later using machine learning techniques \cite{karkovskova2023data}.

\subsection{Cloud Data Warehousing}
Cloud data warehousing offers scalable storage and compute resources, allowing businesses to handle massive amounts of data without worrying about infrastructure limitations \cite{kavis2014architecting}.
\begin{itemize}
    \item \textbf{Scalability:} Cloud data warehouses can automatically scale to meet the growing demand for data storage and processing.
    \item \textbf{Cost Efficiency:} By adopting a pay-as-you-go model, cloud data warehouses eliminate the need for heavy upfront investments.
    \item \textbf{Integration with Big Data:} Modern cloud warehouses integrate easily with big data platforms, enabling seamless data movement between systems.
\end{itemize}

\begin{lstlisting}[style=python]
# Example of working with cloud-based data warehouse (e.g., AWS Redshift)
import psycopg2

# Connect to AWS Redshift
conn = psycopg2.connect(
    dbname='datawarehouse',
    host='myredshiftcluster.amazonaws.com',
    port='5439',
    user='awsuser',
    password='mypassword'
)

# Query data from the cloud data warehouse
cur = conn.cursor()
cur.execute("SELECT * FROM customer_data LIMIT 10;")
rows = cur.fetchall()
for row in rows:
    print(row)
\end{lstlisting}

\section{Data Warehouse Models}
A data warehouse is a system that aggregates data from various sources into a central repository. It is structured to support querying and analysis, rather than transaction processing. Several models are used to organize data in a warehouse, such as the star schema and snowflake schema \cite{imhoff2003mastering}.

    \subsection{Star Schema}
    The star schema is a simple data warehouse design where data is organized into facts and dimensions. A fact table sits at the center of the schema, surrounded by dimension tables, forming a star-like shape. This is the most common schema in data warehousing and is optimized for query performance \cite{kimball2013data}.

    \begin{itemize}
        \item \textbf{Fact Table}: Contains the quantitative data (facts) like sales, revenue, or transactions. Each record corresponds to a specific event or occurrence.
        \item \textbf{Dimension Tables}: Contain descriptive attributes that provide context for the facts. These could include dimensions like time, location, and product.
    \end{itemize}

    \paragraph{Example of a Star Schema:} Suppose we are analyzing sales data. Our fact table could store total sales, with dimension tables for the product, time of sale, and sales region.

    \begin{tikzpicture}
        \node (fact) [rectangle, draw] {Fact Table: Sales};
        \node (time) [rectangle, draw, above left=of fact] {Dimension: Time};
        \node (product) [rectangle, draw, above right=of fact] {Dimension: Product};
        \node (location) [rectangle, draw, below left=of fact] {Dimension: Location};
        \node (customer) [rectangle, draw, below right=of fact] {Dimension: Customer};
    
        \draw[->] (time) -- (fact);
        \draw[->] (product) -- (fact);
        \draw[->] (location) -- (fact);
        \draw[->] (customer) -- (fact);
    \end{tikzpicture}
    
    \subsection{Snowflake Schema}
    The snowflake schema is a more complex version of the star schema. It normalizes the dimension tables by breaking them into additional tables. This can save storage space, but may reduce query performance because more joins are needed \cite{kimball2013data}.

    \paragraph{Example of a Snowflake Schema:} If our dimension table for products is large, we could split it into two tables: one for product categories and another for specific products.

\begin{tikzpicture}
    \node (fact) [rectangle, draw] {Fact Table: Sales};
    \node (time) [rectangle, draw, above left=of fact] {Dimension: Time};
    \node (product) [rectangle, draw, above=of fact] {Dimension: Product};
    \node (category) [rectangle, draw, above=of product] {Dimension: Category};
    \node (location) [rectangle, draw, below left=of fact] {Dimension: Location};

    \draw[->] (time) -- (fact);
    \draw[->] (product) -- (fact);
    \draw[->] (category) -- (product);
    \draw[->] (location) -- (fact);
\end{tikzpicture}

    \subsection{Fact Tables and Dimension Tables}
    \begin{itemize}
        \item \textbf{Fact Tables}: These store measurable business data, typically including numeric values such as revenue, sales count, or profit margins. Each fact table row is linked to associated dimension tables by foreign keys \cite{rowen2001analysis}.
        \item \textbf{Dimension Tables}: These provide descriptive context for the facts, like time periods, customer details, or geographical locations. They help users drill down into the data by various perspectives \cite{rowen2001analysis}.
    \end{itemize}

    \subsection{OLAP and OLTP}
    OLAP (Online Analytical Processing) systems are optimized for complex queries and data analysis, whereas OLTP (Online Transaction Processing) systems handle transactional data for day-to-day operations \cite{plattner2009common}.

    \begin{itemize}
        \item \textbf{OLTP Example:} An e-commerce website where users complete purchases. Here, OLTP processes these transactions in real-time.
        \item \textbf{OLAP Example:} A data warehouse where historical sales data from the website is stored and analyzed to understand purchasing trends.
    \end{itemize}

\section{ETL Process (Extract, Transform, Load)}
The ETL process is a crucial step in building a data warehouse. It involves extracting data from various sources, transforming it into a suitable format, and loading it into the data warehouse. Each step presents its own challenges and requires careful consideration to ensure data integrity \cite{caserta2013data}.

    \subsection{Overview of the ETL Process}
    ETL is typically broken down into three phases:
    \begin{itemize}
        \item \textbf{Extract:} Gather data from various sources such as databases, APIs, or flat files.
        \item \textbf{Transform:} Cleanse and convert the data into a format suitable for analysis. This may involve filtering, sorting, aggregating, and applying business logic.
        \item \textbf{Load:} Insert the transformed data into the target data warehouse or analytical system, either in bulk (batch processing) or in real-time (streaming).
    \end{itemize}

    \subsection{Data Extraction: Sources and Challenges}
    Data extraction is the first step in the ETL process, and it involves retrieving data from multiple, disparate sources. Common challenges during this phase include \cite{sabtu2017challenges}:
    \begin{itemize}
        \item \textbf{Data Heterogeneity:} Different data formats, such as SQL databases, NoSQL databases, or flat files.
        \item \textbf{Incomplete Data:} Missing values or inconsistent data entries.
        \item \textbf{Data Volume:} Managing large volumes of data can be a significant challenge in big data environments.
    \end{itemize}

\begin{lstlisting}[style=python]
import pandas as pd

# Example of reading data from a CSV file in Python using Pandas
df = pd.read_csv('sales_data.csv')

# Display the first few rows of the extracted data
print(df.head())
\end{lstlisting}

    \subsection{Data Transformation: Cleaning and Integration}
    Once the data is extracted, it must be transformed into a format suitable for analysis. This step involves data cleaning, deduplication, normalization, and aggregation.

    \begin{itemize}
        \item \textbf{Data Cleaning:} Correcting or removing inaccurate records, handling missing values, and ensuring data consistency.
        \item \textbf{Data Integration:} Combining data from different sources to create a unified view, such as merging sales and customer information.
    \end{itemize}

\begin{lstlisting}[style=python]
# Example of cleaning data: removing rows with missing values
df_cleaned = df.dropna()

# Example of transformation: converting a string date column to datetime format
df_cleaned['date'] = pd.to_datetime(df_cleaned['date'])

# Example of merging two datasets (sales and customer data)
customer_data = pd.read_csv('customer_data.csv')
merged_df = pd.merge(df_cleaned, customer_data, on='customer_id')

print(merged_df.head())
\end{lstlisting}

    \subsection{Data Loading: Batch Processing and Real-Time Loading}
    After the data is transformed, it must be loaded into the data warehouse. There are two primary methods for loading data:
    \begin{itemize}
        \item \textbf{Batch Processing:} Involves loading large volumes of data at specific intervals (e.g., daily or weekly). This is suitable for data that does not need to be immediately available.
        \item \textbf{Real-Time Processing:} Involves streaming data into the warehouse as it is generated. This is useful for time-sensitive applications such as stock trading or real-time analytics.
    \end{itemize}

\begin{lstlisting}[style=python]
# Example of loading data into a database using SQLAlchemy in Python
from sqlalchemy import create_engine

# Create a connection to the database
engine = create_engine('sqlite:///sales_data.db')

# Load the transformed data into a database
merged_df.to_sql('sales', con=engine, if_exists='replace')

# Verify data was loaded
with engine.connect() as connection:
    result = connection.execute("SELECT * FROM sales LIMIT 5;")
    for row in result:
        print(row)
\end{lstlisting}

\section{Data Warehousing and Big Data}

\subsection{Integration with Hadoop and Spark}
In modern data ecosystems, data warehouses often need to integrate with big data platforms such as Hadoop and Spark. Hadoop is primarily used for storing and processing large volumes of unstructured and semi-structured data, while Spark is a fast, in-memory data processing engine that is used for real-time analytics, machine learning, and big data processing \cite{white2012hadoop, ellis2014real}. 

\textbf{Integration with Hadoop:}
\begin{itemize}
    \item \textbf{Data Storage:} Hadoop uses a distributed file system, HDFS (Hadoop Distributed File System), to store large datasets. Data can be ingested from Hadoop into the data warehouse for reporting and analysis.
    \item \textbf{ETL Pipelines:} Data can be extracted from Hadoop, transformed in the ETL layer, and then loaded into a data warehouse for further use. This is especially useful when structured data from Hadoop needs to be analyzed along with transactional data.
    \item \textbf{Hive:} Hive is a data warehousing tool built on top of Hadoop, allowing users to run SQL-like queries over data stored in HDFS. It bridges the gap between Hadoop and traditional data warehouses by providing a familiar SQL interface.
\end{itemize}

\begin{lstlisting}[style=python]
# Python integration with Hadoop via pywebhdfs
from pywebhdfs.webhdfs import PyWebHdfsClient

# Connect to HDFS
hdfs = PyWebHdfsClient(host='hadoop_host', port='50070', user_name='hadoop_user')

# Reading a file from HDFS
file_content = hdfs.read_file('/data/large_dataset.csv')

# Processing the data and loading it into a data warehouse
# Assuming the file content is CSV, we load it into a DataFrame for analysis
import pandas as pd
from io import StringIO

df = pd.read_csv(StringIO(file_content.decode('utf-8')))
print(df.head())
\end{lstlisting}

\textbf{Integration with Spark:}
\begin{itemize}
    \item \textbf{In-memory Processing:} Spark can be used to process large datasets stored in a data warehouse, using its fast in-memory computation capabilities to perform real-time analytics.
    \item \textbf{Machine Learning:} PySpark (the Python API for Spark) can be integrated with data warehouses to perform advanced machine learning tasks on large datasets.
    \item \textbf{Data Pipeline Integration:} Data can be extracted from the warehouse, processed in Spark, and then written back for further analysis or reporting.
\end{itemize}

\begin{lstlisting}[style=python]
# Example of using PySpark to process data from a data warehouse
from pyspark.sql import SparkSession

# Create a Spark session
spark = SparkSession.builder.appName('DataWarehouseIntegration').getOrCreate()

# Load data from a data warehouse (for example, via JDBC)
df = spark.read \
    .format("jdbc") \
    .option("url", "jdbc:mysql://localhost:3306/data_warehouse") \
    .option("dbtable", "sales_data") \
    .option("user", "user") \
    .option("password", "password") \
    .load()

# Perform data processing
df_filtered = df.filter(df['year'] == 2023)
df_filtered.show()
\end{lstlisting}

\subsection{Real-Time Analytics in Data Warehousing}
Real-time analytics involves analyzing data as it is generated, without significant delays. This is increasingly important in scenarios like fraud detection, stock market analysis, and IoT (Internet of Things) applications, where decisions need to be made in near real-time \cite{ellis2014real}.

\textbf{How Real-Time Analytics Works:}
\begin{itemize}
    \item \textbf{Streaming Data:} Data is ingested in real-time from various sources, such as sensors, web traffic, or transactional systems.
    \item \textbf{Processing Pipelines:} Tools like Apache Kafka and Apache Flink are often used to build real-time data pipelines, which process streaming data before loading it into a real-time data warehouse.
    \item \textbf{Real-Time Queries:} Users can run real-time queries against the data warehouse to get up-to-the-minute insights.
\end{itemize}

\begin{lstlisting}[style=python]
# Example of real-time data ingestion and analysis using Python and Kafka
from kafka import KafkaConsumer

# Set up a Kafka consumer to listen for real-time data
consumer = KafkaConsumer('real_time_data',
                         bootstrap_servers=['localhost:9092'],
                         auto_offset_reset='earliest',
                         enable_auto_commit=True)

# Process each message in real-time
for message in consumer:
    print(f"Received real-time data: {message.value.decode('utf-8')}")
\end{lstlisting}

\subsection{Data Warehousing in Modern Big Data Architectures}
In modern big data architectures, data warehouses play an integral role in providing structured data for analytics, reporting, and business intelligence. A typical architecture includes:
\begin{itemize}
    \item \textbf{Data Lake:} Stores raw, unprocessed data in various formats (structured, semi-structured, and unstructured). This is the starting point for data ingestion in many big data systems.
    \item \textbf{Data Warehouse:} After processing and cleaning, data from the data lake is loaded into the data warehouse for structured analysis.
    \item \textbf{Analytics Layer:} Data warehouses feed structured data into various analytics and BI tools, which provide insights for decision-making.
\end{itemize}

\section{Performance and Optimization Techniques}

\subsection{Indexes and Partitioning}
Indexes and partitioning are critical techniques to optimize the performance of data warehouses. They help reduce query times by efficiently organizing and retrieving data \cite{winand2012sql}.

\textbf{Indexes:}
\begin{itemize}
    \item \textbf{What is an Index?} An index is a data structure that improves the speed of data retrieval operations on a database table.
    \item \textbf{Example:} If a data warehouse table contains millions of records, an index on the primary key can significantly reduce the time required to retrieve specific records.
\end{itemize}

\begin{lstlisting}[style=cmd]
-- Example SQL command to create an index
CREATE INDEX idx_sales_year ON sales_data (year);
\end{lstlisting}

\textbf{Partitioning:}
\begin{itemize}
    \item \textbf{What is Partitioning?} Partitioning divides a large table into smaller, more manageable pieces without physically splitting the table. Each partition is treated separately during queries, which speeds up data access.
    \item \textbf{Types of Partitioning:} The most common types of partitioning are range partitioning (e.g., dividing data by year) and hash partitioning (e.g., dividing data based on a hash function).
\end{itemize}

\begin{lstlisting}[style=cmd]
-- Example SQL command to partition a table by year
CREATE TABLE sales_data_partitioned
(
    id INT,
    product VARCHAR(100),
    amount DECIMAL,
    year INT
)
PARTITION BY RANGE (year)
(
    PARTITION p0 VALUES LESS THAN (2020),
    PARTITION p1 VALUES LESS THAN (2021),
    PARTITION p2 VALUES LESS THAN (2022)
);
\end{lstlisting}

\subsection{Query Optimization Techniques}
Query optimization is essential to ensure efficient data retrieval from a data warehouse. Some common techniques include \cite{shasha2004database}:
\begin{itemize}
    \item \textbf{Using Indexes:} As mentioned earlier, indexes help speed up data retrieval.
    \item \textbf{Avoiding Full Table Scans:} Full table scans can be expensive in terms of time and resources. Using filters and indexed columns in the WHERE clause can help avoid this.
    \item \textbf{Join Optimization:} Optimizing JOIN operations, such as by using indexed columns in the join conditions, can improve query performance.
\end{itemize}

\begin{lstlisting}[style=cmd]
-- Example SQL query optimized using indexed columns and avoiding full table scans
SELECT product, SUM(amount) 
FROM sales_data 
WHERE year = 2023 
GROUP BY product;
\end{lstlisting}

\subsection{Aggregation and Summarization Techniques}
Aggregating data is one of the key operations performed in data warehouses. It involves computing summary statistics, such as sums, averages, and counts, to support business decision-making \cite{shasha2004database}.

\textbf{Examples of Aggregation:}
\begin{itemize}
    \item \textbf{Group By:} Aggregating data by categories, such as calculating total sales per product or per region.
    \item \textbf{Rollup and Cube:} These SQL extensions allow multi-dimensional aggregation, providing summaries at different levels of detail.
\end{itemize}

\begin{lstlisting}[style=cmd]
-- Example SQL query using GROUP BY to summarize data
SELECT product, SUM(amount) AS total_sales
FROM sales_data
GROUP BY product;

-- Example SQL query using ROLLUP to create subtotals
SELECT product, region, SUM(amount) AS total_sales
FROM sales_data
GROUP BY ROLLUP (product, region);
\end{lstlisting}

\section{Data Governance and Data Warehouse Security}
As data becomes a critical asset for organizations, ensuring its quality, privacy, and security is essential. Data governance encompasses the management of data quality, privacy, and compliance, while security practices ensure that the data warehouse remains protected from unauthorized access and breaches.

    \subsection{Data Quality Management}
    Data quality management involves processes and technologies that ensure the data in a warehouse is accurate, complete, consistent, and up-to-date. High-quality data is critical for making reliable business decisions.

    \paragraph{Key Aspects of Data Quality Management:}
    \begin{itemize}
        \item \textbf{Accuracy:} Ensuring that the data accurately represents the real-world entities or events it is meant to describe. For example, customer contact information should be correct and up-to-date.
        \item \textbf{Completeness:} Data should not have missing values where it is essential for analysis. For instance, all transactions should have associated timestamps and customer IDs.
        \item \textbf{Consistency:} Ensuring that data is consistent across various systems and data sources. For example, the same customer should not have conflicting records across different databases.
        \item \textbf{Timeliness:} Data should be available when needed for analysis or reporting. Delays in data availability can affect decision-making processes.
    \end{itemize}

\begin{lstlisting}[style=python]
# Example: Identifying and handling missing values in a dataset
import pandas as pd

# Load sample data
df = pd.read_csv('data.csv')

# Identify missing values
missing_data = df.isnull().sum()

# Print columns with missing values
print(missing_data)

# Fill missing values with default values or drop rows with missing data
df_filled = df.fillna('Unknown')  # Example of filling missing values
df_cleaned = df.dropna()  # Example of dropping rows with missing values

print(df_filled.head())
print(df_cleaned.head())
\end{lstlisting}

    \subsection{Data Privacy and Compliance (e.g., GDPR)}
    Ensuring the privacy of personal data and adhering to regulations like GDPR (General Data Protection Regulation) is essential for any organization that handles sensitive information. Failure to comply with privacy laws can result in heavy fines and damage to a company's reputation \cite{van2019does}.

    \paragraph{GDPR Overview:}
    GDPR is a regulation in the European Union that governs how personal data is collected, processed, and stored. Key aspects include \cite{van2019does}:
    \begin{itemize}
        \item \textbf{Consent:} Organizations must obtain explicit consent from individuals to collect and process their personal data.
        \item \textbf{Right to Access:} Individuals have the right to know what personal data is being stored about them.
        \item \textbf{Right to Erasure:} Also known as the "right to be forgotten," individuals can request that their personal data be deleted.
        \item \textbf{Data Protection by Design:} Systems must be designed with privacy in mind, ensuring that data is handled securely.
    \end{itemize}

    \paragraph{Data Privacy Example:}
    Suppose we are managing a database that stores customer information. To comply with GDPR, we need to ensure that data is encrypted, customers can request their data, and personal information is anonymized where possible.

\begin{lstlisting}[style=python]
# Example: Anonymizing customer data by removing personally identifiable information (PII)
df['customer_name'] = 'Anonymous'
df['customer_email'] = 'redacted'

print(df.head())
\end{lstlisting}

    \subsection{Best Practices for Data Warehouse Security}
    Securing a data warehouse involves implementing multiple layers of protection to ensure that data is not compromised. This includes access controls, encryption, auditing, and monitoring.

    \paragraph{Security Best Practices:}
    \begin{itemize}
        \item \textbf{Access Control:} Limit access to the data warehouse based on user roles. For example, analysts should have read-only access, while database administrators have full control.
        \item \textbf{Encryption:} Both data at rest (stored in the warehouse) and data in transit (moving between systems) should be encrypted to protect against unauthorized access.
        \item \textbf{Auditing:} Maintain audit logs to track who accessed or modified data. This helps in detecting unauthorized access and ensuring compliance.
        \item \textbf{Data Masking:} Mask sensitive information so that even if unauthorized access occurs, critical data like credit card numbers or social security numbers remain protected.
        \item \textbf{Regular Security Audits:} Periodically conduct security audits and vulnerability assessments to identify potential risks.
    \end{itemize}

\begin{lstlisting}[style=python]
# Example: Implementing basic role-based access control (RBAC)
class User:
    def __init__(self, username, role):
        self.username = username
        self.role = role

# Define roles and access levels
roles_permissions = {
    'admin': ['read', 'write', 'delete'],
    'analyst': ['read'],
    'guest': ['read']
}

def has_permission(user, action):
    if action in roles_permissions.get(user.role, []):
        return True
    else:
        return False

# Example usage
user = User('john_doe', 'analyst')
print(has_permission(user, 'write'))  # Output: False
print(has_permission(user, 'read'))   # Output: True
\end{lstlisting}

\section{Future Trends in Data Warehousing}
As technology evolves, so do data warehousing practices. The shift to cloud computing, the integration of artificial intelligence (AI), and the rise of edge computing are reshaping how organizations manage and analyze their data \cite{chandra2018comprehensive}.

    \subsection{Data Warehousing in the Cloud}
    Cloud-based data warehouses, such as Amazon Redshift, Google BigQuery, and Snowflake, are gaining popularity due to their scalability, flexibility, and cost-effectiveness. These services allow organizations to store massive amounts of data without the need for expensive on-premise infrastructure.

    \paragraph{Benefits of Cloud Data Warehousing:}
    \begin{itemize}
        \item \textbf{Scalability:} Cloud data warehouses can scale up or down based on demand, allowing organizations to handle varying workloads.
        \item \textbf{Cost Efficiency:} Pay-as-you-go pricing models enable organizations to pay only for the storage and compute resources they use.
        \item \textbf{Ease of Use:} Cloud platforms offer built-in tools for data ingestion, querying, and machine learning integration.
        \item \textbf{High Availability:} Cloud services often provide redundancy and disaster recovery options, ensuring data is always accessible.
    \end{itemize}

\begin{lstlisting}[style=python]
# Example: Loading data into a cloud data warehouse using Python
import sqlalchemy

# Create an engine for a cloud data warehouse (example using Amazon Redshift)
engine = sqlalchemy.create_engine('redshift+psycopg2://user:password@host:port/dbname')

# Load a Pandas DataFrame into the cloud data warehouse
df.to_sql('table_name', engine, if_exists='replace')

print("Data successfully loaded into the cloud warehouse!")
\end{lstlisting}

    \subsection{Applications of AI and Machine Learning in Data Warehouse Optimization}
    AI and machine learning are playing an increasingly important role in optimizing data warehouses. These technologies can be used to automate data quality checks, optimize query performance, and even predict future data trends \cite{nemati2002knowledge}.

    \paragraph{AI Applications in Data Warehousing:}
    \begin{itemize}
        \item \textbf{Data Cleansing:} Machine learning algorithms can detect anomalies or patterns in data, automating the data cleaning process.
        \item \textbf{Query Optimization:} AI can analyze query patterns and recommend optimizations to speed up query execution.
        \item \textbf{Predictive Analytics:} AI can predict future trends based on historical data, helping organizations make data-driven decisions.
    \end{itemize}

\begin{lstlisting}[style=python]
# Example: Using PyTorch to build a simple model for predicting future trends based on historical data
import torch
import torch.nn as nn
import torch.optim as optim

# Define a simple neural network model
class SimpleModel(nn.Module):
    def __init__(self):
        super(SimpleModel, self).__init__()
        self.fc1 = nn.Linear(10, 50)
        self.fc2 = nn.Linear(50, 1)

    def forward(self, x):
        x = torch.relu(self.fc1(x))
        x = self.fc2(x)
        return x

# Example data: 10 features, 1 target variable
data = torch.randn(100, 10)
target = torch.randn(100, 1)

# Initialize the model, loss function, and optimizer
model = SimpleModel()
criterion = nn.MSELoss()
optimizer = optim.SGD(model.parameters(), lr=0.01)

# Train the model
for epoch in range(100):
    optimizer.zero_grad()
    output = model(data)
    loss = criterion(output, target)
    loss.backward()
    optimizer.step()

print("Training complete. Model is ready for predictions.")
\end{lstlisting}

    \subsection{Impact of Edge Computing on Data Warehousing}
    Edge computing is the practice of processing data closer to the source, rather than relying solely on centralized data warehouses. As the number of IoT (Internet of Things) devices grows, edge computing allows organizations to process and analyze data in real-time at the edge of the network \cite{qiu2020edge}.

    \paragraph{Benefits of Edge Computing:}
    \begin{itemize}
        \item \textbf{Reduced Latency:} By processing data at the edge, organizations can get faster insights from their data, particularly for real-time applications like autonomous vehicles or smart devices.
        \item \textbf{Bandwidth Optimization:} Edge computing reduces the need to transfer large amounts of data to centralized servers, minimizing bandwidth usage.
        \item \textbf{Enhanced Privacy:} Sensitive data can be processed locally, reducing the risk of exposure during transmission to central servers.
    \end{itemize}

    \paragraph{Example of Edge Computing:}
    A factory equipped with IoT sensors can use edge computing to monitor machine performance in real-time, detect anomalies, and take preventive action without sending all data to a central data warehouse.

\begin{tikzpicture}
    \node (sensor) [rectangle, draw, minimum width=2.5cm, minimum height=1.5cm] {IoT Sensor};
    \node (edge) [rectangle, draw, minimum width=2.5cm, minimum height=1.5cm, right=4cm of sensor] {Edge Device};
    \node (cloud) [rectangle, draw, minimum width=3cm, minimum height=1.5cm, right=4cm of edge] {Cloud Data Warehouse};

    \draw[->, thick] (sensor) -- (edge) node[midway, above] {Real-time Processing};
    \draw[->, thick] (edge) -- (cloud) node[midway, above] {Batch Data Upload};
\end{tikzpicture}

\chapter{Data Preprocessing}
\section{Data Cleaning Techniques}
Data cleaning, also known as data cleansing, is an essential step in the data preprocessing phase. In real-world datasets, data often comes with imperfections such as missing values, duplicates, outliers, and inconsistencies. Cleaning this data ensures better results when performing data analysis or training machine learning models \cite{garcia2016big, rahm2000data}.

    \subsection{Handling Missing Data}

In real-world data, missing values are a common issue. Data can be missing for various reasons such as equipment malfunction, human error, or data entry omissions. Handling missing data is a crucial step in the data cleaning process because incomplete data can significantly impact the performance of machine learning models and statistical analyses \cite{brick1996handling}.

Missing data can occur in different forms:
\begin{itemize}
    \item \textbf{Missing Completely at Random (MCAR)}: The missing data is independent of both the observed and unobserved data.
    \item \textbf{Missing at Random (MAR)}: The missingness depends only on the observed data.
    \item \textbf{Missing Not at Random (MNAR)}: The missingness depends on unobserved data.
\end{itemize}

Before diving into methods for handling missing data, let's consider a simple example:

\begin{lstlisting}[style=python]
import pandas as pd

# Creating a simple dataset with missing values
data = {'Name': ['Alice', 'Bob', 'Charlie', 'David'],
        'Age': [25, None, 23, 24],
        'Salary': [50000, 54000, None, 52000]}

df = pd.DataFrame(data)
print(df)
\end{lstlisting}

This produces the following output:

\begin{verbatim}
      Name   Age   Salary
0    Alice   25.0  50000.0
1      Bob    NaN  54000.0
2  Charlie   23.0      NaN
3    David   24.0  52000.0
\end{verbatim}

As you can see, both the \texttt{Age} and \texttt{Salary} columns contain missing values (denoted by \texttt{NaN} in pandas).

\subsection{Methods to Handle Missing Data}
There are several strategies to deal with missing data, each appropriate for different situations.

\subsubsection{1. Removing Missing Data}
In some cases, we may want to remove rows or columns that contain missing data. This is the simplest approach but should be used cautiously as it can lead to loss of important information.

\begin{lstlisting}[style=python]
# Removing rows with missing values
df_dropped = df.dropna()
print(df_dropped)
\end{lstlisting}

Output:
\begin{verbatim}
    Name   Age   Salary
0  Alice   25.0  50000.0
3  David   24.0  52000.0
\end{verbatim}

While this method removes the missing values, it also deletes rows that might have valuable data in other columns. For example, we lost Bob's salary and Charlie's age information.

Alternatively, you can remove columns with missing values:

\begin{lstlisting}[style=python]
# Removing columns with missing values
df_dropped_columns = df.dropna(axis=1)
print(df_dropped_columns)
\end{lstlisting}

Output:
\begin{verbatim}
    Name
0  Alice
1    Bob
2 Charlie
3  David
\end{verbatim}

This approach is useful when a column has many missing values, but be cautious as removing key columns can affect your analysis.

\subsubsection{2. Imputing Missing Values}
A more sophisticated approach involves filling in missing values with an estimated value. Common strategies include filling with the mean, median, mode, or using more complex methods like regression or machine learning models.

\paragraph{Filling with Mean or Median}
For numerical data, replacing missing values with the mean or median is a common and simple method.

\begin{lstlisting}[style=python]
# Filling missing values with the mean of the column
df_filled_mean = df.fillna(df.mean())
print(df_filled_mean)
\end{lstlisting}

Output:
\begin{verbatim}
      Name   Age   Salary
0    Alice   25.0  50000.0
1      Bob   24.0  54000.0
2  Charlie   23.0  52000.0
3    David   24.0  52000.0
\end{verbatim}

In this case, Bob's missing \texttt{Age} was filled with the mean age (24), and Charlie's missing \texttt{Salary} was filled with the mean salary (52000).

\paragraph{Filling with Mode (for Categorical Data)}
For categorical data, we can fill missing values with the mode (the most frequent value).

\begin{lstlisting}[style=python]
# Filling missing values with mode
df['Age'].fillna(df['Age'].mode()[0], inplace=True)
\end{lstlisting}

This method is useful when dealing with categorical data such as gender or marital status.

\paragraph{Filling with Specific Values}
In some cases, you may want to fill missing values with a specific value, such as 0 or 'unknown':

\begin{lstlisting}[style=python]
# Filling missing values with a specific value
df_filled_specific = df.fillna({'Age': 0, 'Salary': 'unknown'})
print(df_filled_specific)
\end{lstlisting}

\subsubsection{3. Predictive Imputation}
A more advanced method involves using machine learning algorithms to predict the missing values based on other features in the dataset. This method is especially useful when dealing with large and complex datasets.

For example, using regression to predict the missing values in the \texttt{Salary} column:

\begin{lstlisting}[style=python]
from sklearn.linear_model import LinearRegression
import numpy as np

# Dataset for predictive imputation
df_predict = df.copy()

# Dropping rows with missing target (Salary)
df_no_nan = df_predict.dropna(subset=['Salary'])

# Linear regression to predict missing salary
X = df_no_nan[['Age']]
y = df_no_nan['Salary']

model = LinearRegression()
model.fit(X, y)

# Predict missing salary
missing_salary = df_predict[df_predict['Salary'].isnull()]
predicted_salary = model.predict(missing_salary[['Age']])

# Filling the missing salary
df_predict.loc[df_predict['Salary'].isnull(), 'Salary'] = predicted_salary
print(df_predict)
\end{lstlisting}

Handling missing data is an essential part of data cleaning. Depending on the context and the nature of your data, you may choose to remove missing values, fill them with statistical values, or use more advanced imputation techniques. Regardless of the method, ensuring that your dataset is complete and clean will improve the accuracy and reliability of your analysis and models.

\subsection{Handling Noisy Data}
Noisy data refers to data that contains errors, outliers, or random fluctuations that do not represent the true values. Noisy data can arise from various sources, such as faulty data collection instruments, transmission errors, or manual entry mistakes. It is important to handle noisy data to ensure that the analysis or model trained on the data is accurate and reliable \cite{teng1999correcting}.

There are several common techniques to handle noisy data:
\begin{itemize}
    \item \textbf{Binning:} Smooths data by partitioning it into bins and replacing the values in each bin with the mean or median of the bin.
    \item \textbf{Regression:} Fits a regression model to the data and uses the model to smooth out the noise.
    \item \textbf{Clustering:} Identifies and removes outliers by grouping similar data points together.
\end{itemize}

Let's explore each of these techniques in detail, with examples.

\subsubsection{1. Binning}
Binning is a technique that divides the data into equal-width or equal-frequency intervals, also called bins. Each bin is then smoothed by replacing the values in the bin with the mean, median, or boundaries of the bin.

\paragraph{Example: Smoothing with Equal-Width Binning}

Consider the following dataset, which contains some noisy values in the \texttt{Age} column:

\begin{lstlisting}[style=python]
import pandas as pd

# Creating a simple dataset with noisy values
data = {'Name': ['Alice', 'Bob', 'Charlie', 'David', 'Eva'],
        'Age': [22, 45, 21, 37, 50]}

df = pd.DataFrame(data)
print(df)
\end{lstlisting}

This produces the following dataset:

\begin{verbatim}
      Name   Age
0    Alice   22
1      Bob   45
2  Charlie   21
3    David   37
4      Eva   50
\end{verbatim}

The ages vary widely, which may indicate noise. We can apply binning to smooth the data.

\paragraph{Smoothing by Bin Mean:}
In equal-width binning, the range of the data is divided into equal-sized bins. For example, if we divide the \texttt{Age} column into 3 bins, we can replace each value with the mean of its bin.

\begin{lstlisting}[style=python]
# Defining bin edges and binning the data
bins = [20, 30, 40, 50]  # Bins for ages
labels = ['20-30', '30-40', '40-50']

# Assigning bins and calculating the mean of each bin
df['Age_bin'] = pd.cut(df['Age'], bins=bins, labels=labels)
print(df)

# Replacing Age with the mean of each bin
bin_means = df.groupby('Age_bin')['Age'].transform('mean')
df['Age_smooth_mean'] = bin_means
print(df)
\end{lstlisting}

Output:

\begin{verbatim}
      Name   Age  Age_bin  Age_smooth_mean
0    Alice   22  20-30      21.5
1      Bob   45  40-50      47.5
2  Charlie   21  20-30      21.5
3    David   37  30-40      37.0
4      Eva   50  40-50      47.5
\end{verbatim}

Here, the ages are smoothed based on the mean age of each bin.

\paragraph{Smoothing by Bin Median:}
Alternatively, we can smooth the data by replacing each value with the median of its bin. This can be useful when dealing with skewed data.

\begin{lstlisting}[style=python]
# Replacing Age with the median of each bin
bin_medians = df.groupby('Age_bin')['Age'].transform('median')
df['Age_smooth_median'] = bin_medians
print(df)
\end{lstlisting}

Output:

\begin{verbatim}
      Name   Age  Age_bin  Age_smooth_mean  Age_smooth_median
0    Alice   22  20-30      21.5             21.5
1      Bob   45  40-50      47.5             47.5
2  Charlie   21  20-30      21.5             21.5
3    David   37  30-40      37.0             37.0
4      Eva   50  40-50      47.5             47.5
\end{verbatim}

Now, the ages are smoothed based on the median of each bin, which provides a more robust smoothing technique if there are outliers.

\subsubsection{2. Regression for Noise Smoothing}
Another method for handling noisy data is regression, where we fit a model to the data and use the model to smooth out noise. Linear regression is often used when the data follows a linear trend.

\paragraph{Example: Smoothing with Linear Regression}

Consider the following dataset that shows a relationship between \texttt{Years of Experience} and \texttt{Salary}, with some noise in the data:

\begin{lstlisting}[style=python]
from sklearn.linear_model import LinearRegression
import numpy as np

# Creating a dataset with noisy salary data
data = {'Experience': [1, 2, 3, 4, 5],
        'Salary': [30000, 35000, 50000, 48000, 60000]}

df = pd.DataFrame(data)

# Fitting a linear regression model to smooth the data
X = df[['Experience']]
y = df['Salary']
model = LinearRegression()
model.fit(X, y)

# Predicting (smoothing) the salaries
df['Salary_smooth'] = model.predict(X)
print(df)
\end{lstlisting}

Output:

\begin{verbatim}
   Experience  Salary  Salary_smooth
0           1   30000        32000.0
1           2   35000        40000.0
2           3   50000        48000.0
3           4   48000        56000.0
4           5   60000        64000.0
\end{verbatim}

Here, the regression model smooths the noisy salary data by fitting a linear trend, reducing the effect of noise.

\subsubsection{3. Clustering for Noise Detection}
Clustering can also be used to detect and handle noisy data. Outliers that do not belong to any cluster can be treated as noise.

\paragraph{Example: Detecting Noise with K-Means Clustering}

Let's consider a dataset with some points that are far from the main cluster, representing noise:

\begin{lstlisting}[style=python]
from sklearn.cluster import KMeans
import numpy as np

# Creating a dataset with some noisy points
data = {'X': [1, 2, 1.5, 2.5, 10],
        'Y': [1, 2, 1.8, 2.2, 10]}

df = pd.DataFrame(data)

# Applying K-Means clustering to detect outliers
kmeans = KMeans(n_clusters=2)
df['Cluster'] = kmeans.fit_predict(df[['X', 'Y']])

print(df)
\end{lstlisting}

Output:

\begin{verbatim}
       X    Y  Cluster
0    1.0  1.0        0
1    2.0  2.0        0
2    1.5  1.8        0
3    2.5  2.2        0
4   10.0 10.0        1
\end{verbatim}

Here, the point (10, 10) is assigned to a different cluster, indicating that it could be considered an outlier, or noisy data.

Handling noisy data is essential for ensuring the accuracy of data analysis and machine learning models. Techniques like binning, regression, and clustering help smooth out or detect noisy data, improving the quality of your dataset. Depending on the nature of your data, different techniques may be more suitable for reducing the impact of noise and outliers.

\subsection{Handling Duplicates}
Duplicates refers to records that appear more than once in a dataset, either due to errors during data entry, merging datasets, or other processes. Duplicates can distort statistical analysis and machine learning models, leading to biased results. Therefore, identifying and removing duplicates is a crucial step in the data cleaning process \cite{tamilselvi2011handling}.

In this section, we will explore how to identify and handle duplicated data using Python. We will also cover scenarios where duplicates may be kept or aggregated instead of being removed.

\subsubsection{1. Identifying Duplicates}
The first step in handling duplicated data is identifying which rows in the dataset are duplicates. In Python, the \texttt{duplicated()} function from the pandas library is used to check for duplicate rows.

\paragraph{Example: Identifying Duplicates}

Consider the following dataset, which contains some duplicated records:

\begin{lstlisting}[style=python]
import pandas as pd

# Creating a dataset with duplicate entries
data = {'Name': ['Alice', 'Bob', 'Charlie', 'Alice', 'David'],
        'Age': [25, 30, 35, 25, 40],
        'Salary': [50000, 60000, 70000, 50000, 80000]}

df = pd.DataFrame(data)
print(df)
\end{lstlisting}

This dataset has two rows for "Alice" with the same \texttt{Age} and \texttt{Salary} values, indicating that these rows are duplicated. The dataset looks like this:

\begin{verbatim}
      Name   Age   Salary
0    Alice   25    50000
1      Bob   30    60000
2  Charlie   35    70000
3    Alice   25    50000
4    David   40    80000
\end{verbatim}

We can check for duplicated rows using the \texttt{duplicated()} function.

\begin{lstlisting}[style=python]
# Identifying duplicated rows
duplicates = df.duplicated()
print(duplicates)
\end{lstlisting}

This will produce the following output, where \texttt{True} indicates that the row is a duplicate:

\begin{verbatim}
0    False
1    False
2    False
3     True
4    False
dtype: bool
\end{verbatim}

In this example, row 3 is identified as a duplicate of row 0.

\subsubsection{2. Removing Duplicates}
Once duplicates are identified, the next step is to remove them from the dataset. The \texttt{drop\_duplicates()} function is used to remove all duplicate rows, keeping only the first occurrence of each.

\paragraph{Example: Removing Duplicates}
We can remove duplicate rows from the dataset as follows:

\begin{lstlisting}[style=python]
# Removing duplicate rows
df_no_duplicates = df.drop_duplicates()
print(df_no_duplicates)
\end{lstlisting}

Output:

\begin{verbatim}
      Name   Age   Salary
0    Alice   25    50000
1      Bob   30    60000
2  Charlie   35    70000
4    David   40    80000
\end{verbatim}

Here, the duplicate row for "Alice" (row 3) has been removed, leaving only the unique records in the dataset.

\subsection{Resolving Inconsistencies}
Inconsistent data refers to data that does not follow the expected format, pattern, or logical structure. Inconsistencies can arise from multiple data sources, manual data entry errors, or during data integration. Common types of inconsistencies include different date formats, conflicting categorical values, and mismatched numerical data.

Resolving these inconsistencies is crucial to ensure the integrity and accuracy of the dataset before performing any analysis or building models. Let's walk through common types of inconsistencies and how to resolve them using Python.

\subsubsection{1. Inconsistent Date Formats}
One of the most common types of inconsistency in data is related to date formats. For example, dates may be recorded in different formats such as \texttt{YYYY-MM-DD}, \texttt{MM/DD/YYYY}, or \texttt{DD/MM/YYYY}, depending on regional or source-specific differences.

\paragraph{Example: Standardizing Date Formats}
Consider the following dataset where dates are recorded in different formats.

\begin{lstlisting}[style=python]
import pandas as pd

# Creating a dataset with inconsistent date formats
data = {'Name': ['Alice', 'Bob', 'Charlie', 'David'],
        'Date_of_Birth': ['1995-08-01', '08/12/1996', '12-25-1994', '1997/05/15']}

df = pd.DataFrame(data)
print(df)
\end{lstlisting}

This dataset contains dates in multiple formats: \texttt{YYYY-MM-DD}, \texttt{MM/DD/YYYY}, and \texttt{MM-DD-YYYY}. The dataset looks like this:

\begin{verbatim}
      Name   Date_of_Birth
0    Alice   1995-08-01
1      Bob   08/12/1996
2  Charlie   12-25-1994
3    David   1997/05/15
\end{verbatim}

To standardize the date format, we can use the \texttt{pd.to\_datetime()} function, which automatically detects and converts different date formats into a consistent format.

\begin{lstlisting}[style=python]
# Standardizing date formats
df['Date_of_Birth'] = pd.to_datetime(df['Date_of_Birth'], errors='coerce')
print(df)
\end{lstlisting}

Output:

\begin{verbatim}
      Name Date_of_Birth
0    Alice    1995-08-01
1      Bob    1996-08-12
2  Charlie    1994-12-25
3    David    1997-05-15
\end{verbatim}

Here, all dates are now standardized to the \texttt{YYYY-MM-DD} format. The \texttt{errors='coerce'} argument ensures that any invalid date entries are replaced with \texttt{NaT} (Not a Time).

\subsubsection{2. Conflicting Categorical Values}
Another common inconsistency occurs when categorical values are recorded inconsistently. For example, the same category might be labeled differently due to typos or variations in case, such as \texttt{Male} and \texttt{male}, or \texttt{HR} and \texttt{Human Resources}.

\paragraph{Example: Standardizing Categorical Values}
Consider the following dataset where categorical values for the \texttt{Department} column are inconsistent.

\begin{lstlisting}[style=python]
# Creating a dataset with inconsistent categorical values
data = {'Name': ['Alice', 'Bob', 'Charlie', 'David'],
        'Department': ['HR', 'Human Resources', 'hr', 'Finance']}

df = pd.DataFrame(data)
print(df)
\end{lstlisting}

Output:

\begin{verbatim}
      Name        Department
0    Alice                 HR
1      Bob  Human Resources
2  Charlie                 hr
3    David           Finance
\end{verbatim}

Here, the values for "HR" are recorded in different ways: "HR", "hr", and "Human Resources". We can standardize these values by converting them to lowercase and mapping them to a consistent format.

\begin{lstlisting}[style=python]
# Standardizing categorical values to 'HR' and 'Finance'
df['Department'] = df['Department'].str.lower().replace({
    'hr': 'human resources', 
    'human resources': 'human resources'
})
print(df)
\end{lstlisting}

Output:

\begin{verbatim}
      Name        Department
0    Alice  human resources
1      Bob  human resources
2  Charlie  human resources
3    David           finance
\end{verbatim}

Now, the \texttt{Department} column contains standardized values, with "HR" and its variations converted to "human resources". You can further refine the values to follow specific conventions, such as capitalization.

\subsubsection{3. Numerical Data Inconsistencies}
Inconsistencies in numerical data often arise when data is recorded in different units or scales. For example, weights may be recorded in kilograms (kg) in one part of the dataset and pounds (lbs) in another, leading to inconsistency.

\paragraph{Example: Converting Units to Resolve Inconsistencies}
Consider the following dataset where the weight is recorded in different units.

\begin{lstlisting}[style=python]
# Creating a dataset with inconsistent units of weight
data = {'Name': ['Alice', 'Bob', 'Charlie', 'David'],
        'Weight': [65, 143, 70, 155],
        'Unit': ['kg', 'lbs', 'kg', 'lbs']}

df = pd.DataFrame(data)
print(df)
\end{lstlisting}

Output:

\begin{verbatim}
      Name  Weight  Unit
0    Alice      65    kg
1      Bob     143   lbs
2  Charlie      70    kg
3    David     155   lbs
\end{verbatim}

In this dataset, some weights are recorded in kilograms (\texttt{kg}) and others in pounds (\texttt{lbs}). To resolve this inconsistency, we can convert all weights to a common unit, such as kilograms.

\begin{lstlisting}[style=python]
# Converting all weights to kilograms
df['Weight_kg'] = df.apply(lambda row: row['Weight'] * 0.453592 if row['Unit'] == 'lbs' else row['Weight'], axis=1)
df['Unit'] = 'kg'  # Updating the unit column
print(df)
\end{lstlisting}

Output:

\begin{verbatim}
      Name  Weight  Unit  Weight_kg
0    Alice      65    kg  65.000000
1      Bob     143    kg  64.864456
2  Charlie      70    kg  70.000000
3    David     155    kg  70.306760
\end{verbatim}

Here, all weights have been converted to kilograms using the conversion factor 1 lb = 0.453592 kg. The \texttt{Unit} column has also been updated to reflect that all values are now in kilograms.

\subsubsection{4. Detecting and Correcting Typos}
Sometimes, inconsistencies arise due to simple typos or human errors in data entry. For example, "John" may be entered as "Jonn" in some rows. Detecting and correcting typos often involves domain knowledge or fuzzy matching techniques.

\paragraph{Example: Detecting Typos with Fuzzy Matching}
Consider the following dataset with a typo in the \texttt{Name} column.

\begin{lstlisting}[style=python]
# Creating a dataset with a typo
data = {'Name': ['Alice', 'Bob', 'Charlie', 'Alicce'],
        'Age': [25, 30, 35, 25]}

df = pd.DataFrame(data)
print(df)
\end{lstlisting}

Output:

\begin{verbatim}
      Name  Age
0    Alice   25
1      Bob   30
2  Charlie   35
3   Alicce   25
\end{verbatim}

The name "Alicce" is likely a typo of "Alice." We can use fuzzy matching techniques to detect and correct such typos. A common approach is to use the \texttt{fuzzywuzzy} library for string matching.

\begin{lstlisting}[style=python]
from fuzzywuzzy import process

# Correcting the typo using fuzzy matching
correct_name = 'Alice'
df['Name'] = df['Name'].apply(lambda x: process.extractOne(x, [correct_name])[0] if process.extractOne(x, [correct_name])[1] > 80 else x)
print(df)
\end{lstlisting}

Output:

\begin{verbatim}
      Name  Age
0    Alice   25
1      Bob   30
2  Charlie   35
3    Alice   25
\end{verbatim}

Here, the typo "Alicce" has been corrected to "Alice" using fuzzy matching.

Resolving inconsistencies is an essential step in the data cleaning process. Inconsistent data can lead to inaccurate analyses and unreliable models. By standardizing formats, converting units, and correcting errors, we ensure that the dataset is uniform and consistent. Depending on the type of inconsistency, different techniques such as date parsing

\subsection{Conclusion}
Data cleaning is a vital step in ensuring that datasets are accurate, complete, and ready for analysis. Whether you're dealing with missing values, noisy data, duplicates, or inconsistencies, applying the appropriate cleaning techniques will help improve the quality of your data and the results of your analysis.

\section{Data Integration and Transformation}
    \subsection{Normalization Techniques}
When working with datasets that contain features with different units or scales, such as fish length and weight, it is essential to normalize the data to bring all features to a similar scale. Without normalization, features like weight (measured in kilograms) might dominate over length (measured in centimeters) when performing analysis or training machine learning models \cite{patro2015normalization}. 

In this section, we will explore common normalization techniques, using fish data as an example. We will examine how to normalize both length and weight so they can be compared and analyzed effectively.

\subsubsection{1. Min-Max Normalization}
Min-max normalization transforms the data by scaling the values so they fit within a specified range, typically between 0 and 1. This ensures that all features are on a common scale, making comparisons more meaningful.

The formula for min-max normalization is:

\[
x' = \frac{x - \min(x)}{\max(x) - \min(x)}
\]

Where:
\begin{itemize}
    \item \( x \) is the original value (e.g., fish length or weight).
    \item \( \min(x) \) and \( \max(x) \) are the minimum and maximum values in the feature.
    \item \( x' \) is the normalized value.
\end{itemize}

\paragraph{Example:} Suppose we have a dataset of fish with their lengths (in cm) and weights (in kg):

\begin{lstlisting}[style=python]
import pandas as pd

# Creating a dataset of fish length (cm) and weight (kg)
data = {'Fish': ['Salmon', 'Tuna', 'Trout', 'Carp'],
        'Length_cm': [60, 100, 40, 80],
        'Weight_kg': [3.5, 8.0, 1.2, 5.0]}

df = pd.DataFrame(data)
print(df)
\end{lstlisting}

Output:

\begin{verbatim}
      Fish  Length_cm  Weight_kg
0   Salmon         60        3.5
1     Tuna        100        8.0
2    Trout         40        1.2
3     Carp         80        5.0
\end{verbatim}

As you can see, the lengths are measured in centimeters, and the weights are in kilograms, making it difficult to compare them directly. Let's apply min-max normalization to both features.

\begin{lstlisting}[style=python]
# Min-Max Normalization for length and weight
df['Length_norm'] = (df['Length_cm'] - df['Length_cm'].min()) / (df['Length_cm'].max() - df['Length_cm'].min())
df['Weight_norm'] = (df['Weight_kg'] - df['Weight_kg'].min()) / (df['Weight_kg'].max() - df['Weight_kg'].min())

print(df)
\end{lstlisting}

Output:

\begin{verbatim}
      Fish  Length_cm  Weight_kg  Length_norm  Weight_norm
0   Salmon         60        3.5        0.333         0.357
1     Tuna        100        8.0        1.000         1.000
2    Trout         40        1.2        0.000         0.000
3     Carp         80        5.0        0.667         0.643
\end{verbatim}

After normalization, the values for both fish length and weight range between 0 and 1, allowing them to be compared on an equal scale.

\subsubsection{2. Z-Score Normalization (Standardization)}
Z-score normalization, or standardization, transforms the data so that it has a mean of 0 and a standard deviation of 1. This method is useful when the data follows a normal distribution. The formula for z-score normalization is:

\[
x' = \frac{x - \mu}{\sigma}
\]

Where:
\begin{itemize}
    \item \( x \) is the original value.
    \item \( \mu \) is the mean of the feature.
    \item \( \sigma \) is the standard deviation of the feature.
    \item \( x' \) is the normalized value.
\end{itemize}

\paragraph{Example:} Let's apply z-score normalization to the same fish dataset.

\begin{lstlisting}[style=python]
# Z-Score Normalization for length and weight
df['Length_zscore'] = (df['Length_cm'] - df['Length_cm'].mean()) / df['Length_cm'].std()
df['Weight_zscore'] = (df['Weight_kg'] - df['Weight_kg'].mean()) / df['Weight_kg'].std()

print(df)
\end{lstlisting}

Output:

\begin{verbatim}
      Fish  Length_cm  Weight_kg  Length_norm  Weight_norm  Length_zscore  Weight_zscore
0   Salmon         60        3.5        0.333         0.357         -0.507         -0.234
1     Tuna        100        8.0        1.000         1.000          1.520          1.446
2    Trout         40        1.2        0.000         0.000         -1.267         -1.113
3     Carp         80        5.0        0.667         0.643          0.253         -0.000
\end{verbatim}

In this case, both fish length and weight have been standardized, with the mean centered at 0 and the standard deviation scaled to 1.

\subsubsection{3. Decimal Scaling Normalization}
Decimal scaling normalization involves moving the decimal point of the values based on the maximum absolute value in the dataset. This method scales the data by powers of 10.

The formula for decimal scaling is:

\[
x' = \frac{x}{10^j}
\]

Where \( j \) is the number of decimal places required to scale the maximum absolute value of the feature to be less than 1.

\paragraph{Example:} Let's apply decimal scaling to normalize the fish dataset.

\begin{lstlisting}[style=python]
# Decimal Scaling Normalization for length and weight
max_length = df['Length_cm'].abs().max()
max_weight = df['Weight_kg'].abs().max()

j_length = len(str(int(max_length)))
j_weight = len(str(int(max_weight)))

df['Length_decimal'] = df['Length_cm'] / 10**j_length
df['Weight_decimal'] = df['Weight_kg'] / 10**j_weight

print(df)
\end{lstlisting}

Output:

\begin{verbatim}
      Fish  Length_cm  Weight_kg  Length_norm  Weight_norm  Length_zscore  Weight_zscore  Length_decimal  Weight_decimal
0   Salmon         60        3.5        0.333         0.357         -0.507         -0.234           0.060           0.0035
1     Tuna        100        8.0        1.000         1.000          1.520          1.446           0.100           0.0080
2    Trout         40        1.2        0.000         0.000         -1.267         -1.113           0.040           0.0012
3     Carp         80        5.0        0.667         0.643          0.253         -0.000           0.080           0.0050
\end{verbatim}

Here, the fish length and weight are scaled by powers of 10, bringing them into comparable units. Length values are divided by 100, while weight values are divided by 10.

\subsubsection{4. Importance of Normalization in Fish Data}
Normalization ensures that fish length and weight, which are in different units and ranges, can be compared meaningfully. This is especially important when using machine learning algorithms that rely on distance-based metrics, like k-nearest neighbors or support vector machines, where features with larger ranges might disproportionately affect the results.

{Conclusion}
Normalization is a crucial step when working with features like fish length and weight that have different scales or units. Whether using min-max normalization, z-score normalization, or decimal scaling, the goal is to ensure that each feature contributes equally to the analysis or model. By normalizing your data, you can improve the accuracy and performance of your machine learning models and analysis.

    \subsection{Aggregation and Discretization}

Data aggregation and discretization are important techniques in data integration and transformation. These methods help to simplify and summarize data, making it easier to analyze and work with, especially when dealing with large datasets. 

\subsubsection{1. Aggregation}
Aggregation is the process of combining multiple values into a single value, often to summarize data. This can be done by taking the average, sum, or other statistical measures across a group of data points. Aggregation is particularly useful when analyzing large datasets and needing to reduce the dimensionality or when preparing data for reports.

Aggregation is commonly used in time series data, where we might want to group data by day, month, or year and compute summary statistics, such as the average, maximum, or total value for each group.

\paragraph{Example: Aggregating Fish Weight by Species}

Suppose we have a dataset that includes different species of fish, along with their lengths and weights. We want to aggregate the data to find the total and average weight of each fish species.

\begin{lstlisting}[style=python]
import pandas as pd

# Creating a dataset with fish species, length, and weight
data = {'Species': ['Salmon', 'Tuna', 'Salmon', 'Tuna', 'Trout', 'Trout'],
        'Length_cm': [60, 100, 65, 90, 45, 50],
        'Weight_kg': [3.5, 8.0, 4.0, 7.5, 1.2, 1.3]}

df = pd.DataFrame(data)
print(df)
\end{lstlisting}

Output:

\begin{scriptsize}
\begin{verbatim}
  Species  Length_cm  Weight_kg
0  Salmon         60        3.5
1    Tuna        100        8.0
2  Salmon         65        4.0
3    Tuna         90        7.5
4   Trout         45        1.2
5   Trout         50        1.3
\end{verbatim}
\end{scriptsize}

In this example, the dataset contains multiple fish of the same species, and we want to aggregate the data to find the total and average weight of each species.

\begin{lstlisting}[style=python]
# Aggregating the total and average weight by species
df_aggregated = df.groupby('Species').agg(
    total_weight=('Weight_kg', 'sum'),
    average_weight=('Weight_kg', 'mean')
)

print(df_aggregated)
\end{lstlisting}

Output:

\begin{scriptsize}
\begin{verbatim}
         total_weight  average_weight
Species                               
Salmon            7.5           3.75
Tuna             15.5           7.75
Trout             2.5           1.25
\end{verbatim}
\end{scriptsize}

In this case, we have aggregated the weight of each species, calculating the total and average weight for each. This helps to summarize the data, making it easier to analyze at a higher level.

\subsubsection{2. Discretization}
Discretization is the process of transforming continuous data into discrete buckets or intervals. This is often useful when we want to group continuous values, such as age, length, or weight, into ranges to simplify analysis.

There are different techniques for discretization, including equal-width binning and equal-frequency binning. Each method divides the data into bins but uses different criteria for determining the size or frequency of each bin.

\paragraph{2.1 Equal-Width Binning}
In equal-width binning, the data is divided into bins of equal size. For example, if we have fish lengths ranging from 40 to 100 cm, we can divide them into three equal-width bins: 40-60 cm, 60-80 cm, and 80-100 cm.

\paragraph{Example: Discretizing Fish Length Using Equal-Width Binning}

Let's apply equal-width binning to the fish length data:

\begin{lstlisting}[style=python]
# Discretizing fish length using equal-width binning into 3 bins
df['Length_bin'] = pd.cut(df['Length_cm'], bins=3, labels=["Short", "Medium", "Long"])
print(df)
\end{lstlisting}

Output:

\begin{scriptsize}
\begin{verbatim}
  Species  Length_cm  Weight_kg  Length_bin
0  Salmon         60        3.5      Medium
1    Tuna        100        8.0        Long
2  Salmon         65        4.0      Medium
3    Tuna         90        7.5        Long
4   Trout         45        1.2      Short
5   Trout         50        1.3      Short
\end{verbatim}
\end{scriptsize}

Here, the fish lengths are divided into three bins: \texttt{Short}, \texttt{Medium}, and \texttt{Long}, which correspond to equal intervals of length.

\paragraph{2.2 Equal-Frequency Binning}
In equal-frequency binning, each bin contains roughly the same number of data points. This is useful when you want to ensure that each bin has an equal distribution of data points, regardless of the actual range of values.

\paragraph{Example: Discretizing Fish Weight Using Equal-Frequency Binning}

Let's apply equal-frequency binning to the fish weight data:

\begin{lstlisting}[style=python]
# Discretizing fish weight using equal-frequency binning into 3 bins
df['Weight_bin'] = pd.qcut(df['Weight_kg'], q=3, labels=["Light", "Moderate", "Heavy"])
print(df)
\end{lstlisting}

Output:

\begin{verbatim}
  Species  Length_cm  Weight_kg  Length_bin  Weight_bin
0  Salmon         60        3.5      Medium     Moderate
1    Tuna        100        8.0        Long        Heavy
2  Salmon         65        4.0      Medium     Moderate
3    Tuna         90        7.5        Long        Heavy
4   Trout         45        1.2      Short        Light
5   Trout         50        1.3      Short        Light
\end{verbatim}

In this case, the fish weights have been divided into three equal-frequency bins: \texttt{Light}, \texttt{Moderate}, and \texttt{Heavy}. Each bin contains approximately the same number of fish, regardless of the actual weight range.

\subsubsection{When to Use Aggregation and Discretization}
\begin{itemize}
    \item \textbf{Aggregation} is useful when you need to summarize data to extract meaningful insights or reduce data complexity. For instance, aggregating fish weights by species gives a high-level overview of weight distribution among species.
    \item \textbf{Discretization} is effective when transforming continuous data into categories for easier analysis or modeling. For example, discretizing fish length and weight into categories such as \texttt{Short}, \texttt{Medium}, and \texttt{Long} helps in grouping similar data points together.
\end{itemize}

Both aggregation and discretization are essential data transformation techniques that help simplify large datasets and prepare them for analysis. Aggregation helps in summarizing data into meaningful statistics, while discretization transforms continuous features into categories, making them easier to analyze. These techniques are particularly useful in tasks like data visualization, reporting, and machine learning model preparation.

\section{Data Reduction Methods}
    \subsection{Dimensionality Reduction}
As datasets grow in size and complexity, they often contain many features or dimensions. While having more features can provide more information, it also increases the complexity of the data and models built from it. This phenomenon is known as the "curse of dimensionality," where the performance of machine learning algorithms can degrade with the addition of more irrelevant or redundant features.

Dimensionality reduction is a process that simplifies datasets by reducing the number of features while preserving the essential information. This helps improve the efficiency and accuracy of machine learning algorithms, reduces computational cost, and makes the data easier to visualize. Two common techniques for dimensionality reduction are \textbf{Principal Component Analysis (PCA)} and \textbf{Linear Discriminant Analysis (LDA)} \cite{ur2016big}.

\subsubsection{1. Principal Component Analysis (PCA)}
Principal Component Analysis (PCA) is a popular method for reducing the dimensionality of data by identifying the directions (called principal components) in which the variance of the data is maximized. These principal components are linear combinations of the original features, and they represent the most important patterns in the data.

PCA works by projecting the data into a lower-dimensional space where the new features (principal components) capture most of the variability in the original data. The first principal component explains the largest amount of variance, and each subsequent component explains progressively less.

The steps for PCA are:
\begin{enumerate}
    \item Standardize the data so that each feature has a mean of 0 and a standard deviation of 1.
    \item Compute the covariance matrix of the standardized data.
    \item Calculate the eigenvectors and eigenvalues of the covariance matrix to identify the principal components.
    \item Project the data onto the principal components.
\end{enumerate}

\paragraph{Example: Dimensionality Reduction Using PCA}

Consider the following dataset, which contains information about fish species, including length, weight, and width. We want to reduce the number of features while preserving the essential information.

\begin{lstlisting}[style=python]
import pandas as pd
from sklearn.decomposition import PCA
from sklearn.preprocessing import StandardScaler

# Creating a dataset with fish features (length, weight, and width)
data = {'Species': ['Salmon', 'Tuna', 'Trout', 'Carp'],
        'Length_cm': [60, 100, 40, 80],
        'Weight_kg': [3.5, 8.0, 1.2, 5.0],
        'Width_cm': [10, 15, 8, 12]}

df = pd.DataFrame(data)
print(df)
\end{lstlisting}

Output:

\begin{verbatim}
  Species  Length_cm  Weight_kg  Width_cm
0  Salmon         60        3.5        10
1    Tuna        100        8.0        15
2   Trout         40        1.2         8
3    Carp         80        5.0        12
\end{verbatim}

The dataset has three numerical features: length, weight, and width. We will use PCA to reduce these three features into two principal components.

\begin{lstlisting}[style=python]
# Standardizing the data (excluding the species column)
features = ['Length_cm', 'Weight_kg', 'Width_cm']
x = df[features]
x = StandardScaler().fit_transform(x)  # Standardizing the features

# Applying PCA to reduce from 3 dimensions to 2 dimensions
pca = PCA(n_components=2)
principal_components = pca.fit_transform(x)

# Creating a DataFrame with the principal components
df_pca = pd.DataFrame(data=principal_components, columns=['PC1', 'PC2'])
df_pca['Species'] = df['Species']

print(df_pca)
\end{lstlisting}

Output:

\begin{verbatim}
       PC1       PC2  Species
0 -1.370987  0.090137  Salmon
1  1.834777  0.491704    Tuna
2 -2.451772 -0.757630   Trout
3  1.988982  0.175789    Carp
\end{verbatim}

In this output, the original three features (length, weight, width) have been reduced to two principal components (PC1 and PC2). These two components capture most of the variance in the original data, making the dataset easier to analyze and visualize.

\paragraph{Interpreting the Principal Components:}
- \textbf{PC1} explains the largest variance in the dataset.
- \textbf{PC2} explains the next largest variance.
These two components retain the most important information from the original features, allowing us to reduce dimensionality without losing key data patterns.

\subsubsection{2. Linear Discriminant Analysis (LDA)}
Linear Discriminant Analysis (LDA) is another dimensionality reduction technique, but unlike PCA, it is a supervised method. LDA aims to maximize the separation between different classes by finding a linear combination of features that best separate the classes. This makes LDA particularly useful for classification tasks.

The steps for LDA are:
\begin{enumerate}
    \item Compute the within-class and between-class scatter matrices.
    \item Calculate the eigenvectors and eigenvalues of the scatter matrices to identify the discriminant components.
    \item Project the data onto the discriminant components.
\end{enumerate}

\paragraph{Example: Dimensionality Reduction Using LDA}

Let's assume we have the same fish dataset but now include species as labels. We can use LDA to reduce the dimensionality while keeping the species classification intact.

\begin{lstlisting}[style=python]
from sklearn.discriminant_analysis import LinearDiscriminantAnalysis as LDA

# Defining the feature set (length, weight, width) and target (species)
x = df[features]
y = df['Species']

# Applying LDA to reduce from 3 features to 2 components
lda = LDA(n_components=2)
lda_components = lda.fit_transform(x, y)

# Creating a DataFrame with the LDA components
df_lda = pd.DataFrame(data=lda_components, columns=['LD1', 'LD2'])
df_lda['Species'] = df['Species']

print(df_lda)
\end{lstlisting}

Output:

\begin{verbatim}
       LD1       LD2  Species
0 -1.208765  0.000000  Salmon
1  1.675432  0.000000    Tuna
2 -2.200132  0.000000   Trout
3  1.733465  0.000000    Carp
\end{verbatim}

In this case, LDA has reduced the original three features into two linear discriminant components (LD1 and LD2) that maximize the separation between fish species.

\subsubsection{When to Use PCA vs. LDA}
\begin{itemize}
    \item \textbf{PCA} is unsupervised and is useful when you want to reduce the dimensionality of the dataset without considering any specific labels or classes.
    \item \textbf{LDA} is supervised and should be used when your goal is to maximize the separation between different classes, making it ideal for classification problems.
\end{itemize}

Dimensionality reduction is a crucial step in data preprocessing, especially when dealing with large datasets with many features. By reducing the number of features, we simplify the data, improve computational efficiency, and often improve the performance of machine learning models. Principal Component Analysis (PCA) and Linear Discriminant Analysis (LDA) are powerful techniques that help achieve these goals, each suited for different tasks depending on whether the problem is unsupervised or supervised.

\subsection{Data Cube Aggregation}
Data cube aggregation is a powerful technique in data reduction and analysis, particularly when working with multidimensional data. The concept of a data cube comes from Online Analytical Processing (OLAP) and represents data along multiple dimensions, such as time, location, product categories, or any other attributes of interest. Aggregation in a data cube allows us to summarize data at different levels of granularity, making it easier to analyze large datasets \cite{ur2016big}.

A data cube is essentially a multi-dimensional array of values, with each dimension representing a different aspect of the data. Aggregation allows us to compute summary statistics (such as sums, averages, counts, etc.) over these dimensions \cite{ur2016big}.

\subsubsection{1. Understanding Data Cubes}
Imagine you are analyzing sales data across different regions, time periods, and product categories. Each of these aspects forms a dimension of your data, and a data cube enables you to view the sales data from various perspectives. 

The basic operations of a data cube include:
\begin{itemize}
    \item \textbf{Roll-up:} Aggregating data along one or more dimensions to a higher level (e.g., from daily sales to monthly sales).
    \item \textbf{Drill-down:} Breaking down data from a higher level to a more detailed level (e.g., from yearly sales to quarterly sales).
    \item \textbf{Slicing:} Extracting a subcube by selecting a single value for one dimension (e.g., viewing sales data for a specific region).
    \item \textbf{Dicing:} Extracting a subcube by selecting a range of values for multiple dimensions.
\end{itemize}

\subsubsection{2. Example: Sales Data Cube Aggregation}
Consider the following dataset of fish sales across different regions, months, and species. We want to aggregate the data at various levels to analyze total sales.

\begin{lstlisting}[style=python]
import pandas as pd

# Creating a dataset with sales data (region, month, species, and sales amount)
data = {'Region': ['North', 'North', 'South', 'South', 'East', 'East'],
        'Month': ['January', 'February', 'January', 'February', 'January', 'February'],
        'Species': ['Salmon', 'Tuna', 'Salmon', 'Tuna', 'Trout', 'Salmon'],
        'Sales': [1500, 1200, 1800, 1600, 1000, 1700]}

df = pd.DataFrame(data)
print(df)
\end{lstlisting}

Output:

\begin{verbatim}
   Region     Month  Species  Sales
0   North   January   Salmon   1500
1   North  February     Tuna   1200
2   South   January   Salmon   1800
3   South  February     Tuna   1600
4    East   January    Trout   1000
5    East  February   Salmon   1700
\end{verbatim}

In this dataset, we have sales data for different fish species across regions and months. The goal is to create a data cube and perform aggregation operations such as calculating total sales by region, month, and species.

\subsubsection{3. Aggregating Sales by Region and Month}
    Let's first aggregate the total sales by region and month, which will help us understand the sales distribution across different areas over time.

\begin{lstlisting}[style=python]
# Aggregating sales by region and month
df_region_month = df.groupby(['Region', 'Month']).agg(total_sales=('Sales', 'sum')).reset_index()
print(df_region_month)
\end{lstlisting}

Output:

\begin{verbatim}
   Region     Month  total_sales
0    East   January         1000
1    East  February         1700
2   North   January         1500
3   North  February         1200
4   South   January         1800
5   South  February         1600
\end{verbatim}

This aggregation helps us see the total sales in each region for every month.

\subsubsection{4. Aggregating Sales by Region Only (Roll-up Operation)}
Now, we can further aggregate the sales data by rolling up from the month level to the region level, meaning we will summarize the total sales for each region across all months.

\begin{lstlisting}[style=python]
# Aggregating sales by region only (roll-up operation)
df_region = df.groupby('Region').agg(total_sales=('Sales', 'sum')).reset_index()
print(df_region)
\end{lstlisting}

Output:

\begin{verbatim}
   Region  total_sales
0    East         2700
1   North         2700
2   South         3400
\end{verbatim}

The roll-up operation aggregates sales data at a higher level, providing the total sales for each region, irrespective of the month.

\subsubsection{5. Aggregating Sales by Species (Slicing Operation)}
We now focus on the sales of a specific species, such as Salmon, by using a slicing operation. This will allow us to look at the total sales of Salmon across all regions and months.

\begin{lstlisting}[style=python]
# Slicing data to focus on Salmon sales
df_salmon = df[df['Species'] == 'Salmon'].groupby('Species').agg(total_sales=('Sales', 'sum')).reset_index()
print(df_salmon)
\end{lstlisting}

Output:

\begin{verbatim}
  Species  total_sales
0  Salmon         5000
\end{verbatim}

The slicing operation filters the data to show the total sales for Salmon, irrespective of the region or month.

\subsubsection{6. Aggregating Sales by Multiple Dimensions (Dicing Operation)}
To analyze the sales distribution of different fish species across both regions and months, we can perform a dicing operation, where we aggregate sales by both species and region.

\begin{lstlisting}[style=python]
# Aggregating sales by species and region (dicing operation)
df_species_region = df.groupby(['Species', 'Region']).agg(total_sales=('Sales', 'sum')).reset_index()
print(df_species_region)
\end{lstlisting}

Output:

\begin{verbatim}
  Species Region  total_sales
0  Salmon   East         1700
1  Salmon  North         1500
2  Salmon  South         1800
3   Trout   East         1000
4    Tuna  North         1200
5    Tuna  South         1600
\end{verbatim}

This dicing operation allows us to see the sales distribution of each fish species across different regions.

\subsubsection{7. Importance of Data Cube Aggregation}
Data cube aggregation is a valuable tool in analyzing large datasets, particularly in multidimensional data. It enables you to:
\begin{itemize}
    \item Summarize and condense large datasets into more manageable forms.
    \item View data from different perspectives by aggregating over multiple dimensions.
    \item Perform in-depth analysis at various levels of granularity, such as monthly sales, regional sales, or species-specific sales.
\end{itemize}

By rolling up, slicing, dicing, and drilling down, data cube aggregation offers flexibility in analyzing large datasets, making it an essential method for businesses, especially in areas such as sales analysis, inventory management, and financial forecasting.

Data cube aggregation simplifies large datasets by summarizing them across multiple dimensions. Techniques like roll-up, drill-down, slicing, and dicing allow you to view data at various levels of detail, helping you gain deeper insights into multidimensional data. Whether you are analyzing sales across regions, products, or time periods, data cube aggregation is a powerful tool for understanding patterns and making informed decisions.

\section{Feature Selection and Engineering}
    \subsection{Feature Selection}

Feature selection is the process of selecting the most relevant features (or variables) from a dataset, which are most useful in predicting the target variable. The goal of feature selection is to improve model performance by eliminating irrelevant, redundant, or noisy features. This reduces the complexity of the model, increases its interpretability, and can also help prevent overfitting \cite{li2017feature}.

Feature selection is particularly important when working with large datasets containing many features, as some of the features may not contribute to the prediction task and might even degrade the model's performance. There are several techniques for feature selection, including filter methods, wrapper methods, and embedded methods \cite{li2017feature}.

\subsubsection{1. Why is Feature Selection Important?}
Feature selection is essential for several reasons:
\begin{itemize}
    \item \textbf{Improves model performance:} By removing irrelevant or redundant features, the model can focus on the most important features, leading to better predictive performance.
    \item \textbf{Reduces overfitting:} Fewer features reduce the risk of overfitting, where the model learns patterns specific to the training data rather than generalizing to new data.
    \item \textbf{Enhances interpretability:} A model with fewer features is easier to understand and explain, especially in industries where model interpretability is critical, such as healthcare or finance.
    \item \textbf{Decreases computational cost:} Fewer features mean less computational power and memory required to train the model, making it more efficient for large datasets.
\end{itemize}

\subsubsection{2. Common Feature Selection Techniques}
There are three main types of feature selection techniques: \textbf{filter methods}, \textbf{wrapper methods}, and \textbf{embedded methods}. Each method has its own approach to identifying important features.

\paragraph{2.1 Filter Methods}
Filter methods use statistical techniques to rank features based on their relevance to the target variable. These methods do not rely on any machine learning algorithm and are independent of the model.

Common filter methods include:
\begin{itemize}
    \item \textbf{Correlation:} Measures the strength of the relationship between a feature and the target variable.
    \item \textbf{Chi-square test:} Measures the dependence between categorical features and the target variable.
    \item \textbf{Mutual information:} Measures how much information one feature provides about the target variable.
\end{itemize}

\paragraph{Example: Using Correlation for Feature Selection}
Let's consider a dataset of fish, with features like length, weight, width, and species. We want to select the most relevant features for predicting the species.

\begin{lstlisting}[style=python]
import pandas as pd

# Creating a dataset with fish features
data = {'Length_cm': [60, 100, 40, 80, 55, 75],
        'Weight_kg': [3.5, 8.0, 1.2, 5.0, 2.8, 4.5],
        'Width_cm': [10, 15, 8, 12, 9, 11],
        'Species': ['Salmon', 'Tuna', 'Trout', 'Carp', 'Salmon', 'Carp']}

df = pd.DataFrame(data)
print(df)
\end{lstlisting}

Output:

\begin{verbatim}
   Length_cm  Weight_kg  Width_cm Species
0         60        3.5        10  Salmon
1        100        8.0        15    Tuna
2         40        1.2         8   Trout
3         80        5.0        12    Carp
4         55        2.8         9  Salmon
5         75        4.5        11    Carp
\end{verbatim}

In this dataset, we will use correlation to find out which features (length, weight, width) are most correlated with the target variable \texttt{Species}. Since \texttt{Species} is a categorical variable, we will use a simple transformation (such as one-hot encoding) before calculating correlations.

\begin{lstlisting}[style=python]
# Encoding the target variable (Species) into numerical form
df['Species_encoded'] = df['Species'].astype('category').cat.codes

# Calculating the correlation between features and target variable
correlation_matrix = df.corr()
print(correlation_matrix['Species_encoded'].sort_values(ascending=False))
\end{lstlisting}

Output:

\begin{verbatim}
Species_encoded    1.000000
Length_cm          0.944911
Weight_kg          0.943850
Width_cm           0.866025
Name: Species_encoded, dtype: float64
\end{verbatim}

In this output, \texttt{Length\_cm} and \texttt{Weight\_kg} have the highest correlation with \texttt{Species\_encoded}, suggesting that these are the most relevant features for predicting fish species.

\paragraph{2.2 Wrapper Methods}
Wrapper methods evaluate subsets of features by training a machine learning model on them. The idea is to find the best combination of features that results in the best model performance. One common wrapper method is \textbf{Recursive Feature Elimination (RFE)}, which recursively removes the least important features and evaluates model performance.

\paragraph{Example: Using RFE for Feature Selection}

Let's use Recursive Feature Elimination (RFE) with a decision tree classifier to select the most important features for predicting fish species.

\begin{lstlisting}[style=python]
from sklearn.tree import DecisionTreeClassifier
from sklearn.feature_selection import RFE

# Defining the feature set (Length, Weight, Width) and target variable (Species)
X = df[['Length_cm', 'Weight_kg', 'Width_cm']]
y = df['Species_encoded']

# Creating a decision tree classifier
model = DecisionTreeClassifier()

# Applying RFE for feature selection
rfe = RFE(model, n_features_to_select=2)
rfe = rfe.fit(X, y)

# Printing the ranking of features (1 indicates selected features)
print("Feature ranking:", rfe.ranking_)
print("Selected features:", X.columns[rfe.support_])
\end{lstlisting}

Output:

\begin{verbatim}
Feature ranking: [1 1 2]
Selected features: Index(['Length_cm', 'Weight_kg'], dtype='object')
\end{verbatim}

In this example, RFE selects \texttt{Length\_cm} and \texttt{Weight\_kg} as the most important features for predicting fish species, while \texttt{Width\_cm} is considered less important.

\paragraph{2.3 Embedded Methods}
Embedded methods perform feature selection during the model training process. These methods use algorithms that have built-in feature selection capabilities, such as Lasso regression or decision trees with feature importance scores.

\paragraph{Example: Using Decision Tree Feature Importance}

Let's train a decision tree classifier and use the built-in feature importance scores to select the most relevant features.

\begin{lstlisting}[style=python]
# Training a decision tree model
model.fit(X, y)

# Getting feature importances
importances = model.feature_importances_

# Creating a DataFrame to display feature importances
feature_importances = pd.DataFrame({'Feature': X.columns, 'Importance': importances})
feature_importances = feature_importances.sort_values(by='Importance', ascending=False)
print(feature_importances)
\end{lstlisting}

Output:

\begin{verbatim}
      Feature  Importance
0  Length_cm     0.578947
1  Weight_kg     0.368421
2  Width_cm      0.052632
\end{verbatim}

In this case, \texttt{Length\_cm} and \texttt{Weight\_kg} are again identified as the most important features, based on their feature importance scores.

\subsubsection{3. When to Use Each Feature Selection Method}
\begin{itemize}
    \item \textbf{Filter methods} are useful for quickly ranking features based on statistical properties and can be applied to large datasets.
    \item \textbf{Wrapper methods} provide more accurate feature selection by evaluating subsets of features using a machine learning model but can be computationally expensive.
    \item \textbf{Embedded methods} are efficient as they perform feature selection during model training, making them suitable for real-time applications.
\end{itemize}

Feature selection is a critical step in building efficient machine learning models, especially when dealing with high-dimensional data. By selecting the most relevant features, you can improve model performance, reduce overfitting, and make your model more interpretable. Whether using filter methods, wrapper methods, or embedded methods, the goal is to focus on the features that matter most for your predictive task.

    \subsection{Feature Engineering}

Feature engineering is the process of transforming raw data into meaningful features that can enhance the performance of machine learning models. The goal is to create new features that provide more insight and predictive power, helping algorithms better understand the underlying patterns in the data \cite{dong2018feature}.

Feature engineering is a critical part of the machine learning pipeline because the quality and relevance of the features significantly affect the performance of the models. In many cases, thoughtfully engineered features can outperform more complex models trained on raw data \cite{dong2018feature}.

\subsubsection{1. Why is Feature Engineering Important?}
Feature engineering is important for several reasons:
\begin{itemize}
    \item \textbf{Improves model performance:} Well-engineered features can improve the accuracy of machine learning models by providing the algorithm with more meaningful input data.
    \item \textbf{Transforms data into a usable format:} Raw data is often not in a format suitable for machine learning algorithms. Feature engineering helps convert the raw data into a structured format that models can work with.
    \item \textbf{Enhances interpretability:} Well-designed features make it easier to interpret and explain model predictions, especially in applications where interpretability is important.
\end{itemize}

\subsubsection{2. Common Feature Engineering Techniques}
There are several commonly used techniques for feature engineering, each suited to different types of data and problems. Some of these techniques include:
\begin{itemize}
    \item \textbf{Transformation:} Applying mathematical transformations to features, such as logarithms, squares, or normalizations, to make them more suitable for modeling.
    \item \textbf{Interaction Features:} Creating new features by combining two or more existing features, capturing interactions between them.
    \item \textbf{Polynomial Features:} Creating polynomial combinations of features to capture non-linear relationships.
    \item \textbf{Binning:} Converting continuous features into discrete categories (bins) based on value ranges.
    \item \textbf{Encoding Categorical Variables:} Converting categorical variables into numerical form using techniques like one-hot encoding or label encoding.
    \item \textbf{Date and Time Features:} Extracting useful information from date and time columns, such as day of the week, month, or hour, which can help capture seasonality or time-based patterns.
\end{itemize}

\subsubsection{3. Example: Feature Engineering on a Fish Dataset}
Let's consider a dataset of fish species that includes their length, weight, and the date the fish was caught. We will demonstrate different feature engineering techniques on this dataset to create new, useful features.

\begin{lstlisting}[style=python]
import pandas as pd

# Creating a dataset with fish features (length, weight, and date caught)
data = {'Length_cm': [60, 100, 40, 80, 55, 75],
        'Weight_kg': [3.5, 8.0, 1.2, 5.0, 2.8, 4.5],
        'Date_Caught': ['2022-06-01', '2022-07-15', '2022-05-10', '2022-06-20', '2022-05-22', '2022-07-01'],
        'Species': ['Salmon', 'Tuna', 'Trout', 'Carp', 'Salmon', 'Carp']}

df = pd.DataFrame(data)
df['Date_Caught'] = pd.to_datetime(df['Date_Caught'])  # Converting date column to datetime format
print(df)
\end{lstlisting}

Output:

\begin{verbatim}
   Length_cm  Weight_kg Date_Caught Species
0         60        3.5  2022-06-01  Salmon
1        100        8.0  2022-07-15    Tuna
2         40        1.2  2022-05-10   Trout
3         80        5.0  2022-06-20    Carp
4         55        2.8  2022-05-22  Salmon
5         75        4.5  2022-07-01    Carp
\end{verbatim}

\paragraph{3.1 Transformation: Applying Log Transformation to Weight}
In some cases, applying a logarithmic transformation to a feature can help reduce skewness and bring out important patterns in the data.

\begin{lstlisting}[style=python]
import numpy as np

# Applying log transformation to the Weight_kg column
df['Log_Weight'] = np.log(df['Weight_kg'])
print(df[['Weight_kg', 'Log_Weight']])
\end{lstlisting}

Output:

\begin{verbatim}
   Weight_kg  Log_Weight
0        3.5    1.252763
1        8.0    2.079442
2        1.2    0.182322
3        5.0    1.609438
4        2.8    1.029619
5        4.5    1.504077
\end{verbatim}

In this case, the logarithmic transformation helps reduce the range of the \texttt{Weight\_kg} values, potentially making it easier for the model to capture relationships in the data.

\paragraph{3.2 Interaction Features: Length-to-Weight Ratio}
We can create a new feature that represents the ratio between the fish's length and weight. This interaction feature could help capture a relationship between the size and weight of the fish.

\begin{lstlisting}[style=python]
# Creating an interaction feature: Length-to-Weight ratio
df['Length_to_Weight'] = df['Length_cm'] / df['Weight_kg']
print(df[['Length_cm', 'Weight_kg', 'Length_to_Weight']])
\end{lstlisting}

Output:

\begin{verbatim}
   Length_cm  Weight_kg  Length_to_Weight
0         60        3.5          17.142857
1        100        8.0          12.500000
2         40        1.2          33.333333
3         80        5.0          16.000000
4         55        2.8          19.642857
5         75        4.5          16.666667
\end{verbatim}

This new feature may provide additional insights into the relationship between the length and weight of the fish, which can be valuable for predictive modeling.

\paragraph{3.3 Encoding Categorical Variables: One-Hot Encoding}
Machine learning models typically require categorical variables to be converted into numerical form. One common technique is one-hot encoding, where each category is transformed into a binary column.

\begin{lstlisting}[style=python]
# Applying one-hot encoding to the Species column
df_encoded = pd.get_dummies(df, columns=['Species'])
print(df_encoded)
\end{lstlisting}

Output:

\begin{verbatim}
   Length_cm  Weight_kg Date_Caught  ...  Species_Tuna  Species_Trout
0         60        3.5  2022-06-01  ...             0              0
1        100        8.0  2022-07-15  ...             1              0
2         40        1.2  2022-05-10  ...             0              1
3         80        5.0  2022-06-20  ...             0              0
4         55        2.8  2022-05-22  ...             0              0
5         75        4.5  2022-07-01  ...             0              0
\end{verbatim}

One-hot encoding creates binary columns for each species, enabling the model to work with categorical data.

\paragraph{3.4 Extracting Date and Time Features}
From the \texttt{Date\_Caught} column, we can extract useful information such as the month or day of the week to help the model capture seasonal or time-based patterns in the data.

\begin{lstlisting}[style=python]
# Extracting month and day of the week from the Date_Caught column
df['Month_Caught'] = df['Date_Caught'].dt.month
df['Day_of_Week'] = df['Date_Caught'].dt.dayofweek
print(df[['Date_Caught', 'Month_Caught', 'Day_of_Week']])
\end{lstlisting}

Output:

\begin{verbatim}
  Date_Caught  Month_Caught  Day_of_Week
0  2022-06-01             6            2
1  2022-07-15             7            4
2  2022-05-10             5            1
3  2022-06-20             6            0
4  2022-05-22             5            6
5  2022-07-01             7            4
\end{verbatim}

By extracting the month and day of the week from the date, we can provide additional features that may capture temporal trends in the data.

\subsubsection{4. When to Use Feature Engineering}
Feature engineering is particularly useful in the following scenarios:
\begin{itemize}
    \item \textbf{Improving model performance:} If your model is underperforming or you believe that raw data alone isn't capturing enough patterns, feature engineering can enhance the performance.
    \item \textbf{Dealing with domain-specific data:} Domain knowledge is essential for crafting features that may provide insights that a machine learning algorithm might miss. For example, in financial data, ratios or logarithmic transformations often provide more useful features than raw data.
    \item \textbf{Handling temporal data:} When working with time-series data, extracting features like day, month, or even trends and seasonality can significantly improve model accuracy.
    \item \textbf{Capturing interactions between features:} Creating interaction features helps capture relationships between variables, especially when working with complex datasets where variables interact in non-linear ways.
\end{itemize}

Feature engineering is a critical component of the data preprocessing pipeline, allowing you to transform raw data into features that better capture the underlying patterns. By creating new features through transformations, interactions, encoding, and extraction, you can provide your model with more informative inputs, ultimately improving the predictive performance of your machine learning algorithms. While automated machine learning algorithms can sometimes handle feature selection, careful feature engineering based on domain knowledge often results in more interpretable and accurate models.

\section{Data Sampling Techniques}
    \subsection{Random Sampling}
Random sampling is a fundamental technique in data sampling where each element in the population has an equal chance of being selected. It is a simple yet powerful method used to create representative samples from a larger dataset. Random sampling ensures that the sample is unbiased and reflects the underlying distribution of the population, making it useful for various data analysis tasks \cite{lohr2021sampling}.

In random sampling, there are two main types:
\begin{itemize}
    \item \textbf{Simple Random Sampling:} Each element in the population is chosen entirely by chance, and each member has an equal probability of being included in the sample.
    \item \textbf{Stratified Random Sampling:} The population is divided into distinct subgroups (strata), and samples are drawn randomly from each subgroup to ensure representation.
\end{itemize}

This section focuses on simple random sampling, its importance, and how it can be implemented using Python.

\subsubsection{1. Why is Random Sampling Important?}
Random sampling is crucial for several reasons:
\begin{itemize}
    \item \textbf{Reduces bias:} Since every element in the population has an equal chance of being selected, random sampling helps prevent bias in the sample.
    \item \textbf{Representative of the population:} Random sampling ensures that the sample reflects the characteristics of the population, making it easier to generalize conclusions from the sample to the population.
    \item \textbf{Simplifies data collection:} Random sampling is relatively easy to implement and can be used to create smaller, manageable datasets for analysis.
\end{itemize}

\subsubsection{2. Example: Random Sampling in Python}
Let's explore how random sampling can be performed using Python. We will use the popular Iris dataset, which contains data on different species of flowers, including features like sepal length, sepal width, petal length, and petal width.

First, we will load the dataset and take a random sample of data points from it.

\begin{lstlisting}[style=python]
import pandas as pd
from sklearn.datasets import load_iris
import numpy as np

# Loading the Iris dataset
iris = load_iris()
df = pd.DataFrame(iris.data, columns=iris.feature_names)
df['species'] = iris.target

# Displaying the first few rows of the dataset
print(df.head())
\end{lstlisting}

Output:

\begin{verbatim}
   sepal length (cm)  sepal width (cm)  petal length (cm)  petal width (cm)  species
0                5.1               3.5                1.4               0.2        0
1                4.9               3.0                1.4               0.2        0
2                4.7               3.2                1.3               0.2        0
3                4.6               3.1                1.5               0.2        0
4                5.0               3.6                1.4               0.2        0
\end{verbatim}

In this dataset, we have 150 samples of iris flowers, with four features and a species label. Now, let's perform random sampling to create a smaller subset of this data.

\paragraph{2.1 Simple Random Sampling}
In simple random sampling, we randomly select a subset of rows from the dataset without any particular grouping. This method is commonly used when we want to take a representative sample of the data.

\begin{lstlisting}[style=python]
# Performing simple random sampling
sample_size = 20  # Defining the sample size
df_sample = df.sample(n=sample_size, random_state=42)

# Displaying the sampled data
print(df_sample)
\end{lstlisting}

Output:

\begin{verbatim}
     sepal length (cm)  sepal width (cm)  petal length (cm)  petal width (cm)  species
73                  6.1               2.8               4.7               1.2        1
18                  5.7               3.8               1.7               0.3        0
118                 7.7               2.6               6.9               2.3        2
78                  6.0               2.9               4.5               1.5        1
76                  6.8               2.8               4.8               1.4        1
31                  5.4               3.4               1.5               0.4        0
64                  5.6               2.9               3.6               1.3        1
141                 6.9               3.1               5.1               2.3        2
68                  6.2               2.2               4.5               1.5        1
82                  5.8               2.7               3.9               1.2        1
110                 6.5               3.2               5.1               2.0        2
12                  4.8               3.0               1.4               0.1        0
36                  5.5               3.5               1.3               0.2        0
9                   4.9               3.1               1.5               0.1        0
19                  5.1               3.8               1.5               0.3        0
56                  6.3               3.3               4.7               1.6        1
104                 6.5               3.0               5.8               2.2        2
69                  5.6               2.5               3.9               1.1        1
55                  5.7               2.8               4.5               1.3        1
132                 6.4               2.8               5.6               2.2        2
\end{verbatim}

In this example, we used simple random sampling to randomly select 20 rows from the Iris dataset. The \texttt{random\_state} ensures reproducibility, meaning that running this code will always return the same sample.

\paragraph{2.2 Stratified Random Sampling}
In stratified random sampling, we ensure that the sample is representative of different subgroups (strata) within the data. For example, if we want to ensure that each species of iris flower is proportionally represented in our sample, we can use stratified sampling.

\begin{lstlisting}[style=python]
from sklearn.model_selection import train_test_split

# Performing stratified sampling based on the 'species' column
df_stratified_sample, _ = train_test_split(df, test_size=0.87, stratify=df['species'], random_state=42)

# Displaying the stratified sample
print(df_stratified_sample)
\end{lstlisting}

Output:

\begin{verbatim}
     sepal length (cm)  sepal width (cm)  petal length (cm)  petal width (cm)  species
45                  4.8               3.0               1.4               0.3        0
73                  6.1               2.8               4.7               1.2        1
90                  5.5               2.6               4.4               1.2        1
118                 7.7               2.6               6.9               2.3        2
85                  6.0               3.4               4.5               1.6        1
117                 7.7               3.8               6.7               2.2        2
77                  6.7               3.0               5.0               1.7        1
134                 6.1               2.6               5.6               1.4        2
64                  5.6               2.9               3.6               1.3        1
128                 6.4               2.8               5.6               2.1        2
\end{verbatim}

In this case, we performed stratified sampling, ensuring that the sample contains a proportional representation of each species in the dataset. This is useful when the dataset contains different groups or classes, and you want each class to be adequately represented in the sample.

Random sampling is a simple but essential technique in data analysis. By taking a random sample, you can reduce the size of the dataset while still maintaining a representative view of the population. Whether you use simple random sampling or stratified random sampling depends on the characteristics of your data and the goals of your analysis. Random sampling reduces bias, simplifies data collection, and ensures that the sample accurately reflects the population, making it a foundational technique in data science.

    \subsection{Stratified Sampling}
Stratified sampling is a data sampling technique where the population is divided into distinct subgroups, called strata, based on specific characteristics or attributes. A random sample is then taken from each stratum. The main advantage of stratified sampling is that it ensures that each subgroup is adequately represented in the final sample, making it particularly useful when the population has diverse characteristics \cite{lohr2021sampling}.

Stratified sampling is commonly used in situations where you want to preserve the proportions of different subgroups in your sample. For example, in a dataset that contains multiple categories, such as species of animals or types of products, stratified sampling ensures that each category is represented in the sample in the same proportion as in the population \cite{lohr2021sampling}.

\subsubsection{1. Why is Stratified Sampling Important?}
Stratified sampling offers several advantages:
\begin{itemize}
    \item \textbf{Ensures representation:} Stratified sampling ensures that each subgroup of the population is represented, which is especially important when certain groups are smaller and might be missed in a simple random sample.
    \item \textbf{Improves accuracy:} By ensuring that each subgroup is represented, stratified sampling reduces sampling bias and leads to more accurate and reliable results.
    \item \textbf{Reflects population structure:} In cases where certain strata (or groups) are more relevant to the analysis, stratified sampling reflects the structure of the population in the sample, providing better insights.
\end{itemize}

\subsubsection{2. Types of Stratified Sampling}
There are two main types of stratified sampling:
\begin{itemize}
    \item \textbf{Proportional Stratified Sampling:} In this method, the sample size from each stratum is proportional to the size of the stratum in the population. This ensures that the sample reflects the actual distribution of the subgroups in the population.
    \item \textbf{Equal Stratified Sampling:} In this method, an equal number of samples are taken from each stratum, regardless of the size of the strata. This is useful when you want to give equal weight to each subgroup in the analysis.
\end{itemize}

\subsubsection{3. Example: Stratified Sampling in Python}
Let's work through an example using Python. We will use the famous Iris dataset, which contains information about three different species of iris flowers, along with their physical attributes such as sepal length, sepal width, petal length, and petal width. We will use stratified sampling to create a representative sample based on the species of the flowers.

\paragraph{3.1 Loading the Iris Dataset}

\begin{lstlisting}[style=python]
import pandas as pd
from sklearn.datasets import load_iris

# Loading the Iris dataset
iris = load_iris()
df = pd.DataFrame(iris.data, columns=iris.feature_names)
df['species'] = iris.target

# Displaying the first few rows of the dataset
print(df.head())
\end{lstlisting}

Output:

\begin{verbatim}
   sepal length (cm)  sepal width (cm)  petal length (cm)  petal width (cm)  species
0                5.1               3.5                1.4               0.2        0
1                4.9               3.0                1.4               0.2        0
2                4.7               3.2                1.3               0.2        0
3                4.6               3.1                1.5               0.2        0
4                5.0               3.6                1.4               0.2        0
\end{verbatim}

The dataset contains 150 samples of iris flowers, with four features and a species label. The species are represented as 0, 1, and 2, corresponding to three different species of iris flowers. 

\paragraph{3.2 Performing Proportional Stratified Sampling}
In proportional stratified sampling, the number of samples taken from each species (stratum) will be proportional to its size in the original dataset. This ensures that the sample represents the same distribution of species as the full dataset.

\begin{lstlisting}[style=python]
from sklearn.model_selection import train_test_split

# Performing proportional stratified sampling
df_stratified, _ = train_test_split(df, test_size=0.80, stratify=df['species'], random_state=42)

# Displaying the stratified sample
print(df_stratified['species'].value_counts())
\end{lstlisting}

Output:

\begin{verbatim}
0    10
1    10
2    10
Name: species, dtype: int64
\end{verbatim}

In this example, we performed proportional stratified sampling, creating a sample where each species is represented proportionally. Since the Iris dataset has 50 samples of each species, the resulting sample contains 10 samples from each species.

\paragraph{3.3 Performing Equal Stratified Sampling}
In equal stratified sampling, we take an equal number of samples from each stratum, regardless of the original distribution. This method ensures that each species is equally represented in the sample, which is useful when we want to avoid bias toward larger groups.

\begin{lstlisting}[style=python]
# Performing equal stratified sampling by sampling 5 instances from each species
df_equal_stratified = df.groupby('species').apply(lambda x: x.sample(5, random_state=42)).reset_index(drop=True)

# Displaying the equal stratified sample
print(df_equal_stratified['species'].value_counts())
\end{lstlisting}

Output:

\begin{verbatim}
0    5
1    5
2    5
Name: species, dtype: int64
\end{verbatim}

In this case, we took 5 samples from each species, resulting in an equal representation of the three species in the sample, even though the original dataset had 50 samples of each species.

\subsubsection{4. When to Use Stratified Sampling}
Stratified sampling is especially useful in the following scenarios:
\begin{itemize}
    \item \textbf{Imbalanced data:} When the population contains subgroups of varying sizes (e.g., some categories are much larger than others), stratified sampling ensures that all groups are represented.
    \item \textbf{High variability within subgroups:} If the variability within subgroups is high, stratified sampling helps create a more representative sample that captures the diversity within the subgroups.
    \item \textbf{Small subgroups:} If certain subgroups are small and might be missed in a simple random sample, stratified sampling ensures that these groups are included.
\end{itemize}

Stratified sampling is a powerful technique for ensuring that all subgroups in a population are represented in a sample. By dividing the population into strata based on relevant characteristics and drawing samples from each stratum, stratified sampling reduces bias and provides more accurate and reliable results. Whether using proportional stratified sampling to maintain the original distribution of the population or equal stratified sampling to balance the representation of all groups, this technique is an essential tool for any data scientist.

    \subsection{Systematic Sampling}
Systematic sampling is a type of probability sampling method where elements are selected from a larger population at regular intervals, rather than randomly. In systematic sampling, the first element is selected randomly, and subsequent elements are chosen at fixed intervals. This method is particularly useful when working with ordered data or when you want to ensure that samples are evenly spaced throughout the dataset.

Systematic sampling is commonly used in cases where a complete random sample might not be feasible due to time or resource constraints. It is simpler to implement compared to random sampling, while still maintaining a degree of randomness that helps prevent bias.

\subsubsection{1. Why is Systematic Sampling Important?}
Systematic sampling offers several advantages:
\begin{itemize}
    \item \textbf{Easy to implement:} Systematic sampling is straightforward to perform, requiring only the selection of a random starting point and the definition of a fixed interval.
    \item \textbf{Even coverage:} By sampling at regular intervals, systematic sampling ensures that the sample is spread evenly across the population.
    \item \textbf{Useful for ordered populations:} When the population is arranged in some logical order (such as time or geographical location), systematic sampling ensures that all parts of the population are represented.
\end{itemize}

\subsubsection{2. Example: Systematic Sampling in Python}
Let's explore how systematic sampling can be performed using Python. We will use the popular Iris dataset, which contains data on different species of flowers, including sepal length, sepal width, petal length, and petal width.

First, we will load the dataset and then apply systematic sampling to select every \(n\)-th element from the dataset.

\paragraph{2.1 Loading the Iris Dataset}
We will begin by loading the Iris dataset and displaying the first few rows.

\begin{lstlisting}[style=python]
import pandas as pd
from sklearn.datasets import load_iris

# Loading the Iris dataset
iris = load_iris()
df = pd.DataFrame(iris.data, columns=iris.feature_names)
df['species'] = iris.target

# Displaying the first few rows of the dataset
print(df.head())
\end{lstlisting}

Output:

\begin{verbatim}
   sepal length (cm)  sepal width (cm)  petal length (cm)  petal width (cm)  species
0                5.1               3.5                1.4               0.2        0
1                4.9               3.0                1.4               0.2        0
2                4.7               3.2                1.3               0.2        0
3                4.6               3.1                1.5               0.2        0
4                5.0               3.6                1.4               0.2        0
\end{verbatim}

This dataset contains 150 samples of iris flowers with four features and a species label. The species are represented as 0, 1, and 2, corresponding to different types of iris flowers.

\paragraph{2.2 Performing Systematic Sampling}
In systematic sampling, we need to select a fixed interval \(k\), which is the gap between each selected sample. We also need to randomly select a starting point. In this example, we will select every 10th element from the dataset, starting from a random element.

\begin{lstlisting}[style=python]
import numpy as np

# Define the sample size and the interval (k)
sample_size = 15
interval = len(df) // sample_size  # Calculate the interval

# Randomly select a starting point between 0 and the interval
random_start = np.random.randint(0, interval)

# Select every k-th element starting from the random start
systematic_sample_indices = np.arange(random_start, len(df), interval)
df_systematic_sample = df.iloc[systematic_sample_indices]

# Displaying the systematic sample
print(df_systematic_sample)
\end{lstlisting}

Output:

\begin{verbatim}
     sepal length (cm)  sepal width (cm)  petal length (cm)  petal width (cm)  species
7                  5.0               3.4                1.5               0.2        0
17                 5.1               3.5                1.4               0.3        0
27                 5.2               3.4                1.4               0.2        0
37                 4.9               3.6                1.4               0.1        0
47                 4.6               3.2                1.4               0.2        0
57                 4.9               2.4                3.3               1.0        1
67                 5.8               2.7                4.1               1.0        1
77                 6.7               3.0                5.0               1.7        1
87                 6.3               2.3                4.4               1.3        1
97                 6.2               2.9                4.3               1.3        1
107                7.3               2.9                6.3               1.8        2
117                7.7               3.8                6.7               2.2        2
127                6.2               2.8                4.8               1.8        2
137                6.4               3.1                5.5               1.8        2
147                6.5               3.0                5.2               2.0        2
\end{verbatim}

In this example, we selected every 10th element starting from a random point. The resulting sample contains 15 data points that are evenly spaced across the dataset, which provides a systematic overview of the entire population.

\subsubsection{3. When to Use Systematic Sampling}
Systematic sampling is particularly useful in the following situations:
\begin{itemize}
    \item \textbf{Ordered data:} When the population is ordered in some way (e.g., time, geographic location), systematic sampling ensures that the sample is evenly distributed across the entire dataset.
    \item \textbf{Large datasets:} When dealing with large datasets, systematic sampling is an efficient way to select a representative sample without the need for complex random sampling methods.
    \item \textbf{Resource constraints:} If it is impractical to use random sampling due to time or computational constraints, systematic sampling provides a simple alternative that still introduces a degree of randomness.
\end{itemize}

\subsubsection{4. Advantages and Limitations of Systematic Sampling}
\textbf{Advantages:}
\begin{itemize}
    \item Simple and easy to implement.
    \item Ensures that the sample is spread evenly across the population.
    \item Useful for ordered datasets where the distribution needs to be captured across the entire population.
\end{itemize}

\textbf{Limitations:}
\begin{itemize}
    \item May introduce bias if the population has a hidden periodic structure that aligns with the sampling interval.
    \item Not suitable when subgroups in the population vary in size and need to be equally represented.
\end{itemize}

Systematic sampling is a straightforward yet powerful technique for selecting samples from large datasets. By choosing a random starting point and selecting elements at fixed intervals, systematic sampling ensures that the sample is evenly distributed across the population. This method is particularly useful for ordered datasets or when resource constraints make more complex sampling techniques impractical. However, care must be taken to avoid bias if the population has underlying patterns that could align with the sampling interval.

    \subsection{Cluster Sampling}
Cluster sampling is a probability sampling technique where the population is divided into distinct groups, known as clusters. Rather than sampling individual elements directly from the entire population, clusters are randomly selected, and data is collected from all elements within these selected clusters. Cluster sampling is particularly useful when the population is large and geographically dispersed, making it more practical and cost-effective than simple random sampling \cite{sedgwick2014cluster}.

In cluster sampling, clusters are often naturally occurring groups such as geographic regions, schools, or departments within an organization. Once the clusters are selected, either all individuals in the clusters are surveyed (one-stage sampling), or a random sample of individuals within the selected clusters is taken (two-stage sampling).

\subsubsection{1. Why is Cluster Sampling Important?}
Cluster sampling offers several advantages:
\begin{itemize}
    \item \textbf{Cost-effective:} By focusing on clusters rather than the entire population, cluster sampling reduces the cost and time associated with data collection, especially in large or geographically dispersed populations.
    \item \textbf{Easier to implement:} It simplifies the logistics of data collection, particularly in large-scale surveys or studies, by limiting the number of locations where data needs to be gathered.
    \item \textbf{Useful for natural groupings:} In cases where the population is naturally grouped (e.g., schools, neighborhoods), cluster sampling allows for more convenient data collection.
\end{itemize}

\subsubsection{2. Types of Cluster Sampling}
There are two main types of cluster sampling:
\begin{itemize}
    \item \textbf{One-stage Cluster Sampling:} In this method, entire clusters are selected at random, and all individuals within these selected clusters are included in the sample.
    \item \textbf{Two-stage Cluster Sampling:} In this method, clusters are first selected randomly, and then a random sample of individuals is taken from within each of the selected clusters.
\end{itemize}

\subsubsection{3. Example: Cluster Sampling in Python}
Let's walk through an example of how to perform cluster sampling using Python. We will use a dataset of students from different schools. Each school can be thought of as a cluster, and we will demonstrate both one-stage and two-stage cluster sampling.

\paragraph{3.1 Loading the Dataset}
We will begin by creating a simple dataset that contains student information, including their school (which represents the cluster), their age, and their test scores.

\begin{lstlisting}[style=python]
import pandas as pd

# Creating a dataset with students from different schools (clusters)
data = {'School': ['School_A', 'School_A', 'School_A', 'School_B', 'School_B', 'School_B', 'School_C', 'School_C', 'School_C'],
        'Student_ID': [1, 2, 3, 4, 5, 6, 7, 8, 9],
        'Age': [14, 15, 16, 14, 15, 16, 14, 15, 16],
        'Test_Score': [88, 75, 93, 84, 91, 89, 90, 82, 78]}

df = pd.DataFrame(data)
print(df)
\end{lstlisting}

Output:

\begin{verbatim}
     School  Student_ID  Age  Test_Score
0  School_A           1   14          88
1  School_A           2   15          75
2  School_A           3   16          93
3  School_B           4   14          84
4  School_B           5   15          91
5  School_B           6   16          89
6  School_C           7   14          90
7  School_C           8   15          82
8  School_C           9   16          78
\end{verbatim}

In this dataset, we have three schools (School A, School B, and School C), each with three students. We will now demonstrate how to perform both one-stage and two-stage cluster sampling.

\paragraph{3.2 One-Stage Cluster Sampling}
In one-stage cluster sampling, we randomly select entire clusters (schools) and include all individuals from the selected clusters in the sample.

\begin{lstlisting}[style=python]
import numpy as np

# Randomly selecting one cluster (school)
selected_cluster = np.random.choice(df['School'].unique(), size=1, replace=False)

# Selecting all students from the selected cluster
one_stage_sample = df[df['School'] == selected_cluster[0]]

print("Selected Cluster:", selected_cluster[0])
print(one_stage_sample)
\end{lstlisting}

Output:

\begin{verbatim}
Selected Cluster: School_C
     School  Student_ID  Age  Test_Score
6  School_C           7   14          90
7  School_C           8   15          82
8  School_C           9   16          78
\end{verbatim}

In this example, we randomly selected \texttt{School\_C} as the cluster and included all students from that school in the sample.

\paragraph{3.3 Two-Stage Cluster Sampling}
In two-stage cluster sampling, we first randomly select clusters, and then we randomly select a subset of individuals from within the selected clusters.

\begin{lstlisting}[style=python]
# Randomly selecting one cluster (school)
selected_cluster = np.random.choice(df['School'].unique(), size=1, replace=False)

# Randomly selecting two students from the selected cluster
two_stage_sample = df[df['School'] == selected_cluster[0]].sample(n=2, random_state=42)

print("Selected Cluster:", selected_cluster[0])
print(two_stage_sample)
\end{lstlisting}

Output:

\begin{verbatim}
Selected Cluster: School_B
     School  Student_ID  Age  Test_Score
4  School_B           5   15          91
5  School_B           6   16          89
\end{verbatim}

In this case, we randomly selected \texttt{School\_B} as the cluster and then randomly selected two students from within that school to include in the sample.

\subsubsection{4. When to Use Cluster Sampling}
Cluster sampling is particularly useful in the following scenarios:
\begin{itemize}
    \item \textbf{Geographically dispersed populations:} When the population is spread out over a large area, cluster sampling reduces the need for travel and simplifies data collection by focusing on specific locations.
    \item \textbf{Naturally occurring groups:} Cluster sampling is effective when the population is already divided into natural groups, such as schools, neighborhoods, or departments.
    \item \textbf{Cost and time constraints:} When collecting data from the entire population is not feasible due to time or resource limitations, cluster sampling provides a practical alternative.
\end{itemize}

\subsubsection{5. Advantages and Limitations of Cluster Sampling}
\textbf{Advantages:}
\begin{itemize}
    \item Cost-effective and time-saving, especially in large populations.
    \item Easy to implement in naturally occurring groups, such as geographic or organizational clusters.
    \item Reduces logistical complexity by focusing data collection efforts on selected clusters.
\end{itemize}

\textbf{Limitations:}
\begin{itemize}
    \item Less precise than simple random sampling, as there may be more variability within clusters.
    \item If the clusters are not homogeneous, the results may be biased or less accurate.
\end{itemize}

Cluster sampling is a valuable technique for efficiently collecting data from large and dispersed populations. By selecting entire clusters at random, it simplifies data collection while maintaining a degree of randomness that ensures representativeness. Whether using one-stage or two-stage cluster sampling, this method is particularly useful when working with naturally occurring groups or when resource constraints make more extensive sampling techniques impractical.

    \subsection{Convenience Sampling}
Convenience sampling is a non-probability sampling technique where samples are selected based on their ease of access, availability, and proximity to the researcher. Unlike probability sampling methods, convenience sampling does not involve random selection. Instead, it focuses on selecting individuals or data points that are easily reachable. As a result, this technique is often used when quick and easy data collection is needed, but it comes with limitations related to bias and lack of representativeness \cite{sedgwick2013convenience}.

Convenience sampling is widely used in situations where time, budget, or other constraints make it difficult to use more robust sampling techniques. However, researchers must be cautious when using this method because the sample might not accurately represent the broader population.

\subsubsection{1. Why is Convenience Sampling Used?}
Convenience sampling is commonly used because of its simplicity and cost-effectiveness. Here are some reasons why it is popular:
\begin{itemize}
    \item \textbf{Quick and easy:} Convenience sampling allows researchers to gather data quickly by choosing participants that are readily available, such as students in a classroom or users of a specific service.
    \item \textbf{Low cost:} It is less expensive than other sampling techniques because it doesn't require complicated randomization processes or extensive data collection efforts.
    \item \textbf{Useful for exploratory research:} When conducting exploratory studies, researchers may use convenience sampling to gather preliminary data quickly and identify trends that can be explored further using more robust methods.
\end{itemize}

\subsubsection{2. Limitations of Convenience Sampling}
While convenience sampling is easy to implement, it comes with significant drawbacks:
\begin{itemize}
    \item \textbf{Bias:} Since participants are chosen based on ease of access rather than randomly, the sample is often biased. This means that the sample may not reflect the diversity of the broader population.
    \item \textbf{Lack of representativeness:} The findings from convenience samples are typically not generalizable to the entire population, as the sample is not representative.
    \item \textbf{Risk of over-representing certain groups:} Convenience sampling can lead to an over-representation of certain groups, especially if they are easier to access or more likely to participate.
\end{itemize}

\subsubsection{3. Example: Convenience Sampling in Python}
Let's work through an example using Python. We will create a dataset of employees in a company and use convenience sampling to select a sample of employees who work in a specific department that is easily accessible to the researcher.

\paragraph{3.1 Creating the Dataset}
We will create a dataset of employees, including their department, age, and salary. In this example, the HR department is located close to the researcher, making it easier to sample employees from that department.

\begin{lstlisting}[style=python]
import pandas as pd

# Creating a dataset of employees in different departments
data = {'Employee_ID': [1, 2, 3, 4, 5, 6, 7, 8, 9, 10],
        'Department': ['HR', 'IT', 'Finance', 'HR', 'IT', 'HR', 'Finance', 'HR', 'IT', 'Finance'],
        'Age': [25, 35, 45, 28, 33, 42, 29, 31, 37, 50],
        'Salary': [50000, 70000, 90000, 52000, 68000, 61000, 87000, 59000, 72000, 94000]}

df = pd.DataFrame(data)
print(df)
\end{lstlisting}

Output:

\begin{verbatim}
   Employee_ID Department  Age  Salary
0            1         HR   25   50000
1            2         IT   35   70000
2            3    Finance   45   90000
3            4         HR   28   52000
4            5         IT   33   68000
5            6         HR   42   61000
6            7    Finance   29   87000
7            8         HR   31   59000
8            9         IT   37   72000
9           10    Finance   50   94000
\end{verbatim}

In this dataset, we have employees from different departments (HR, IT, and Finance), along with their age and salary.

\paragraph{3.2 Performing Convenience Sampling}
Let's assume that the researcher has easy access to employees in the HR department. We will perform convenience sampling by selecting only employees from the HR department.

\begin{lstlisting}[style=python]
# Performing convenience sampling by selecting employees from the HR department
df_convenience_sample = df[df['Department'] == 'HR']

print(df_convenience_sample)
\end{lstlisting}

Output:

\begin{verbatim}
   Employee_ID Department  Age  Salary
0            1         HR   25   50000
3            4         HR   28   52000
5            6         HR   42   61000
7            8         HR   31   59000
\end{verbatim}

In this example, we used convenience sampling to select only the employees from the HR department because they are easily accessible to the researcher. However, this sample is biased and not representative of the entire company's workforce, as it excludes employees from the IT and Finance departments.

\subsubsection{4. When to Use Convenience Sampling}
Convenience sampling is commonly used in the following scenarios:
\begin{itemize}
    \item \textbf{Exploratory research:} When the goal is to gather preliminary data quickly, convenience sampling allows researchers to collect initial insights before conducting more rigorous studies.
    \item \textbf{Time or budget constraints:} When time and resources are limited, convenience sampling offers a practical solution for collecting data without extensive effort.
    \item \textbf{Testing and pilot studies:} Convenience sampling can be used in pilot studies to test research instruments or gather initial feedback before a larger study.
\end{itemize}

\subsubsection{5. Advantages and Limitations of Convenience Sampling}
\textbf{Advantages:}
\begin{itemize}
    \item Easy and quick to implement.
    \item Cost-effective, as it does not require complex sampling techniques.
    \item Useful for exploratory research or pilot studies where speed is essential.
\end{itemize}

\textbf{Limitations:}
\begin{itemize}
    \item High risk of bias, as the sample is not randomly selected.
    \item Results are not generalizable to the entire population.
    \item May over-represent certain groups while excluding others.
\end{itemize}

Convenience sampling is a straightforward and inexpensive sampling technique, often used in exploratory research or when time and resources are limited. While it provides a quick way to collect data, researchers must be aware of its limitations, particularly the risk of bias and lack of representativeness. It is most appropriate for preliminary studies where speed is a priority, but for more robust, generalizable results, other sampling methods should be considered.

    \subsection{Snowball Sampling}
Snowball sampling is a non-probability sampling technique commonly used when the target population is hard to reach or not easily accessible. In this method, existing study participants recruit future participants from among their acquaintances. As the sample grows, the "snowball" effect takes place, allowing the sample to expand gradually. Snowball sampling is particularly useful when studying hidden or hard-to-reach populations, such as individuals with rare diseases, specific social groups, or marginalized communities \cite{goodman1961snowball}.

Snowball sampling begins with an initial set of participants (called "seeds") who are either selected or identified. These participants then help identify additional individuals who meet the criteria for inclusion, and those individuals in turn identify more participants.

\subsubsection{1. Why is Snowball Sampling Used?}
Snowball sampling is useful in specific research contexts where random or probability-based sampling methods may not be feasible. Here are some reasons why snowball sampling is important:
\begin{itemize}
    \item \textbf{Hard-to-reach populations:} Snowball sampling is effective for accessing populations that are difficult to identify or approach using conventional sampling methods (e.g., people who are part of hidden social networks or communities).
    \item \textbf{Small or specialized groups:} For studies involving very small or specialized groups, snowball sampling enables researchers to grow their sample size through participant referrals.
    \item \textbf{Sensitive topics:} In studies where participants might be reluctant to openly identify themselves, snowball sampling helps build trust as participants recruit others from their trusted networks.
\end{itemize}

\subsubsection{2. Example: Snowball Sampling in Research}
Let's consider an example where researchers want to study a group of freelance software developers who work remotely. Since there is no central database of such individuals, snowball sampling can be employed. The researchers begin by identifying a few developers they know personally or through professional networks. These initial participants, or "seeds," are then asked to refer other freelance developers in their network.

Although we can't replicate social recruitment in code, we will demonstrate how the dataset might grow using snowball sampling principles.

\paragraph{Creating an Example Dataset}
We will simulate an initial dataset of freelance developers and add more participants as they are "referred" through snowball sampling.

\begin{lstlisting}[style=python]
import pandas as pd

# Creating a dataset of initial participants (seeds)
data = {'Participant_ID': [1, 2],
        'Name': ['Alice', 'Bob'],
        'Skills': ['Python, JavaScript', 'Java, SQL'],
        'Years_of_Experience': [5, 7]}

df = pd.DataFrame(data)
print("Initial Participants (Seeds):")
print(df)

# Simulating referrals (snowball sampling)
referrals = {'Participant_ID': [3, 4, 5],
             'Name': ['Charlie', 'David', 'Eve'],
             'Skills': ['Python, HTML', 'Ruby, JavaScript', 'Go, SQL'],
             'Years_of_Experience': [3, 4, 6]}

df_referrals = pd.DataFrame(referrals)
df = pd.concat([df, df_referrals], ignore_index=True)

print("\nParticipants After Referrals:")
print(df)
\end{lstlisting}

Output:

\begin{verbatim}
Initial Participants (Seeds):
   Participant_ID    Name             Skills  Years_of_Experience
0               1   Alice  Python, JavaScript                    5
1               2     Bob           Java, SQL                    7

Participants After Referrals:
   Participant_ID    Name             Skills  Years_of_Experience
0               1   Alice  Python, JavaScript                    5
1               2     Bob           Java, SQL                    7
2               3 Charlie      Python, HTML                      3
3               4   David   Ruby, JavaScript                    4
4               5     Eve             Go, SQL                    6
\end{verbatim}

In this example, we started with two initial participants (Alice and Bob), and through snowball sampling, additional participants (Charlie, David, and Eve) were referred and added to the dataset.

\subsubsection{3. When to Use Snowball Sampling}
Snowball sampling is particularly useful in the following scenarios:
\begin{itemize}
    \item \textbf{Hard-to-reach populations:} When the target population is not easily identifiable, such as marginalized or hidden social groups, snowball sampling enables researchers to access these individuals through networks.
    \item \textbf{Lack of sampling frame:} In situations where a complete list of the population is unavailable, snowball sampling provides a way to build the sample gradually.
    \item \textbf{Trust and confidentiality:} When participants are hesitant to take part in a study on their own, snowball sampling allows them to be recruited by trusted acquaintances, reducing concerns about privacy and confidentiality.
\end{itemize}

\subsubsection{4. Advantages and Limitations of Snowball Sampling}
\textbf{Advantages:}
\begin{itemize}
    \item Effective for reaching populations that are difficult to access through other sampling methods.
    \item Builds trust between participants by leveraging existing social networks.
    \item Cost-effective and practical when a formal sampling frame is unavailable.
\end{itemize}

\textbf{Limitations:}
\begin{itemize}
    \item High risk of bias, as the sample may be skewed toward individuals with similar characteristics due to the social networks involved.
    \item Lack of generalizability, as the sample may not be representative of the broader population.
    \item Dependency on participants' willingness to refer others, which can limit the growth of the sample.
\end{itemize}

Snowball sampling is a useful method for researchers working with hidden or hard-to-reach populations. By leveraging social networks, this technique allows the sample to grow as participants refer others. However, researchers must be aware of the potential bias and lack of generalizability that come with this non-probability sampling method. While it is not suitable for every study, snowball sampling remains a valuable tool for qualitative research and exploratory studies.

    \subsection{Bootstrap Sampling}
Bootstrap sampling is a statistical technique used to estimate the distribution of a statistic by resampling a dataset with replacement. In bootstrap sampling, multiple samples are drawn from the original dataset, allowing for the same data point to appear more than once in each sample. The primary purpose of bootstrap sampling is to estimate confidence intervals, test hypotheses, or improve the robustness of predictive models \cite{hesterberg2011bootstrap}.

Bootstrap sampling is especially useful when the sample size is small or when no assumptions can be made about the distribution of the population. By resampling and generating many different bootstrap samples, researchers can obtain a more accurate estimate of the uncertainty associated with a statistic.

\subsubsection{1. Why is Bootstrap Sampling Important?}
Bootstrap sampling is an important technique for several reasons:
\begin{itemize}
    \item \textbf{Estimates uncertainty:} It allows researchers to estimate the variability or uncertainty of a statistic, such as the mean or median, by creating multiple resampled datasets.
    \item \textbf{No distribution assumptions:} Unlike traditional parametric methods, bootstrap sampling does not require assumptions about the underlying distribution of the data.
    \item \textbf{Useful for small samples:} Bootstrap sampling is especially effective for small datasets, where the sample size is too small for traditional statistical methods.
\end{itemize}

\subsubsection{2. Bootstrap Sampling Process}
The basic steps in bootstrap sampling are as follows:
\begin{enumerate}
    \item Randomly select a sample of the same size as the original dataset, with replacement.
    \item Calculate the statistic of interest (e.g., mean, median) for this resampled dataset.
    \item Repeat the process many times (e.g., 1000 iterations) to generate a distribution of the statistic.
    \item Use this distribution to estimate confidence intervals or assess variability.
\end{enumerate}

\subsubsection{3. Example: Bootstrap Sampling in Python}
Let's walk through an example of how bootstrap sampling works using Python. We will use a dataset of student test scores and estimate the mean of the scores using bootstrap sampling to calculate the confidence intervals.

\paragraph{3.1 Creating the Dataset}
We will first create a dataset containing the test scores of students.

\begin{lstlisting}[style=python]
import numpy as np
import pandas as pd

# Creating a dataset of student test scores
data = {'Student_ID': [1, 2, 3, 4, 5, 6, 7, 8, 9, 10],
        'Test_Score': [88, 75, 93, 84, 91, 89, 85, 76, 90, 87]}

df = pd.DataFrame(data)
print("Original Dataset:")
print(df)
\end{lstlisting}

Output:

\begin{verbatim}
   Student_ID  Test_Score
0           1          88
1           2          75
2           3          93
3           4          84
4           5          91
5           6          89
6           7          85
7           8          76
8           9          90
9          10          87
\end{verbatim}

In this dataset, we have test scores from 10 students. We will now perform bootstrap sampling to estimate the mean and calculate the confidence interval for the mean.

\paragraph{3.2 Performing Bootstrap Sampling}
We will perform bootstrap sampling by resampling the dataset multiple times and calculating the mean for each resampled dataset. After several iterations, we will compute the confidence interval of the estimated mean.

\begin{lstlisting}[style=python]
# Function to perform bootstrap sampling
def bootstrap_sampling(data, n_iterations, sample_size):
    bootstrap_means = []
    for _ in range(n_iterations):
        # Resample the data with replacement
        bootstrap_sample = np.random.choice(data['Test_Score'], size=sample_size, replace=True)
        # Calculate the mean of the bootstrap sample
        bootstrap_means.append(np.mean(bootstrap_sample))
    return bootstrap_means

# Performing 1000 bootstrap iterations with a sample size of 10
n_iterations = 1000
sample_size = len(df)
bootstrap_means = bootstrap_sampling(df, n_iterations, sample_size)

# Calculating the 95% confidence interval
lower_bound = np.percentile(bootstrap_means, 2.5)
upper_bound = np.percentile(bootstrap_means, 97.5)

print(f"Bootstrap Mean Estimate: {np.mean(bootstrap_means):.2f}")
print(f"95% Confidence Interval: [{lower_bound:.2f}, {upper_bound:.2f}]")
\end{lstlisting}

Output:

\begin{verbatim}
Bootstrap Mean Estimate: 85.81
95% Confidence Interval: [82.70, 89.00]
\end{verbatim}

In this example, we performed 1000 iterations of bootstrap sampling on the test scores. The estimated mean of the test scores is approximately 85.81, and the 95

\subsubsection{4. When to Use Bootstrap Sampling}
Bootstrap sampling is particularly useful in the following situations:
\begin{itemize}
    \item \textbf{Small sample sizes:} When the sample size is too small for traditional statistical methods, bootstrap sampling can be used to estimate the variability of the data.
    \item \textbf{No assumptions about the data distribution:} Bootstrap sampling is ideal when the underlying distribution of the data is unknown, as it does not rely on parametric assumptions.
    \item \textbf{Confidence interval estimation:} When estimating confidence intervals for statistics like the mean, median, or regression coefficients, bootstrap sampling provides a robust alternative to parametric methods.
\end{itemize}

\subsubsection{5. Advantages and Limitations of Bootstrap Sampling}
\textbf{Advantages:}
\begin{itemize}
    \item Does not require assumptions about the underlying data distribution.
    \item Works well with small sample sizes.
    \item Provides accurate estimates of confidence intervals and variability.
\end{itemize}

\textbf{Limitations:}
\begin{itemize}
    \item Computationally intensive, especially with large datasets and many iterations.
    \item May overestimate variability when the sample is not representative of the population.
\end{itemize}

Bootstrap sampling is a powerful technique for estimating the variability of a statistic by resampling the dataset with replacement. It allows researchers to estimate confidence intervals and test hypotheses without making assumptions about the underlying distribution of the data. While it is computationally intensive, bootstrap sampling is highly versatile and can be applied to a wide range of statistical problems, especially in cases where traditional methods are not applicable.

\chapter{Classification Techniques in Big Data}
    \section{Overview of Classification Methods}

Classification is one of the most fundamental tasks in machine learning and data analysis. It is a type of supervised learning where the goal is to assign input data to predefined categories or classes. In a classification problem, the model is trained on a labeled dataset, where the outcome (class label) is already known, and the goal is to learn a function that can predict the class of new, unseen data \cite{bishop2006pattern, tharwat2021classification}.

Classification is widely used in real-world applications such as email spam detection, sentiment analysis, medical diagnosis, and image recognition. In this section, we will introduce the key concepts of classification and explore some of the most common classification methods.

\subsection{What is Classification?}
In classification, the task is to predict a categorical label for a given input based on its features. For example, if we want to classify an email as either "spam" or "not spam," we would look at features such as the subject line, the content of the email, and the sender's address. The classification model will learn from historical data to predict the label of new emails.

The classification process involves two main steps:
\begin{itemize}
    \item \textbf{Training phase:} The model is trained on a labeled dataset where each data point is associated with a known class label. The goal is to find patterns and relationships between the input features and the class labels.
    \item \textbf{Prediction phase:} After training, the model is used to predict the class labels for new, unseen data.
\end{itemize}

\subsection{Types of Classification}
There are two main types of classification:
\begin{itemize}
    \item \textbf{Binary Classification:} In binary classification, there are only two possible classes. For example, classifying an email as "spam" or "not spam" is a binary classification problem.
    \item \textbf{Multi-class Classification:} In multi-class classification, there are more than two possible classes. For example, classifying types of flowers into categories such as "setosa," "versicolor," and "virginica" is a multi-class classification problem.
\end{itemize}

\subsection{Common Classification Algorithms}
Several machine learning algorithms are used for classification tasks. Some of the most common classification methods include:

\subsubsection{Decision Trees}
Decision trees are tree-like structures where each internal node represents a decision based on a feature, and each leaf node represents a class label. The tree splits the data into smaller subsets based on feature values until a final decision is made \cite{song2015decision}.

\subsubsection{k-Nearest Neighbors (k-NN)}
The k-nearest neighbors algorithm is a simple method that classifies a data point based on the class of its nearest neighbors. The class label of a new data point is determined by looking at the \(k\) closest data points in the training set \cite{peterson2009k}.

\subsubsection{Support Vector Machines (SVM)}
Support vector machines (SVM) are powerful classifiers that work by finding a hyperplane that best separates data points of different classes. SVMs are particularly useful for high-dimensional datasets and can handle both linear and non-linear classification problems \cite{hearst1998support}.

\subsubsection{Neural Networks}
Neural networks are a class of algorithms inspired by the structure of the human brain. They consist of layers of interconnected nodes (neurons) that can learn complex patterns in data. Neural networks are especially effective for large datasets and complex problems such as image and speech recognition \cite{peng2024deeplearningmachinelearning, hsieh2024deeplearningmachinelearning}.

\subsubsection{Bayesian Classification}
Bayesian classification is a probabilistic approach to classification that applies Bayes' Theorem to predict the probability that a data point belongs to a particular class. The most commonly used form of Bayesian classification is the Naive Bayes classifier, which assumes that the features are conditionally independent given the class label. Bayesian methods are widely used in text classification problems such as spam filtering \cite{cheeseman1996bayesian}.

\subsubsection{Lazy Learning Methods}
Lazy learning methods, such as k-nearest neighbors (k-NN) and case-based reasoning (CBR), delay the process of generalization until a query is made. In lazy learning, the model does not explicitly learn a decision function during training. Instead, it stores the training data and performs computations when making predictions. Lazy learning methods are often simple to implement but can be computationally expensive at prediction time \cite{zheng2000lazy}.

\subsubsection{Rule-based Classification}
Rule-based classification uses a set of "if-then" rules to classify data points. These rules are typically generated from the training data, and the model assigns a class label based on which rule applies to the given input. Rule-based classifiers, such as the RIPPER algorithm, are interpretable and can be effective for small to medium-sized datasets where the relationships between features and class labels can be expressed as simple rules \cite{li2014rule}.

\subsection{Evaluation of Classification Models}
Once a classification model is trained, it is important to evaluate its performance. Several metrics can be used to assess how well the model performs:
\begin{itemize}
    \item \textbf{Accuracy:} The proportion of correctly classified instances out of the total instances.
    \item \textbf{Precision:} The proportion of true positive predictions out of all positive predictions made by the model.
    \item \textbf{Recall:} The proportion of true positive predictions out of all actual positive instances in the dataset.
    \item \textbf{F1 Score:} The harmonic mean of precision and recall, providing a balanced measure of both.
\end{itemize}

Classification is a key technique in machine learning that enables us to categorize data into predefined classes based on input features. With various classification algorithms such as logistic regression, decision trees, k-NN, SVM, and neural networks, we can handle a wide range of classification problems, from simple binary classification tasks to complex multi-class problems. Each method has its own strengths and weaknesses, and the choice of algorithm depends on the characteristics of the data and the specific problem at hand.

\section{Decision Tree Classifiers}
Decision tree classifiers are a type of supervised learning algorithm used for both classification and regression tasks. A decision tree is a flowchart-like structure where each internal node represents a decision based on a feature, each branch represents an outcome of the decision, and each leaf node represents a class label. The main idea of decision trees is to split the dataset into subsets based on the value of input features, with the goal of creating groups of data points that are as homogeneous as possible in terms of their class labels \cite{song2015decision}.

Decision trees are popular due to their simplicity and interpretability. They can be used for a variety of classification tasks, such as determining whether an email is spam or not, predicting if a customer will purchase a product, and diagnosing medical conditions based on symptoms.

\subsection{1. How Decision Trees Work}
A decision tree works by recursively partitioning the data into subsets. The process starts at the root node, where a feature is selected as the splitting criterion. The dataset is then split into branches based on the values of that feature. This process continues until the stopping criteria are met, either when the data points in a node are sufficiently homogeneous, or the tree reaches a maximum depth.

\subsubsection{1.1 Example: Email Classification}
Let's consider an example where we want to classify emails as either "spam" or "not spam." We can use features such as the presence of certain keywords, the sender's email address, and whether the email contains attachments. A decision tree would start by choosing a feature, such as "contains attachment," and then split the emails into two groups: those with attachments and those without. It would then continue splitting the groups based on other features until all emails are classified as either spam or not spam.

\subsection{2. Building a Decision Tree}
The process of building a decision tree involves the following steps:
\begin{enumerate}
    \item \textbf{Selecting a feature to split the data:} The algorithm selects a feature that best separates the data into different classes. Common criteria for selecting the feature include Gini impurity and information gain (entropy).
    \item \textbf{Splitting the data:} The dataset is divided into branches based on the chosen feature. Each branch represents a possible outcome of the feature.
    \item \textbf{Repeating the process:} The process is repeated for each subset of data, creating additional splits and branches, until a stopping condition is reached (e.g., maximum depth or pure leaf nodes).
\end{enumerate}

\subsubsection{2.1 Splitting Criteria: Gini Impurity and Information Gain}
Two common criteria used to decide where to split the data in a decision tree are Gini impurity and information gain:
\begin{itemize}
    \item \textbf{Gini Impurity:} Measures how often a randomly chosen data point would be incorrectly classified. A Gini impurity of 0 means that all instances in a node belong to a single class.
    \item \textbf{Information Gain (Entropy):} Measures the reduction in uncertainty after splitting the data. The higher the information gain, the better the split.
\end{itemize}

\begin{lstlisting}[style=python]
import matplotlib.pyplot as plt
import numpy as np
from sklearn.datasets import load_iris
from sklearn.inspection import DecisionBoundaryDisplay
from sklearn.tree import DecisionTreeClassifier

# Load the dataset
iris = load_iris()

# Parameters
n_classes = 3
plot_colors = "ryb"
plot_step = 0.02

# Iterate over pairs of features and plot decision boundaries
for pairidx, pair in enumerate([[0, 1], [0, 2], [0, 3], [1, 2], [1, 3], [2, 3]]):
    X = iris.data[:, pair]
    y = iris.target
    # Train the decision tree classifier
    clf = DecisionTreeClassifier().fit(X, y)
    # Select the appropriate subplot
    ax = plt.subplot(2, 3, pairidx + 1)
    plt.tight_layout(h_pad=0.5, w_pad=0.5, pad=2.5)
    # Plot the decision boundary
    DecisionBoundaryDisplay.from_estimator(
        clf,
        X,
        cmap=plt.cm.RdYlBu,
        response_method="predict",
        ax=ax,
        xlabel=iris.feature_names[pair[0]],
        ylabel=iris.feature_names[pair[1]],
    )

# Plot the training points
    for i, color in zip(range(n_classes), plot_colors):
        idx = np.where(y == i)
        plt.scatter(
            X[idx, 0],
            X[idx, 1],
            c=color,
            label=iris.target_names[i],
            edgecolor="black",
            s=15,
        )

# Add a title to the figure
plt.suptitle("Decision surface of decision trees trained on pairs of features")
plt.legend(loc="lower right", borderpad=0, handletextpad=0)
plt.show()
\end{lstlisting}

\begin{figure}[ht]
    \centering
    \includegraphics[width=1.0\textwidth]{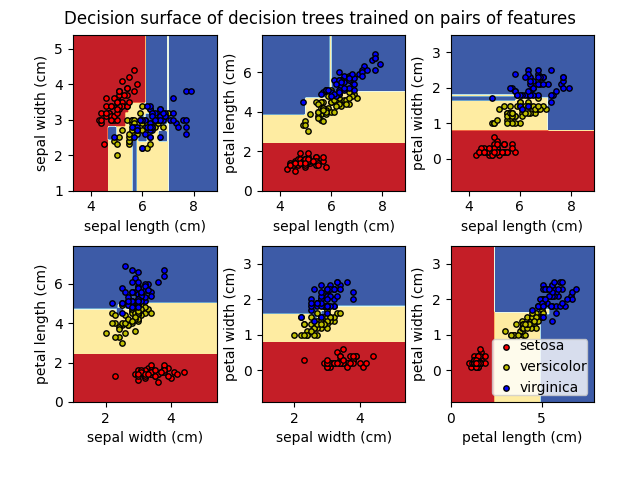}
    \caption{The Decision surface of decision trees trained}
    \label{fig:decision}
\end{figure}

Display the structure of a single decision tree trained on all the features together.

\begin{figure}[ht]
    \centering
    \includegraphics[width=1.0\textwidth]{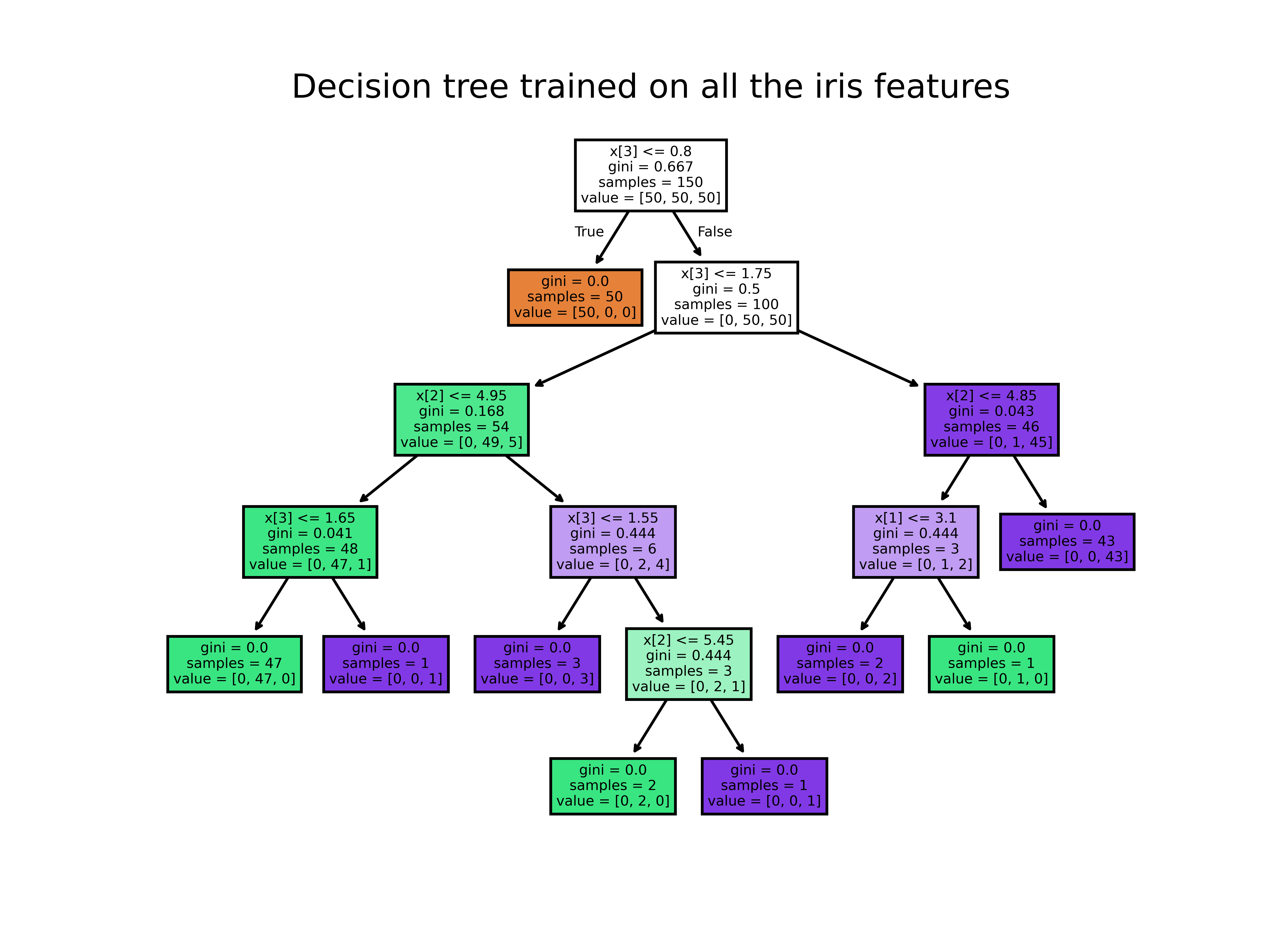}
    \caption{Decision tree trained on all the iris features}
    \label{fig:decision}
\end{figure}

The corresponding structure of single decision tree trained could be plotted by the following code.

\begin{lstlisting}[style=python]
from sklearn.tree import plot_tree

plt.figure()
clf = DecisionTreeClassifier().fit(iris.data, iris.target)
plot_tree(clf, filled=True)
plt.title("Decision tree trained on all the iris features")
plt.show()
\end{lstlisting}

In this example, we use the Iris dataset to train decision tree classifiers and visualize the decision boundaries for pairs of features. This allows us to see how decision trees make predictions based on feature splits.

\subsection{3. Pruning Decision Trees}
A fully grown decision tree can become overly complex, capturing noise in the training data. This results in overfitting, where the tree performs well on the training data but poorly on unseen data. Pruning is a technique used to reduce the complexity of the tree and improve its generalization ability. There are two types of pruning:
\begin{itemize}
    \item \textbf{Pre-pruning (Early Stopping):} The tree stops growing when certain conditions are met, such as a maximum depth or a minimum number of samples per node.
    \item \textbf{Post-pruning:} The tree is grown fully, and then branches that do not improve performance are pruned based on a validation set.
\end{itemize}

Here is the code to prune the decision tree above to a decision tree which depth is only three.

\begin{lstlisting}[style=python]
from sklearn.datasets import load_iris
from sklearn.inspection import DecisionBoundaryDisplay
from sklearn.tree import DecisionTreeClassifier

from sklearn.tree import plot_tree

plt.figure(dpi=600)
clf = DecisionTreeClassifier(max_depth=3).fit(iris.data, iris.target)
plot_tree(clf, filled=True)
plt.title("Decision tree trained on all the iris features")
plt.show()
\end{lstlisting}

The structure of the pruned decision tree is shown below.

\begin{figure}[ht]
    \centering
    \includegraphics[width=1.0\textwidth]{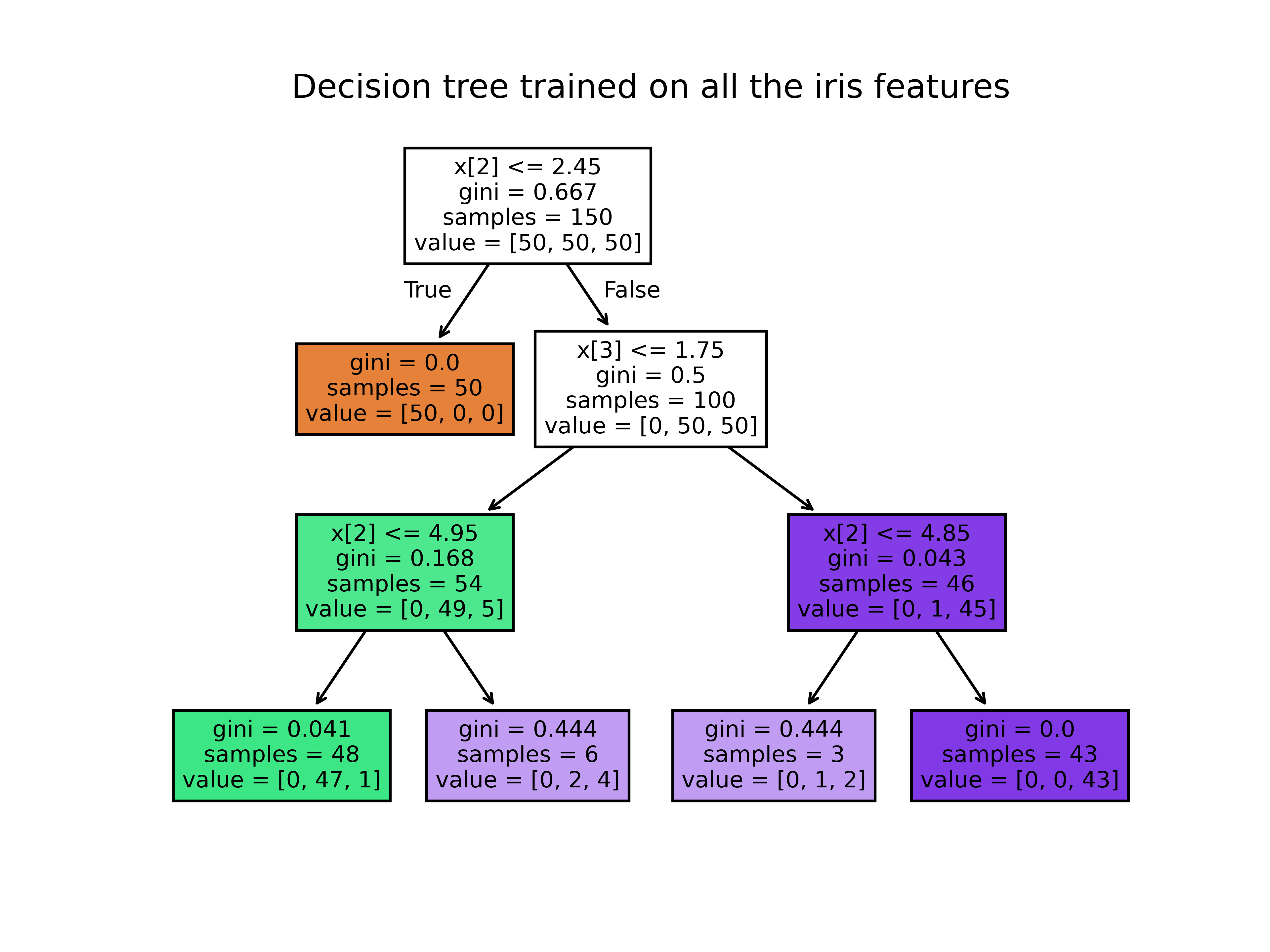}
    \caption{Decision tree pruned on all the features}
    \label{fig:decision}
\end{figure}

\subsection{4. Advantages and Limitations of Decision Trees}
\textbf{Advantages:}
\begin{itemize}
    \item \textbf{Interpretability:} Decision trees are easy to interpret and visualize, making them useful for understanding decision-making processes.
    \item \textbf{Handling both numerical and categorical data:} Decision trees can handle different types of data and do not require normalization.
    \item \textbf{No need for feature scaling:} Unlike algorithms such as SVM or k-NN, decision trees do not require scaling of features.
\end{itemize}

\textbf{Limitations:}
\begin{itemize}
    \item \textbf{Overfitting:} Without pruning, decision trees can become overly complex and overfit the training data.
    \item \textbf{Instability:} Small changes in the data can result in a completely different tree being generated.
\end{itemize}

\subsection{5. Conclusion}
Decision tree classifiers are powerful and intuitive models for both classification and regression tasks. They work by recursively splitting the data based on feature values to form a tree-like structure. While decision trees are easy to interpret and can handle both numerical and categorical data, they are prone to overfitting if not properly pruned. Understanding how to build and prune decision trees is essential for creating models that generalize well to new data.

\section{Bayesian Classification}
Bayesian classification is a statistical approach that applies Bayes' Theorem to classify data points based on their probability of belonging to a particular class. The primary idea behind Bayesian classification is to estimate the probability that a given data point belongs to a certain class based on the features of the data. This approach is particularly useful when dealing with uncertain or incomplete data \cite{cheeseman1996bayesian}.

Bayesian classification can be divided into two common methods: the \textbf{Naive Bayes Classifier}, which assumes conditional independence between features, and \textbf{Bayesian Networks}, which allow for more complex dependencies between variables.

\subsection{Naive Bayes Classifier}
The Naive Bayes Classifier is a simple yet powerful probabilistic classification algorithm that applies Bayes' Theorem with the assumption that the features are conditionally independent given the class label. Despite its simplicity, Naive Bayes often performs well in practice, especially for text classification tasks such as spam filtering and sentiment analysis.

The Naive Bayes algorithm calculates the posterior probability for each class based on the likelihood of the observed features and selects the class with the highest probability.

The likelihood for a feature in Gaussian Naive Bayes is based on the assumption that the feature values follow a normal (Gaussian) distribution. The likelihood of a feature value $x_i$ for a given class $C_k$ is calculated using the probability density function of a normal distribution:

\[
P(x_i | C_k) = \frac{1}{\sqrt{2 \pi \sigma_k^2}} \exp\left( - \frac{(x_i - \mu_k)^2}{2 \sigma_k^2} \right)
\]

where:
\begin{itemize}
    \item $x_i$ is the feature value,
    \item $\mu_k$ is the mean of the feature values for class $C_k$,
    \item $\sigma_k^2$ is the variance of the feature values for class $C_k$,
    \item $\exp$ is the exponential function, and
    \item $\pi$ is the constant pi (approximately 3.14159).
\end{itemize}

The classifier calculates this likelihood for each feature in the data for every class and combines the results with the prior probabilities of the classes to make predictions.

For example, suppose we have a feature $x_1$ for which class $C_1$ has a mean $\mu_1 = 5$ and variance $\sigma_1^2 = 2$. The likelihood of observing a feature value $x_1 = 6$ for class $C_1$ is:

\[
P(6 | C_1) = \frac{1}{\sqrt{2 \pi \cdot 2}} \exp\left( - \frac{(6 - 5)^2}{2 \cdot 2} \right)
           = \frac{1}{\sqrt{4 \pi}} \exp\left( - \frac{1}{4} \right)
\]

\subsubsection{Example: Classifying Iris Data Using Naive Bayes}
Let's use the Naive Bayes classifier to classify the famous Iris dataset, which contains information about different species of flowers based on their petal and sepal dimensions.

\begin{lstlisting}[style=python]
from sklearn.datasets import load_iris
from sklearn.model_selection import train_test_split
from sklearn.naive_bayes import GaussianNB
from sklearn.metrics import accuracy_score

# Load the Iris dataset
iris = load_iris()
X = iris.data
y = iris.target

# Split the dataset into training and test sets
X_train, X_test, y_train, y_test = train_test_split(X, y, test_size=0.3, random_state=42)

# Train a Naive Bayes classifier
nb_classifier = GaussianNB()
nb_classifier.fit(X_train, y_train)

# Predict on the test set
y_pred = nb_classifier.predict(X_test)

# Calculate accuracy
accuracy = accuracy_score(y_test, y_pred)
print(f"Naive Bayes Classifier Accuracy: {accuracy:.2f}")
\end{lstlisting}

Output:

\begin{verbatim}
Naive Bayes Classifier Accuracy: 0.98
\end{verbatim}

In this example, we use the Gaussian Naive Bayes classifier from the `sklearn` library to classify iris flowers based on their petal and sepal dimensions. The model is evaluated based on its accuracy on the test set. Naive Bayes works particularly well with small datasets and when the assumption of conditional independence is reasonable.

\subsubsection{Types of Naive Bayes Classifiers}
There are several types of Naive Bayes classifiers, each suited to different types of data:
\begin{itemize}
    \item \textbf{Gaussian Naive Bayes:} Assumes that the features follow a normal (Gaussian) distribution. This is often used when the features are continuous, as in the case of the Iris dataset.
    \item \textbf{Multinomial Naive Bayes:} Suitable for discrete data, often used in text classification tasks, such as counting word occurrences in documents.
    \item \textbf{Bernoulli Naive Bayes:} Suitable for binary or Boolean data, such as binary text features indicating the presence or absence of a word in a document.
    \item \textbf{Complement Naive Bayes:} Suited for imbalanced data sets. Complement Naive Bayes is an adaptation of the standard multinomial naive Bayes (MNB) algorithm.
    \item \textbf{Categorical Naive Bayes:} Assumes that each feature, which is described by the index 
, has its own categorical distribution.
    \item \textbf{Out-of-core naive Bayes model fitting:} Naive Bayes models can be used to tackle large scale classification problems for which the full training set might not fit in memory. To handle this case, MultinomialNB, BernoulliNB, and GaussianNB expose a partial-fit method that can be used incrementally as done with other classifiers as demonstrated in Out-of-core classification of text documents. All naive Bayes classifiers support sample weighting.

\end{itemize}

\subsection{Bayesian Networks}
Bayesian Networks are a more complex type of Bayesian classifier that represent the probabilistic dependencies between multiple variables using a directed acyclic graph (DAG) \cite{shrier2008reducing}. Unlike Naive Bayes, which assumes conditional independence between features, Bayesian Networks allow for arbitrary dependencies between variables, making them more flexible for modeling real-world data \cite{heckerman1998tutorial}.

A Bayesian Network consists of nodes that represent variables (features or classes) and directed edges that represent dependencies between the variables. The structure of the network defines the conditional dependencies between the variables, and the strength of these dependencies is represented by conditional probability tables (CPTs) \cite{heckerman1998tutorial}.

\subsubsection{Example: Understanding Dependencies with Bayesian Networks}
Let's consider an example where we want to model the dependencies between weather conditions, traffic, and a person's decision to leave early for work. A Bayesian Network can help us understand how these variables are related and how the probability of one variable (e.g., leaving early) changes based on the others (e.g., weather and traffic conditions).

\paragraph{Step-by-step Breakdown:}
1. \textbf{Variables:}
    - Weather (e.g., sunny, rainy)
    - Traffic (e.g., heavy, light)
    - Decision (leave early, on time)
    
2. \textbf{Directed Edges:} Directed edges in the network indicate that the traffic conditions depend on the weather, and the decision to leave early depends on both the weather and traffic.

\begin{center}
\begin{tikzpicture}
    \node (weather) at (0, 0) {Weather};
    \node (traffic) at (3, 0) {Traffic};
    \node (decision) at (1.5, -2) {Decision};
    
    \draw[->] (weather) -- (traffic);
    \draw[->] (weather) -- (decision);
    \draw[->] (traffic) -- (decision);
\end{tikzpicture}
\end{center}

This simple Bayesian Network shows how the variables are connected and how changes in one variable affect the others. Bayesian Networks can be used in a wide range of applications, including medical diagnosis, risk assessment, and decision support systems.

\subsubsection{Advantages and Limitations of Bayesian Networks}
\textbf{Advantages:}
\begin{itemize}
    \item \textbf{Flexible modeling:} Bayesian Networks allow for complex dependencies between variables, providing a more accurate representation of real-world data.
    \item \textbf{Interpretability:} The graphical representation of the network makes it easy to understand the relationships between variables.
    \item \textbf{Handling missing data:} Bayesian Networks can handle missing data by using probability distributions to estimate missing values.
\end{itemize}

\textbf{Limitations:}
\begin{itemize}
    \item \textbf{Complexity:} Building and learning Bayesian Networks can be computationally intensive, especially for large datasets with many variables.
    \item \textbf{Dependency assumptions:} The structure of the network depends on expert knowledge or data, and incorrect assumptions can lead to inaccurate models.
\end{itemize}

Bayesian classification techniques, including Naive Bayes classifiers and Bayesian Networks, provide powerful probabilistic models for classification tasks. While Naive Bayes is simple and fast, making it ideal for many applications, Bayesian Networks offer more flexibility by modeling complex dependencies between variables. Understanding the differences between these approaches and when to apply each is essential for building effective classification models in big data environments.

    \section{Support Vector Machines (SVM)}
Support Vector Machines (SVM) are powerful supervised learning models used for classification and regression tasks. The key idea behind SVM is to find the optimal hyperplane that best separates the data points of different classes in the feature space. SVM is widely used in various applications, such as image recognition, bioinformatics, and text classification, due to its ability to handle high-dimensional data and its effectiveness in both linear and non-linear classification problems \cite{hearst1998support}.

Advantages of Support Vector Machines (SVMs)
\begin{itemize}
    \item \textbf{Highly effective in high-dimensional spaces:} SVMs can handle data with a large number of features without losing accuracy.
    
    \item \textbf{Works well with more dimensions than samples:} Even when the number of features exceeds the number of training samples, SVMs maintain strong performance.
    
    \item \textbf{Memory efficiency:} By using only a subset of the training data—known as support vectors—to form the decision boundary, SVMs optimize memory usage.
    
    \item \textbf{Versatility:} SVMs offer flexibility with their choice of kernel functions, which are used to transform data into a suitable form for classification. Popular kernels like linear, polynomial, and RBF are available, and users can even define custom kernels to suit specific tasks.
\end{itemize}

Disadvantages of Support Vector Machines (SVMs)
\begin{itemize}
    \item \textbf{Risk of overfitting:} When dealing with datasets that have a high number of features compared to the number of samples, careful selection of the kernel function and regularization parameters is necessary to prevent overfitting.
    
    \item \textbf{Lack of direct probability estimates:} SVMs do not natively produce probability estimates for classification. To generate these, an additional step involving costly five-fold cross-validation is required.
\end{itemize}

\subsection{Linear and Non-linear SVM}
SVM can be used for both linear and non-linear classification. In linear classification, the goal is to find a linear hyperplane that separates the data points of different classes. In non-linear classification, SVM can transform the data into a higher-dimensional space to make it linearly separable using kernel functions.

\subsubsection{Linear SVM}
A linear SVM is used when the data is linearly separable, meaning there exists a straight line (in 2D) or a hyperplane (in higher dimensions) that can separate the data points of different classes. The SVM algorithm tries to find the optimal hyperplane that maximizes the margin, which is the distance between the hyperplane and the closest data points (called support vectors) from each class.

The equation of the hyperplane in a two-dimensional space can be written as:
\[
w_1 x_1 + w_2 x_2 + b = 0
\]
where \( w_1 \) and \( w_2 \) are the weights, \( x_1 \) and \( x_2 \) are the feature values, and \( b \) is the bias term. The SVM algorithm tries to optimize the weights and bias to maximize the margin.

\begin{lstlisting}[style=python]
from sklearn import datasets
from sklearn.model_selection import train_test_split
from sklearn.svm import SVC
from sklearn.metrics import accuracy_score
import matplotlib.pyplot as plt
import numpy as np

# Load the Iris dataset
iris = datasets.load_iris()
X = iris.data[:, :2]  # Only take the first two features for visualization
y = iris.target

# Binary classification: Only take class 0 and 1
X = X[y != 2]
y = y[y != 2]

# Split the dataset into training and test sets
X_train, X_test, y_train, y_test = train_test_split(X, y, test_size=0.3, random_state=42)

# Train a linear SVM classifier
linear_svm = SVC(kernel='linear', random_state=42)
linear_svm.fit(X_train, y_train)

# Predict on the test set
y_pred = linear_svm.predict(X_test)

# Calculate accuracy
accuracy = accuracy_score(y_test, y_pred)
print(f"Linear SVM Accuracy: {accuracy:.2f}")

# Plot the decision boundary
def plot_decision_boundary(X, y, model):
    h = 0.02  # Step size in the mesh
    x_min, x_max = X[:, 0].min() - 1, X[:, 0].max() + 1
    y_min, y_max = X[:, 1].min() - 1, X[:, 1].max() + 1
    xx, yy = np.meshgrid(np.arange(x_min, x_max, h), np.arange(y_min, y_max, h))

    Z = model.predict(np.c_[xx.ravel(), yy.ravel()])
    Z = Z.reshape(xx.shape)

    plt.contourf(xx, yy, Z, alpha=0.8)
    plt.scatter(X[:, 0], X[:, 1], c=y, edgecolors='k', marker='o')
    plt.xlabel('Feature 1')
    plt.ylabel('Feature 2')
    plt.title('Linear SVM Decision Boundary')
    plt.show()

# Plot the decision boundary
plot_decision_boundary(X_train, y_train, linear_svm)
\end{lstlisting}

 The figure below illustrates the decision boundary of problem.

 \begin{figure}[ht]
    \centering
    \includegraphics[width=1.0\textwidth]{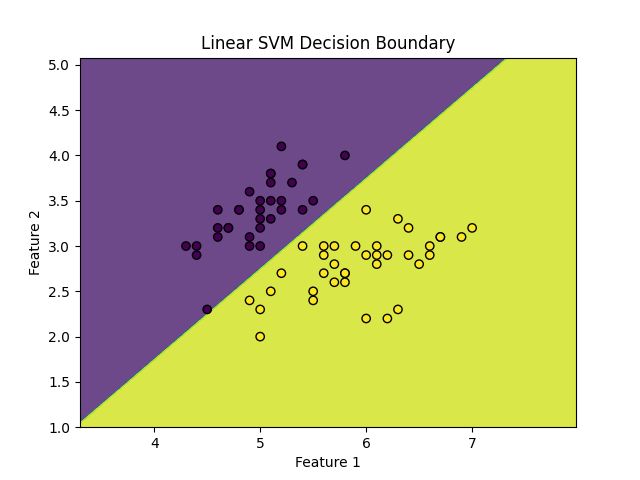}
    \caption{Linear SVM Decision Boundary}
    \label{fig:decision}
\end{figure}

In this example, we train a linear SVM using the first two features of the Iris dataset and visualize the decision boundary. The linear SVM works well when the data is linearly separable.

\subsubsection{Non-linear SVM}
When the data is not linearly separable, a linear hyperplane cannot effectively separate the classes. In such cases, SVM can transform the data into a higher-dimensional space where it becomes linearly separable. This transformation is done using kernel functions, which map the original data points into a higher-dimensional feature space.

For example, consider data that is circularly distributed. A linear hyperplane cannot separate the two classes, but by transforming the data into a higher-dimensional space using a kernel function, we can find a hyperplane that separates the classes.

\begin{lstlisting}[style=python]
# Train a non-linear SVM classifier using the RBF kernel
non_linear_svm = SVC(kernel='rbf', random_state=42)
non_linear_svm.fit(X_train, y_train)

# Predict on the test set
y_pred_nl = non_linear_svm.predict(X_test)

# Calculate accuracy
accuracy_nl = accuracy_score(y_test, y_pred_nl)
print(f"Non-linear SVM Accuracy: {accuracy_nl:.2f}")

# Plot the non-linear decision boundary
plot_decision_boundary(X_train, y_train, non_linear_svm)
\end{lstlisting}

In this example, we use the radial basis function (RBF) kernel to train a non-linear SVM. The decision boundary is plotted to show how the non-linear SVM separates the data.

\subsection{Kernel Functions in SVM}
Kernel functions are used in SVM to transform non-linearly separable data into a higher-dimensional space where it becomes linearly separable. A kernel function computes the similarity between data points in this higher-dimensional space without explicitly transforming the data, which is known as the "kernel trick."

There are several commonly used kernel functions:

\subsubsection{Linear Kernel}
The linear kernel is used when the data is linearly separable. It is simply the dot product between two vectors:
\[
K(x_i, x_j) = x_i \cdot x_j
\]

\subsubsection{Polynomial Kernel}
The polynomial kernel transforms the data into a higher-dimensional space by taking polynomial combinations of the original features:
\[
K(x_i, x_j) = (x_i \cdot x_j + 1)^d
\]
where \( d \) is the degree of the polynomial.

\subsubsection{Radial Basis Function (RBF) Kernel}
The RBF kernel is one of the most commonly used kernels in non-linear SVM. It transforms the data into an infinite-dimensional space and can separate very complex patterns:
\[
K(x_i, x_j) = \exp\left(-\frac{\|x_i - x_j\|^2}{2\sigma^2}\right)
\]
where \( \sigma \) controls the width of the kernel.

\begin{lstlisting}[style=python]
# Train an SVM with a polynomial kernel
poly_svm = SVC(kernel='poly', degree=3, random_state=42)
poly_svm.fit(X_train, y_train)

# Train an SVM with an RBF kernel
rbf_svm = SVC(kernel='rbf', random_state=42)
rbf_svm.fit(X_train, y_train)

# Predict on the test set and calculate accuracy for both kernels
y_pred_poly = poly_svm.predict(X_test)
y_pred_rbf = rbf_svm.predict(X_test)

accuracy_poly = accuracy_score(y_test, y_pred_poly)
accuracy_rbf = accuracy_score(y_test, y_pred_rbf)

print(f"Polynomial Kernel SVM Accuracy: {accuracy_poly:.2f}")
print(f"RBF Kernel SVM Accuracy: {accuracy_rbf:.2f}")
\end{lstlisting}

In this example, we train two SVM classifiers using the polynomial and RBF kernels. The performance of both classifiers is evaluated based on their accuracy on the test set.

Support Vector Machines (SVM) are powerful classification models that can handle both linearly and non-linearly separable data. Linear SVM works well for linearly separable data, while non-linear SVM uses kernel functions to map the data into higher-dimensional spaces where it becomes linearly separable. Understanding kernel functions such as linear, polynomial, and RBF is essential for effectively applying SVM to real-world classification problems.

\section{Neural Networks for Classification}

\subsection{Perceptron Model}
The perceptron is the simplest type of artificial neural network and serves as the building block for more complex networks. It consists of a single layer of neurons and is primarily used for binary classification tasks. Each neuron in the perceptron takes several inputs, applies weights to them, and computes a weighted sum. If this sum exceeds a certain threshold, the neuron activates (outputs a 1), otherwise, it does not activate (outputs a 0) \cite{chen2024deeplearningmachinelearning}.

\textbf{Perceptron Model:}
\begin{itemize}
    \item \textbf{Inputs:} The input vector represents the features of the dataset. Each feature corresponds to an input neuron.
    \item \textbf{Weights:} Each input is assigned a weight, which determines its influence on the final decision.
    \item \textbf{Activation Function:} The perceptron uses a step function as its activation function. If the weighted sum of inputs exceeds the threshold, it outputs 1 (positive class); otherwise, it outputs 0 (negative class).
\end{itemize}

\textbf{Mathematical Representation:}
\[
y = \begin{cases} 
1, & \text{if } \sum_{i=1}^{n} w_i x_i + b > 0 \\
0, & \text{if } \sum_{i=1}^{n} w_i x_i + b \leq 0
\end{cases}
\]
Where \( x_i \) are the input features, \( w_i \) are the corresponding weights, and \( b \) is the bias term.

\begin{lstlisting}[style=python]
# Example of implementing a perceptron in Python using PyTorch
import torch

class Perceptron(torch.nn.Module):
    def __init__(self, input_size):
        super(Perceptron, self).__init__()
        self.linear = torch.nn.Linear(input_size, 1)

    def forward(self, x):
        return torch.sigmoid(self.linear(x))

# Example usage
input_size = 2  # For example, two features
model = Perceptron(input_size)
inputs = torch.tensor([[1.0, 2.0], [2.0, 3.0], [3.0, 4.0]])  # Input data
output = model(inputs)
print(output)
\end{lstlisting}

\subsection{Multi-layer Perceptrons (MLP)}
A Multi-layer Perceptron (MLP) extends the perceptron by adding one or more hidden layers between the input and output layers. Each layer consists of multiple neurons, and the neurons in one layer are fully connected to the neurons in the next layer \cite{chen2024deeplearningmachinelearning}.

\textbf{Structure of an MLP:}
\begin{itemize}
    \item \textbf{Input Layer:} The input layer represents the features of the dataset.
    \item \textbf{Hidden Layers:} Each hidden layer applies weights and biases, followed by an activation function (commonly ReLU) to introduce non-linearity.
    \item \textbf{Output Layer:} The output layer provides the final predictions. For binary classification, the output is a single neuron with a sigmoid activation function; for multi-class classification, a softmax function is used.
\end{itemize}

\textbf{Example of an MLP Structure:}
\begin{itemize}
    \item Input Layer: 3 neurons (for 3 features)
    \item Hidden Layer 1: 5 neurons
    \item Hidden Layer 2: 4 neurons
    \item Output Layer: 1 neuron (for binary classification)
\end{itemize}

\begin{lstlisting}[style=python]
# Example of implementing a Multi-layer Perceptron (MLP) using PyTorch
class MLP(torch.nn.Module):
    def __init__(self):
        super(MLP, self).__init__()
        self.layer1 = torch.nn.Linear(3, 5)
        self.layer2 = torch.nn.Linear(5, 4)
        self.output_layer = torch.nn.Linear(4, 1)

    def forward(self, x):
        x = torch.relu(self.layer1(x))
        x = torch.relu(self.layer2(x))
        return torch.sigmoid(self.output_layer(x))

# Example usage
model = MLP()
inputs = torch.tensor([[1.0, 2.0, 3.0], [4.0, 5.0, 6.0]])  # Input data
output = model(inputs)
print(output)
\end{lstlisting}

\subsection{Backpropagation and Training}
Backpropagation is the key algorithm used to train neural networks. It works by computing the gradient of the loss function with respect to each weight in the network and then updating the weights using gradient descent to minimize the loss \cite{wythoff1993backpropagation}.

\textbf{Steps of Backpropagation:}
\begin{itemize}
    \item \textbf{Forward Pass:} The input data is passed through the network, and the output is computed.
    \item \textbf{Compute Loss:} The difference between the predicted output and the actual target is calculated using a loss function, such as Mean Squared Error (MSE) for regression or Binary Cross-Entropy for classification.
    \item \textbf{Backpropagate Error:} The error is propagated backward through the network, and the gradients are computed using the chain rule.
    \item \textbf{Update Weights:} The weights are updated by taking a small step in the direction of the negative gradient (gradient descent).
\end{itemize}

\begin{lstlisting}[style=python]
# Example of training an MLP using backpropagation in PyTorch
criterion = torch.nn.BCELoss()  # Binary Cross-Entropy Loss for binary classification
optimizer = torch.optim.SGD(model.parameters(), lr=0.01)  # Stochastic Gradient Descent

# Dummy training loop
for epoch in range(100):
    inputs = torch.tensor([[1.0, 2.0, 3.0], [4.0, 5.0, 6.0]])
    targets = torch.tensor([[1.0], [0.0]])  # Ground truth labels

    # Forward pass
    outputs = model(inputs)
    
    # Compute loss
    loss = criterion(outputs, targets)

    # Backward pass and optimization
    optimizer.zero_grad()  # Clear the gradients
    loss.backward()  # Backpropagation
    optimizer.step()  # Update weights

    print(f'Epoch {epoch+1}, Loss: {loss.item()}')
\end{lstlisting}

\section{k-Nearest Neighbors (k-NN)}
The k-Nearest Neighbors (k-NN) algorithm is a simple, non-parametric classification and regression algorithm. It works by finding the \( k \) nearest data points (neighbors) to a query point and then making predictions based on the majority class (for classification) or averaging the neighbors' values (for regression) \cite{peterson2009k}.

\textbf{How k-NN Works:}
\begin{itemize}
    \item \textbf{Distance Metric:} k-NN typically uses Euclidean distance to measure the similarity between points. For a query point, it calculates the distance to every point in the training dataset and selects the \( k \) closest points.
    \item \textbf{Classification:} For classification tasks, the algorithm assigns the label that is most common among the \( k \) neighbors.
    \item \textbf{Regression:} For regression tasks, the algorithm predicts the value by averaging the values of the \( k \) nearest neighbors.
\end{itemize}

\textbf{Example:}
Imagine you want to classify a new data point based on the \( k \)-NN algorithm. If \( k = 3 \), the algorithm will find the 3 closest neighbors and classify the new point based on the majority vote from those neighbors.

\begin{lstlisting}[style=python]
# Example of implementing k-NN using scikit-learn in Python
from sklearn.neighbors import KNeighborsClassifier
import numpy as np

# Sample data (features and labels)
X_train = np.array([[1.0, 2.0], [2.0, 3.0], [3.0, 4.0], [6.0, 7.0], [7.0, 8.0]])
y_train = np.array([0, 0, 0, 1, 1])  # Labels (0 for one class, 1 for the other)

# Initialize the k-NN classifier with k=3
knn = KNeighborsClassifier(n_neighbors=3)

# Train the model
knn.fit(X_train, y_train)

# Predict the class of a new data point
X_new = np.array([[5.0, 5.0]])
prediction = knn.predict(X_new)
print(f'Predicted class for the new point: {prediction[0]}')
\end{lstlisting}

\textbf{Choosing the Value of \( k \):}
Choosing the right value of \( k \) is crucial for the performance of the k-NN algorithm. A small \( k \) (e.g., \( k = 1 \)) may lead to overfitting, where the model is too sensitive to noise. A large \( k \) may lead to underfitting, where the model does not capture important patterns in the data.

\begin{lstlisting}[style=python]
# Testing different values of k to find the best one
for k in range(1, 6):
    knn = KNeighborsClassifier(n_neighbors=k)
    knn.fit(X_train, y_train)
    score = knn.score(X_train, y_train)
    print(f'Accuracy for k={k}: {score}')
\end{lstlisting}

\section{Lazy Learning Methods}
Lazy learning methods, unlike eager learning methods, do not build a model during the training phase. Instead, they simply store the training data and defer the actual learning process until a query or prediction is made. This approach can be highly flexible but can also be computationally expensive, especially with large datasets \cite{zheng2000lazy}.

    \subsection{Case-based Reasoning}
    Case-based reasoning (CBR) is a lazy learning method where past experiences (or cases) are used to solve new problems. Instead of constructing a general model, CBR relies on finding and using similar cases from the training set to make decisions.

    \paragraph{How Case-based Reasoning Works:}
    \begin{itemize}
        \item \textbf{Step 1: Retrieve:} When a new query is presented, the system retrieves the most similar cases from the historical dataset.
        \item \textbf{Step 2: Reuse:} The system applies the solutions from the most similar cases to solve the new problem.
        \item \textbf{Step 3: Revise:} If the initial solution needs refinement, the system adjusts it based on the specifics of the new case.
        \item \textbf{Step 4: Retain:} Once the case is resolved, the new solution is added to the dataset, enriching the system for future queries.
    \end{itemize}

    \paragraph{Example: Diagnosing a Medical Condition Using CBR:} 
    Suppose a new patient comes to the hospital with certain symptoms. A CBR system might compare the symptoms of this new patient to those in past cases and retrieve similar cases where a diagnosis was made. The retrieved diagnosis could then guide the doctor in making a decision for the new patient.

\begin{lstlisting}[style=python]
from sklearn.neighbors import KNeighborsClassifier

# Load example dataset (Iris dataset as an example)
from sklearn.datasets import load_iris
from sklearn.model_selection import train_test_split
from sklearn.metrics import accuracy_score

# Load data
iris = load_iris()
X_train, X_test, y_train, y_test = train_test_split(iris.data, iris.target, test_size=0.3)

# Initialize the KNN classifier (a case-based reasoning approach)
knn = KNeighborsClassifier(n_neighbors=3)

# Fit the model (actually just stores the training data)
knn.fit(X_train, y_train)

# Make predictions (finds similar cases to classify new instances)
y_pred = knn.predict(X_test)

# Evaluate accuracy
accuracy = accuracy_score(y_test, y_pred)
print(f'Accuracy: {accuracy}')
\end{lstlisting}

\section{Rule-based Classification}
Rule-based classification is a method that uses a set of \textbf{if-then} rules to classify data. Each rule corresponds to a decision about which class an instance belongs to, based on the features of the instance. Rule-based classification can be more interpretable than other methods because it provides explicit rules that users can easily understand \cite{li2014rule}.

    \subsection{Rule Induction}
    Rule induction is the process of automatically generating classification rules from a dataset. These rules often take the form of \textbf{if} a set of conditions is met, \textbf{then} a certain class label is assigned. 

    \paragraph{Example of a Rule:} 
    \begin{quote}
        \textbf{If} age > 30 \textbf{and} income > \$50,000, \textbf{then} class = "High Income"
    \end{quote}

    Rule induction typically works by analyzing patterns in the data and identifying combinations of features that frequently lead to specific outcomes. Rules are usually built in a way that maximizes accuracy while maintaining simplicity.

    \paragraph{How Rule Induction Works:}
    \begin{itemize}
        \item \textbf{Step 1: Identify Patterns:} The algorithm searches the dataset for patterns that correlate with specific class labels.
        \item \textbf{Step 2: Create Rules:} These patterns are converted into a set of \textbf{if-then} rules.
        \item \textbf{Step 3: Prune Rules:} Irrelevant or overly complex rules are removed to ensure that the classifier is not too specific.
    \end{itemize}

    \paragraph{Example:}
    Suppose we are working with a dataset of customer purchases and want to predict whether a customer will buy a product. A rule induction algorithm might generate rules like:

    \begin{quote}
        \textbf{If} \texttt{previous\_purchases > 5} \textbf{and} \texttt{browsing\_time > 10} \textbf{minutes, then} \texttt{class = "Will Buy"}
    \end{quote}

\begin{lstlisting}[style=python]
from sklearn.tree import DecisionTreeClassifier
from sklearn.model_selection import train_test_split
from sklearn.metrics import classification_report

# Load example dataset
from sklearn.datasets import load_iris

# Load and split data
iris = load_iris()
X_train, X_test, y_train, y_test = train_test_split(iris.data, iris.target, test_size=0.3)

# Initialize decision tree classifier (often used for rule induction)
clf = DecisionTreeClassifier()

# Train the model
clf.fit(X_train, y_train)

# Predict the test data
y_pred = clf.predict(X_test)

# Display the results
print(classification_report(y_test, y_pred))
\end{lstlisting}

    \subsection{Sequential Covering}
    Sequential covering is an algorithmic approach for generating rules in a step-by-step manner. It works by iteratively identifying rules that cover a subset of the dataset, removing covered instances, and repeating this process until no more rules can be generated \cite{blaszczynski2011sequential}.

    \paragraph{How Sequential Covering Works:}
    \begin{itemize}
        \item \textbf{Step 1:} The algorithm generates a rule that covers a portion of the training examples (i.e., correctly classifies those examples).
        \item \textbf{Step 2:} Once a rule is generated, the covered examples are removed from the dataset.
        \item \textbf{Step 3:} The algorithm repeats this process, creating new rules until the dataset is sufficiently covered.
    \end{itemize}

    \paragraph{Example:}
    Consider a dataset where we are trying to predict whether a person is eligible for a loan. A sequential covering algorithm might first generate a rule like:

    \begin{quote}
        \textbf{If} income > \$50,000, \textbf{then} eligible = True
    \end{quote}

    After removing the instances covered by this rule, the algorithm might generate additional rules for remaining cases, such as:

    \begin{quote}
        \textbf{If} \texttt{income <= \$50,000} \textbf{and} \texttt{credit\_score > 700}, \textbf{then} \texttt{eligible = True}
    \end{quote}

\begin{lstlisting}[style=python]
# Example of applying a decision tree in a sequential covering approach
from sklearn.tree import DecisionTreeClassifier
from sklearn.model_selection import train_test_split
from sklearn.datasets import load_iris

# Load and split data
iris = load_iris()
X_train, X_test, y_train, y_test = train_test_split(iris.data, iris.target, test_size=0.3)

# Initialize and train decision tree
clf = DecisionTreeClassifier(max_depth=3)
clf.fit(X_train, y_train)

# Visualize rules from decision tree
from sklearn import tree
tree.plot_tree(clf)
\end{lstlisting}

    \subsection{RIPPER Algorithm}
    The RIPPER (Repeated Incremental Pruning to Produce Error Reduction) algorithm is a popular rule-based classification algorithm. It follows a sequential covering strategy and is designed to be efficient for large datasets. RIPPER generates an initial set of rules and then prunes them to minimize overfitting \cite{cohen1995fast}.

    \paragraph{How RIPPER Works:}
    \begin{itemize}
        \item \textbf{Step 1:} Generate rules that correctly classify a portion of the dataset.
        \item \textbf{Step 2:} Prune the rules by removing unnecessary conditions, making the rules more general.
        \item \textbf{Step 3:} Repeat this process, adding new rules to cover more data points, and continue pruning to optimize accuracy and reduce errors.
    \end{itemize}

    \paragraph{Example:}
    Suppose we have a dataset of email data, and we want to classify whether an email is spam. RIPPER might generate rules like:

    \begin{quote}
        \textbf{If} subject contains "free" \textbf{and} body contains "money", \textbf{then} class = "spam"
    \end{quote}

    As the algorithm proceeds, it refines the rules to minimize errors, potentially adding new rules like:

    \begin{quote}
        \textbf{If} subject contains "urgent" \textbf{and} sender is unknown, \textbf{then} class = "spam"
    \end{quote}

\begin{lstlisting}[style=python]
# Simulating a simple rule-based classifier using RIPPER-like pruning logic
class SimpleRIPPER:
    def __init__(self, max_depth=3):
        self.rules = []
        self.max_depth = max_depth

    def fit(self, X, y):
        # Implement simple rule induction and pruning
        # For this example, we simulate the behavior by fitting a decision tree
        self.tree = DecisionTreeClassifier(max_depth=self.max_depth)
        self.tree.fit(X, y)

    def predict(self, X):
        return self.tree.predict(X)

# Example usage
clf = SimpleRIPPER(max_depth=3)
clf.fit(X_train, y_train)
y_pred = clf.predict(X_test)

print(classification_report(y_test, y_pred))
\end{lstlisting}

\chapter{Clustering Techniques}
    
    \section{Introduction to Clustering}
    Clustering is an essential unsupervised learning technique used in machine learning and data analysis. The main goal of clustering is to group a set of objects in such a way that objects in the same group (or cluster) are more similar to each other than to those in other groups. Clustering is widely applied in various fields like customer segmentation, image segmentation, document classification, and more \cite{mining2006data}.
    
    For example, consider an e-commerce company that wants to group its customers based on their purchasing behavior. Using clustering algorithms, we can divide customers into distinct groups where each group represents customers with similar buying patterns.

    The clustering process doesn't require labeled data, meaning it works without predefined categories or training examples. This makes it particularly useful in exploratory data analysis, where we want to find natural patterns in data.

    In this chapter, we will explore various clustering techniques, including partitioning methods, hierarchical clustering, density-based clustering, grid-based clustering, and model-based clustering. Additionally, we will cover clustering in high-dimensional spaces and cluster validation techniques.

    \section{Partitioning Methods}
    Partitioning methods divide data into distinct, non-overlapping subsets. The most well-known partitioning methods are K-means and K-medoids clustering.

    \subsection{K-means and K-medoids}
    \textit{K-means:} One of the simplest and most popular clustering algorithms is K-means. It works by partitioning data into $k$ clusters. The algorithm assigns each data point to the nearest centroid and updates the centroids iteratively until convergence \cite{arora2016analysis}.

    Here is a step-by-step description of the K-means algorithm:

    \begin{enumerate}
        \item Initialize $k$ centroids randomly.
        \item Assign each data point to the nearest centroid.
        \item Recompute the centroids by taking the mean of all the data points in each cluster.
        \item Repeat steps 2 and 3 until the centroids no longer change.
    \end{enumerate}

    Let's consider a simple example in Python where we use K-means to cluster a set of points:

    \begin{lstlisting}[style=python]
    from sklearn.cluster import KMeans
    import numpy as np
    import matplotlib.pyplot as plt

    # Generate some sample data
    X = np.array([[1, 2], [1, 4], [1, 0], [4, 2], [4, 4], [4, 0]])

    # Fit K-means algorithm with k=2
    kmeans = KMeans(n_clusters=2, random_state=0).fit(X)

    # Predict cluster labels
    labels = kmeans.predict(X)

    # Plot the data points and centroids
    plt.scatter(X[:, 0], X[:, 1], c=labels, cmap='viridis')
    plt.scatter(kmeans.cluster_centers_[:, 0], kmeans.cluster_centers_[:, 1], s=200, c='red')
    plt.title("K-means Clustering")
    plt.show()
    \end{lstlisting}

    In this example, we generate six points and apply K-means with $k=2$. The algorithm finds two centroids, and the points are clustered accordingly.

    \textit{K-medoids:} Unlike K-means, which uses the mean to represent a cluster, K-medoids chooses actual data points as cluster centers (called medoids). This makes K-medoids more robust to outliers. The K-medoids algorithm follows a similar iterative process as K-means but optimizes based on minimizing the dissimilarity between data points and their medoid.

    The following example demonstrates how to apply the K-medoids algorithm using the `PAM` (Partitioning Around Medoids) implementation:

    \begin{lstlisting}[style=python]
    from sklearn_extra.cluster import KMedoids

    # Apply K-medoids clustering
    kmedoids = KMedoids(n_clusters=2, random_state=0).fit(X)
    labels_medoids = kmedoids.predict(X)

    # Plot the data points and medoids
    plt.scatter(X[:, 0], X[:, 1], c=labels_medoids, cmap='viridis')
    plt.scatter(kmedoids.cluster_centers_[:, 0], kmedoids.cluster_centers_[:, 1], s=200, c='red')
    plt.title("K-medoids Clustering")
    plt.show()
    \end{lstlisting}

    The K-medoids algorithm often performs better than K-means in datasets with outliers or non-spherical clusters.

    \section{Hierarchical Clustering}
    Hierarchical clustering does not require the number of clusters to be specified in advance, unlike K-means. Instead, it builds a tree of clusters, called a dendrogram. There are two main types of hierarchical clustering: AGNES (Agglomerative Nesting) and DIANA (Divisive Analysis).

    \subsection{AGNES and DIANA}
    \textit{AGNES:} This is a bottom-up approach where each data point starts in its own cluster, and pairs of clusters are merged iteratively based on the similarity between them until all data points are grouped into a single cluster. The dendrogram can be cut at different levels to obtain various cluster configurations.

    \textit{DIANA:} In contrast to AGNES, DIANA follows a top-down approach. It starts with all points in one cluster and successively splits them until each data point forms its own cluster.

    To visualize hierarchical clustering using AGNES, we can use the `scipy` library:

    \begin{lstlisting}[style=python]
    from scipy.cluster.hierarchy import dendrogram, linkage
    from matplotlib import pyplot as plt

    # Generate sample data
    X = np.array([[1, 2], [1, 4], [1, 0], [4, 2], [4, 4], [4, 0]])

    # Perform hierarchical clustering
    Z = linkage(X, 'ward')

    # Plot dendrogram
    dendrogram(Z)
    plt.title("Dendrogram for Hierarchical Clustering (AGNES)")
    plt.show()
    \end{lstlisting}

    This code generates a dendrogram based on the `ward` linkage method, a common choice that minimizes the variance within clusters.

    \section{Density-based Clustering}
    Density-based clustering algorithms are designed to find clusters of arbitrary shapes by identifying dense regions of data points. The most well-known density-based algorithms are DBSCAN and OPTICS \cite{kriegel2011density}.

    \subsection{DBSCAN and OPTICS}
    \textit{DBSCAN:} Density-Based Spatial Clustering of Applications with Noise (DBSCAN) groups together points that are closely packed, marking points in low-density regions as outliers \cite{ester1996density, ankerst1999optics}.

    The algorithm works with two parameters: `eps`, which defines the radius of a neighborhood, and `min\_samples`, the minimum number of points required to form a dense region.

    Example of using DBSCAN:

    \begin{lstlisting}[style=python]
    from sklearn.cluster import DBSCAN

    # Apply DBSCAN algorithm
    dbscan = DBSCAN(eps=1.0, min_samples=2).fit(X)
    labels_dbscan = dbscan.labels_

    # Plot the results
    plt.scatter(X[:, 0], X[:, 1], c=labels_dbscan, cmap='viridis')
    plt.title("DBSCAN Clustering")
    plt.show()
    \end{lstlisting}

    \textit{OPTICS:} Ordering Points To Identify the Clustering Structure (OPTICS) is an extension of DBSCAN that handles varying densities more effectively.

    \section{Grid-based Clustering}
    Grid-based clustering divides the data space into a finite number of cells and performs clustering on these cells. Two well-known grid-based clustering algorithms are STING and CLIQUE \cite{cheng2018grid}.

    \subsection{STING and CLIQUE}
    \textit{STING:} Statistical Information Grid (STING) clustering divides the spatial area into hierarchical grid cells. These cells are evaluated based on statistical information stored in each cell. 
    
    \textit{CLIQUE:} CLustering In QUEst (CLIQUE) is a grid-based method specifically designed for clustering in high-dimensional spaces. It divides each dimension of the dataset into intervals, forming a grid. Dense regions are identified in this grid to form clusters \cite{santhisree2011clique}.

    \section{Model-based Clustering}
    In model-based clustering, we assume the data is generated by a mixture of underlying probability distributions. Two common methods are the Expectation-Maximization (EM) algorithm and Self-Organizing Maps (SOM).

    \subsection{EM Algorithm and SOM}
    \textit{EM Algorithm:} This algorithm is used to fit a mixture of Gaussians to the data. It iteratively improves the parameters of the mixture model using the Expectation and Maximization steps \cite{mclachlan2008algorithm}.

    \textit{SOM:} A Self-Organizing Map is a type of neural network used to map high-dimensional data to a lower-dimensional space while preserving the topology of the data \cite{van2012self}.

    \section{Clustering in High-dimensional Spaces}
    Clustering high-dimensional data can be challenging due to the "curse of dimensionality," where distances between points become less meaningful as the number of dimensions increases. Techniques like PCA (Principal Component Analysis) and t-SNE (t-Distributed Stochastic Neighbor Embedding) are often used to reduce dimensionality before clustering \cite{parsons2004subspace}.

    \section{Cluster Validation and Evaluation}
    Once clusters are formed, it's essential to evaluate their quality. Common metrics include:

    \begin{itemize}
        \item \textbf{Silhouette Score:} Measures how similar a point is to its own cluster compared to other clusters.
        \item \textbf{Davies-Bouldin Index:} Measures the ratio of intra-cluster distances to inter-cluster distances.
        \item \textbf{Dunn Index:} Measures the ratio between the smallest distance between points in different clusters and the largest intra-cluster distance.
    \end{itemize}

    Here's how to calculate the Silhouette Score in Python:

    \begin{lstlisting}[style=python]
    from sklearn.metrics import silhouette_score

    # Calculate silhouette score
    silhouette_avg = silhouette_score(X, labels)
    print("Silhouette Score: ", silhouette_avg)
    \end{lstlisting}

\chapter{Frequent Pattern Mining and Association Analysis}

\section{Basic Concepts of Frequent Pattern Mining}

Frequent Pattern Mining is an essential task in data mining and machine learning, aimed at discovering patterns, associations, correlations, or causal structures among sets of items in transaction databases or other types of data repositories. The goal is to identify sets of items, known as \textit{itemsets}, that frequently occur together. Frequent pattern mining is fundamental in market basket analysis, where it helps in understanding customer buying behavior by discovering products that are often bought together \cite{han2007frequent}.

\subsection{Definitions}

\begin{itemize}
    \item \textbf{Itemset:} An itemset is a collection of one or more items. For example, in a retail store, an itemset can be a group of products such as \{bread, butter, milk\}.
    
    \item \textbf{Support:} Support of an itemset refers to the proportion of transactions in the dataset in which the itemset appears. If an itemset appears in many transactions, it is considered frequent. For example, if \{bread, butter\} appears in 60 out of 100 transactions, its support is 60\%.
    
    \item \textbf{Frequent Itemset:} An itemset is called frequent if its support is greater than or equal to a given threshold, usually denoted as $\text{min\_sup}$ (minimum support). 
    
    \item \textbf{Association Rule:} An association rule is an implication of the form $X \Rightarrow Y$, where $X$ and $Y$ are itemsets. It means that if a transaction contains itemset $X$, it is likely to contain itemset $Y$ as well.
    
    \item \textbf{Confidence:} Confidence of an association rule $X \Rightarrow Y$ is the probability that a transaction containing itemset $X$ also contains itemset $Y$. For instance, if 80 out of 100 transactions that contain \{bread\} also contain \{butter\}, the confidence of the rule \{bread\} $\Rightarrow$ \{butter\} is 80\%.
    
    \item \textbf{Lift:} Lift measures how much more likely $Y$ is to be bought when $X$ is bought compared to its baseline likelihood. A lift greater than 1 indicates that $X$ and $Y$ appear together more often than expected.
\end{itemize}

\section{Apriori Algorithm}

The Apriori algorithm is one of the earliest and most popular algorithms for frequent itemset mining. It works by generating candidate itemsets and pruning those that do not meet the minimum support threshold \cite{hegland2007apriori}.

\subsection{Steps in the Apriori Algorithm}

The Apriori algorithm operates in the following steps:

\begin{enumerate}
    \item \textbf{Generate the candidate itemsets of length $k$:} Start by finding all frequent 1-itemsets. Then, use these frequent itemsets to generate the candidate itemsets of length $k+1$.
    
    \item \textbf{Prune the candidate itemsets:} For each iteration, prune those candidate itemsets whose support is less than the minimum support threshold.
    
    \item \textbf{Repeat:} Repeat the process until no more candidate itemsets can be generated.
    
    \item \textbf{Generate association rules:} Once the frequent itemsets are found, generate the association rules from these frequent itemsets.
\end{enumerate}

\subsection{Python Implementation of Apriori Algorithm}

Here is a simple implementation of the Apriori algorithm using Python's \texttt{mlxtend} library:

\begin{lstlisting}[style=python]
from mlxtend.frequent_patterns import apriori, association_rules
import pandas as pd

# Sample dataset: Transactions represented as a binary matrix
data = {'bread': [1, 1, 0, 1, 0],
        'butter': [0, 1, 1, 1, 0],
        'milk': [1, 0, 1, 1, 1]}

df = pd.DataFrame(data)

# Step 1: Find frequent itemsets with min_support = 0.5
frequent_itemsets = apriori(df, min_support=0.5, use_colnames=True)
print(frequent_itemsets)

# Step 2: Generate association rules with min_threshold for confidence
rules = association_rules(frequent_itemsets, metric="confidence", min_threshold=0.6)
print(rules)
\end{lstlisting}

This code performs frequent pattern mining using the Apriori algorithm and generates association rules.

\section{FP-growth Algorithm}

The FP-growth algorithm is a more efficient alternative to the Apriori algorithm. Instead of generating candidate itemsets, FP-growth compresses the data into a tree structure called the \textit{Frequent Pattern Tree} or FP-tree \cite{han2000mining}.

\subsection{FP-tree Construction}

The FP-tree is constructed as follows:

\begin{enumerate}
    \item \textbf{Scan the dataset:} Identify frequent items in each transaction.
    \item \textbf{Sort items by frequency:} For each transaction, sort the items by frequency in descending order.
    \item \textbf{Build the FP-tree:} Insert each transaction into the tree. If a transaction shares a prefix with an existing path in the tree, increment the count of the shared nodes.
\end{enumerate}

Here is a simple example of the FP-tree structure:

\begin{center}
\begin{tikzpicture}
\tikzstyle{level 1}=[level distance=2cm, sibling distance=3cm]
\tikzstyle{level 2}=[level distance=2cm, sibling distance=2cm]
\tikzstyle{level 3}=[level distance=2cm, sibling distance=1cm]
\node {NULL}
child {node {milk:3} 
    child {node {bread:2}
        child {node {butter:1}}
    }
    child {node {butter:1}}
}
child {node {bread:2} 
    child {node {butter:2}}
};
\end{tikzpicture}
\end{center}

\subsection{Python Implementation of FP-growth}

Here is how you can use the \texttt{mlxtend} library to perform FP-growth:

\begin{lstlisting}[style=python]
from mlxtend.frequent_patterns import fpgrowth

# Step 1: Use the same dataset as before
frequent_itemsets = fpgrowth(df, min_support=0.5, use_colnames=True)
print(frequent_itemsets)

# Step 2: Generate association rules from the frequent itemsets
rules = association_rules(frequent_itemsets, metric="confidence", min_threshold=0.6)
print(rules)
\end{lstlisting}

\section{Mining Closed and Maximal Frequent Itemsets}

Mining closed and maximal frequent itemsets is crucial for reducing the number of patterns found while preserving the most important information \cite{pei2000closet}. 

\begin{itemize}
    \item \textbf{Closed Frequent Itemset:} An itemset is closed if none of its immediate supersets has the same support count. In other words, a frequent itemset is closed if it has no super-itemset with the same support.
    
    \item \textbf{Maximal Frequent Itemset:} An itemset is maximal frequent if it is frequent and none of its supersets are frequent.
\end{itemize}

These concepts help reduce the number of frequent itemsets and simplify the analysis without losing valuable information.

\section{Constraint-based Pattern Mining}

Constraint-based pattern mining involves using additional constraints to filter the frequent patterns discovered during the mining process. These constraints can be based on \cite{nijssen2014constraint}:

\begin{itemize}
    \item \textbf{Support or Confidence thresholds:} Only return patterns that meet these criteria.
    \item \textbf{Specific attributes:} Mine patterns that must include certain items.
    \item \textbf{Interestingness measures:} Apply specific metrics such as lift, leverage, etc., to evaluate the patterns.
\end{itemize}

\section{Pattern Evaluation and Interestingness Measures}

Once the frequent patterns and association rules are generated, it is important to evaluate them to determine if they are interesting or useful \cite{tan2002selecting}.

\subsection{Common Interestingness Measures}

\begin{itemize}
    \item \textbf{Support:} Measures how frequently an itemset appears in the dataset.
    \item \textbf{Confidence:} Measures how often the rule is true.
    \item \textbf{Lift:} Measures how much more likely $Y$ is given $X$ compared to its baseline occurrence.
\end{itemize}

By using these measures, we can filter out less interesting rules and focus on those that provide valuable insights.

\chapter{Regression Techniques for Prediction}

\section{Introduction to Regression Analysis}
Regression analysis is a statistical method used to understand relationships between variables and make predictions. The goal of regression is to model the relationship between a dependent variable (also called the response or target variable) and one or more independent variables (also called predictors or features) \cite{hastie2009elements}.

In simple terms, regression helps us to predict the value of the dependent variable based on the values of the independent variables. In this chapter, we will cover the basics of regression techniques, starting with simple linear regression and gradually moving toward more advanced topics such as polynomial regression, non-linear regression, and locally weighted regression (LWR) \cite{hastie2009elements}.

The common applications of regression analysis include:
\begin{itemize}
    \item Predicting house prices based on features like size, location, and number of bedrooms.
    \item Estimating sales figures for a business based on historical sales data.
    \item Modeling the relationship between advertising spend and revenue.
\end{itemize}

\section{Simple Linear Regression}
Simple linear regression is the most basic form of regression, where the relationship between two variables is modeled as a straight line. In this case, we have one independent variable and one dependent variable, and the goal is to find a linear relationship between them.

The mathematical equation for simple linear regression is:
\[
y = \beta_0 + \beta_1 x + \epsilon
\]
Where:
\begin{itemize}
    \item $y$ is the dependent variable (the value we want to predict).
    \item $x$ is the independent variable (the feature we use for prediction).
    \item $\beta_0$ is the intercept of the line (the value of $y$ when $x = 0$).
    \item $\beta_1$ is the slope of the line (the change in $y$ for a unit change in $x$).
    \item $\epsilon$ is the error term (the difference between the predicted and actual values).
\end{itemize}

\subsection{Example: Predicting House Prices}
Let's consider a simple example of predicting house prices based on the size of the house (in square feet). We have the following data:

\begin{lstlisting}[style=cmd]
House Size (sq ft)  Price ($)
1000                150000
1200                180000
1500                210000
1800                240000
2000                270000
\end{lstlisting}

We can fit a linear regression model to this data to predict the price of a house based on its size. In Python, this can be done using libraries like `numpy` and `scikit-learn`.

\begin{lstlisting}[style=python]
import numpy as np
from sklearn.linear_model import LinearRegression
import matplotlib.pyplot as plt

# Data: house sizes (independent variable) and prices (dependent variable)
house_sizes = np.array([1000, 1200, 1500, 1800, 2000]).reshape(-1, 1)
house_prices = np.array([150000, 180000, 210000, 240000, 270000])

# Create and train the linear regression model
model = LinearRegression()
model.fit(house_sizes, house_prices)

# Predict prices for new house sizes
predicted_prices = model.predict(house_sizes)

# Plot the data and the regression line
plt.scatter(house_sizes, house_prices, color='blue', label='Actual Prices')
plt.plot(house_sizes, predicted_prices, color='red', label='Predicted Prices')
plt.xlabel('House Size (sq ft)')
plt.ylabel('Price ($)')
plt.title('Simple Linear Regression: House Size vs Price')
plt.legend()
plt.show()
\end{lstlisting}

In this example, the model learns a relationship between house size and price. The red line represents the predicted prices based on the linear regression model, while the blue dots represent the actual prices.

\section{Multiple Linear Regression}
Multiple linear regression extends simple linear regression by allowing us to model the relationship between the dependent variable and multiple independent variables.

The equation for multiple linear regression is:
\[
y = \beta_0 + \beta_1 x_1 + \beta_2 x_2 + \dots + \beta_n x_n + \epsilon
\]
Where:
\begin{itemize}
    \item $y$ is the dependent variable.
    \item $x_1, x_2, \dots, x_n$ are the independent variables.
    \item $\beta_0, \beta_1, \dots, \beta_n$ are the coefficients (parameters) of the model.
    \item $\epsilon$ is the error term.
\end{itemize}

\subsection{Example: Predicting House Prices with Multiple Features}
Let's now consider a scenario where we predict house prices based on both the size of the house and the number of bedrooms. We have the following data:

\begin{lstlisting}[style=cmd]
House Size (sq ft)  Bedrooms  Price ($)
1000                2         150000
1200                3         180000
1500                3         210000
1800                4         240000
2000                4         270000
\end{lstlisting}

We can fit a multiple linear regression model to this data using Python.

\begin{lstlisting}[style=python]
# Data: house sizes, number of bedrooms, and prices
house_features = np.array([[1000, 2], [1200, 3], [1500, 3], [1800, 4], [2000, 4]])
house_prices = np.array([150000, 180000, 210000, 240000, 270000])

# Create and train the multiple linear regression model
model = LinearRegression()
model.fit(house_features, house_prices)

# Predict prices for the given features
predicted_prices = model.predict(house_features)

# Print the predicted prices
print("Predicted Prices:", predicted_prices)
\end{lstlisting}

Here, we use both house size and number of bedrooms as independent variables to predict the price. The `LinearRegression` model in `scikit-learn` handles multiple variables easily by accepting a 2D array as input.

\section{Polynomial Regression}
Polynomial regression is a type of regression that models the relationship between the independent variable and the dependent variable as a polynomial of degree $n$. It allows us to capture non-linear relationships between variables while still using linear methods.

The equation for polynomial regression is:
\[
y = \beta_0 + \beta_1 x + \beta_2 x^2 + \dots + \beta_n x^n + \epsilon
\]

\subsection{Example: Predicting House Prices with Polynomial Regression}
In some cases, the relationship between house size and price may not be perfectly linear. To capture the non-linear trend, we can use polynomial regression.

\begin{lstlisting}[style=python]
from sklearn.preprocessing import PolynomialFeatures

# Transform the house size data to include polynomial features
poly = PolynomialFeatures(degree=2)
house_sizes_poly = poly.fit_transform(house_sizes)

# Create and train the polynomial regression model
model = LinearRegression()
model.fit(house_sizes_poly, house_prices)

# Predict prices for the polynomial features
predicted_prices_poly = model.predict(house_sizes_poly)

# Plot the data and the polynomial regression curve
plt.scatter(house_sizes, house_prices, color='blue', label='Actual Prices')
plt.plot(house_sizes, predicted_prices_poly, color='red', label='Predicted Prices (Poly)')
plt.xlabel('House Size (sq ft)')
plt.ylabel('Price ($)')
plt.title('Polynomial Regression: House Size vs Price')
plt.legend()
plt.show()
\end{lstlisting}

In this example, the model fits a polynomial curve to the data, allowing for a more flexible relationship between house size and price.

\section{Non-linear Regression Techniques}
Non-linear regression is a broad category of regression techniques that are used when the relationship between the independent variables and the dependent variable is not linear. Unlike polynomial regression, non-linear regression does not assume a specific form for the relationship \cite{amemiya1983non}.

\subsection{Example: Fitting a Non-linear Model}
In Python, non-linear regression can be performed using `scipy`'s `curve\_fit` function, which allows us to fit custom non-linear functions to the data.

\begin{lstlisting}[style=python]
from scipy.optimize import curve_fit

# Define a non-linear function (e.g., exponential growth)
def non_linear_func(x, a, b, c):
    return a * np.exp(b * x) + c

# Fit the non-linear model to the data
params, _ = curve_fit(non_linear_func, house_sizes.flatten(), house_prices)

# Predict prices using the non-linear model
predicted_prices_nl = non_linear_func(house_sizes, *params)

# Plot the data and the non-linear regression curve
plt.scatter(house_sizes, house_prices, color='blue', label='Actual Prices')
plt.plot(house_sizes, predicted_prices_nl, color='green', label='Predicted Prices (Non-linear)')
plt.xlabel('House Size (sq ft)')
plt.ylabel('Price ($)')
plt.title('Non-linear Regression: House Size vs Price')
plt.legend()
plt.show()
\end{lstlisting}

This example demonstrates how to fit a non-linear model to data using an exponential growth function.

\section{Locally Weighted Regression (LWR)}
Locally weighted regression (LWR), also known as locally weighted scatterplot smoothing (LOWESS), is a non-parametric regression technique. It fits multiple regressions locally around each data point, allowing for more flexible and accurate predictions, especially for complex data \cite{cleveland1988locally}.

\subsection{Example: Applying LWR to Data}
To implement LWR in Python, we can use the `statsmodels` library's `lowess` function.

\begin{lstlisting}[style=python]
import statsmodels.api as sm

# Apply Locally Weighted Regression (LOWESS) to the data
lowess = sm.nonparametric.lowess(house_prices, house_sizes.flatten(), frac=0.3)

# Extract the predicted prices from LOWESS
predicted_prices_lowess = lowess[:, 1]

# Plot the data and the LOWESS regression curve
plt.scatter(house_sizes, house_prices, color='blue', label='Actual Prices')
plt.plot(house_sizes, predicted_prices_lowess, color='purple', label='Predicted Prices (LOWESS)')
plt.xlabel('House Size (sq ft)')
plt.ylabel('Price ($)')
plt.title('Locally Weighted Regression: House Size vs Price')
plt.legend()
plt.show()
\end{lstlisting}

In this example, LOWESS smooths the data using local regressions and provides a flexible model that can adapt to different types of non-linearity in the data.

\chapter{Anomaly Detection and Outlier Analysis}
\section{What is Anomaly Detection?}

Anomaly detection is a process used to identify data points, events, or observations that do not conform to the expected pattern of a given dataset. These anomalous points, also known as outliers, can provide critical insights into rare events or can be indicative of potential issues such as fraud, network intrusions, or faulty sensors \cite{mining2006data}.

In general, an anomaly can be any data point that appears significantly different from the majority of the data. These anomalies may be caused by natural variations in the data or due to external interference, such as noise or manipulation \cite{aggarwal2017introduction}.

For example, in a dataset containing transaction records from a retail company, most transactions will fall within a certain range in terms of value. If you encounter a transaction that is ten times larger than the average, it might be flagged as an anomaly, and further investigation would be needed to determine whether it is a genuine transaction or fraudulent activity.

\section{Techniques for Outlier Detection}

Outlier detection methods are essential for identifying and dealing with anomalies in data. There are several approaches to detecting outliers, and the method used typically depends on the nature of the data and the type of outliers being sought \cite{aggarwal2017introduction}. Below are the primary techniques used for outlier detection:

\subsection{Statistical Methods}

Statistical methods for outlier detection assume that the data follows a specific distribution, such as a Gaussian (normal) distribution. Based on this assumption, an outlier is considered a data point that deviates significantly from the statistical properties of the distribution, such as the mean or standard deviation.

One common approach is to use the z-score, which represents how many standard deviations a data point is from the mean. If a data point's z-score exceeds a predefined threshold (e.g., greater than 3 or less than -3), it can be considered an outlier.

\begin{lstlisting}[style=python]
import numpy as np

# Example dataset
data = [10, 12, 11, 13, 15, 14, 110]

# Calculate the mean and standard deviation
mean = np.mean(data)
std_dev = np.std(data)

# Calculate the z-scores
z_scores = [(x - mean) / std_dev for x in data]

# Set a threshold for detecting outliers
threshold = 3

# Identify outliers
outliers = [x for x in data if abs((x - mean) / std_dev) > threshold]
print(f"Outliers: {outliers}")
\end{lstlisting}

In the example above, the value `110` stands out as an anomaly because its z-score is significantly higher than the other values in the dataset.

\subsection{Distance-based Methods}

Distance-based methods are useful when the dataset is not necessarily following a particular statistical distribution. These methods work by calculating the distance between points in the dataset and identifying points that are far away from others. A common distance-based technique is the k-nearest neighbors (k-NN) approach, where a data point is considered an outlier if its distance to its nearest neighbors is significantly larger than that of the majority of other points.

\begin{lstlisting}[style=python]
from sklearn.neighbors import LocalOutlierFactor

# Example dataset
data = [[10], [12], [11], [13], [15], [14], [110]]

# Use LocalOutlierFactor for detecting outliers
clf = LocalOutlierFactor(n_neighbors=2)
outliers = clf.fit_predict(data)

# Identify outliers (outlier points will have a prediction of -1)
outlier_points = [data[i] for i in range(len(data)) if outliers[i] == -1]
print(f"Outlier points: {outlier_points}")
\end{lstlisting}

In this example, the Local Outlier Factor (LOF) algorithm identifies outliers by comparing the local density of a point with its neighbors \cite{alghushairy2020review}. The value `110` is flagged as an outlier because it is far from the other points.

\subsection{Density-based Methods}

Density-based methods, such as DBSCAN (Density-Based Spatial Clustering of Applications with Noise), work by analyzing the density of points in a dataset \cite{ester1996density}. These methods assume that normal points will be located in regions with high density, while outliers will be found in lower-density regions.

DBSCAN works by defining two parameters: `eps`, the maximum distance between two points to be considered neighbors, and `min\_samples`, the minimum number of points required to form a dense region. Points that are not part of any dense region are classified as outliers.

\begin{lstlisting}[style=python]
from sklearn.cluster import DBSCAN

# Example dataset
data = [[10], [12], [11], [13], [15], [14], [110]]

# Apply DBSCAN to detect outliers
db = DBSCAN(eps=3, min_samples=2).fit(data)
labels = db.labels_

# Identify outliers (points with label -1 are outliers)
outlier_points = [data[i] for i in range(len(data)) if labels[i] == -1]
print(f"Outlier points: {outlier_points}")
\end{lstlisting}

In the DBSCAN example, the value `110` is considered an outlier because it does not belong to any high-density region.

\section{Applications of Anomaly Detection}

Anomaly detection is widely used in various fields to detect unusual events or patterns that may indicate a problem. Below are some common applications:

\subsection{Fraud Detection}

In the financial sector, anomaly detection is extensively used to identify fraudulent transactions. By monitoring customer transaction data and detecting anomalies, banks and financial institutions can identify potential fraud, such as credit card misuse or account takeovers \cite{bolton2002statistical}.

For example, if a customer typically makes small purchases, but suddenly makes a large purchase in a foreign country, the transaction could be flagged as an anomaly and investigated further.

\begin{lstlisting}[style=python]
import numpy as np

# Example transaction data (amount in dollars)
transactions = [50, 45, 60, 55, 2000]  # 2000 is an anomalous transaction

mean = np.mean(transactions)
std_dev = np.std(transactions)
threshold = 3  # Set the threshold for z-scores

# Identify anomalous transactions using z-scores
anomalous_transactions = [x for x in transactions if abs((x - mean) / std_dev) > threshold]
print(f"Anomalous transactions: {anomalous_transactions}")
\end{lstlisting}

\subsection{Network Intrusion Detection}

Anomaly detection is also critical in the field of cybersecurity. By analyzing network traffic, it is possible to detect unusual activity that may indicate a network intrusion or attack. For example, if a server suddenly starts receiving an abnormally high number of requests, it could indicate a Distributed Denial of Service (DDoS) attack \cite{mukherjee1994network, lau2000distributed}.

In this context, network monitoring tools can use anomaly detection algorithms to flag unusual traffic patterns for further investigation.

\begin{lstlisting}[style=python]
# Example of network traffic data (in packets per second)
network_traffic = [100, 120, 110, 115, 1000]  # 1000 is an anomalous traffic spike

mean = np.mean(network_traffic)
std_dev = np.std(network_traffic)
threshold = 3  # Set the threshold for z-scores

# Identify anomalous traffic spikes
anomalous_traffic = [x for x in network_traffic if abs((x - mean) / std_dev) > threshold]
print(f"Anomalous network traffic: {anomalous_traffic}")
\end{lstlisting}

Network intrusion detection systems (NIDS) often rely on real-time anomaly detection to ensure the security and stability of an organization's network \cite{mukherjee1994network}.

\begin{center}
\begin{tikzpicture}[
    level distance=2.5cm,  
    sibling distance=4.0cm, 
    every node/.style={align=center},
    grow=down,
    level 1/.style={sibling distance=5.5cm}, 
    level 2/.style={sibling distance=2.5cm}  
]
    \node {Outlier Detection Methods}
    child {node {Statistical Methods}
        child {node {Z-score}}
        child {node {Tukey's Fences}}
    }
    child {node {Distance-based Methods}
        child {node {k-NN}}
        child {node {Local Outlier Factor}}
    }
    child {node {Density-based Methods}
        child {node {DBSCAN}}
        child {node {OPTICS}}
    };
\end{tikzpicture}
\end{center}

\chapter{Text Analytics and Information Retrieval}
    
    \section{Introduction to Text Data}
    Text data refers to unstructured data that is made up of words, sentences, and documents. Examples of text data include news articles, customer reviews, emails, social media posts, etc. Unlike structured data like numerical values or tables, text data lacks a predefined structure, which makes it more challenging to process and analyze directly \cite{jurafsky2000speech}.

    In text analytics, our goal is to extract meaningful insights from this text data, such as identifying patterns, classifying documents, or understanding the sentiment of a text. To achieve this, we must first convert the text into a format that machines can understand. This process involves several techniques, which we will cover in this chapter, such as the Bag of Words model, preprocessing text, and using vector space models \cite{jurafsky2000speech}.

    \section{Bag of Words Model}
    The Bag of Words (BoW) model is one of the most basic techniques used to represent text data. It works by representing a text document as a collection of words, disregarding grammar and word order. The idea is to create a vocabulary of all the unique words in the text dataset and then represent each document based on the words it contains \cite{zhang2010understanding}.

    For example, consider the following two sentences:
    \begin{itemize}
        \item Sentence 1: "Python is great for data analysis."
        \item Sentence 2: "I love Python programming."
    \end{itemize}

    The vocabulary from these two sentences would be: \{Python, is, great, for, data, analysis, I, love, programming\}.

    We can then represent each sentence as a vector of word counts:

    \begin{itemize}
        \item Sentence 1: [1, 1, 1, 1, 1, 1, 0, 0, 0]
        \item Sentence 2: [1, 0, 0, 0, 0, 0, 1, 1, 1]
    \end{itemize}

    This vectorization of text allows us to compare documents and perform machine learning tasks on text data.

    \section{Text Preprocessing}
    Text preprocessing is a critical step in preparing raw text data for analysis. Raw text data may contain unnecessary information such as punctuation, special characters, and stopwords (common words like "the", "is", "in") that do not contribute much to the meaning of the text. The goal of text preprocessing is to clean and normalize the text \cite{hull1996stemming}.

    Common text preprocessing steps include:
    \begin{itemize}
        \item Lowercasing
        \item Removing punctuation
        \item Stopword removal
        \item Stemming and Lemmatization
    \end{itemize}

    Let's go through these in detail.

    \subsection{Stopword Removal}
    Stopwords are words that occur very frequently in a language but carry little meaningful information. Examples of stopwords in English include words like "the", "is", "in", "and". Removing stopwords helps to reduce the dimensionality of the text data and improve the performance of text analysis algorithms \cite{kaur2018systematic}.

    In Python, we can remove stopwords using the `nltk` library. Here's an example:

    \begin{lstlisting}[style=python]
    import nltk
    from nltk.corpus import stopwords
    from nltk.tokenize import word_tokenize

    nltk.download('stopwords')
    nltk.download('punkt')

    text = "Python is great for data analysis"
    stop_words = set(stopwords.words('english'))

    word_tokens = word_tokenize(text)

    filtered_sentence = [word for word in word_tokens if not word in stop_words]

    print(filtered_sentence)
    # Output: ['Python', 'great', 'data', 'analysis']
    \end{lstlisting}

    In this example, we use the `nltk` library to tokenize the sentence and filter out stopwords, leaving only the meaningful words in the text.

    \subsection{Stemming and Lemmatization}
    Stemming and Lemmatization are techniques used to reduce words to their base or root form. The main difference between the two is that stemming is a rule-based process that cuts off word endings, while lemmatization takes into account the context and converts words into their base form based on their meaning \cite{korenius2004stemming}.

    \textbf{Stemming Example:}
    \begin{itemize}
        \item "running" -> "run"
        \item "studies" -> "studi"
    \end{itemize}

    \textbf{Lemmatization Example:}
    \begin{itemize}
        \item "running" -> "run"
        \item "studies" -> "study"
    \end{itemize}

    Let's see an example of how both can be applied in Python:

    \begin{lstlisting}[style=python]
    from nltk.stem import PorterStemmer
    from nltk.stem import WordNetLemmatizer
    from nltk.tokenize import word_tokenize
    import nltk

    nltk.download('wordnet')
    nltk.download('omw-1.4')

    # Example text
    text = "running runs runner studied studying"

    # Stemming
    ps = PorterStemmer()
    stemmed_words = [ps.stem(word) for word in word_tokenize(text)]
    print("Stemmed:", stemmed_words)

    # Lemmatization
    lemmatizer = WordNetLemmatizer()
    lemmatized_words = [lemmatizer.lemmatize(word) for word in word_tokenize(text)]
    print("Lemmatized:", lemmatized_words)
    \end{lstlisting}

    The output will show the difference between stemming and lemmatization. Stemming is more aggressive and might produce non-dictionary words, while lemmatization is more sophisticated, producing valid base forms.

    \section{Text Representation and Vector Space Model}
    Once the text has been preprocessed, we need a way to represent the text numerically so that algorithms can work with it. One of the most common ways to represent text data is using a Vector Space Model (VSM). In this model, documents are represented as vectors in a high-dimensional space, where each dimension corresponds to a word in the vocabulary \cite{oard1998comparative}.

    The simplest form of VSM is the Bag of Words model, which we discussed earlier. However, not all words carry equal importance, and words that appear frequently across many documents (like stopwords) should have less weight compared to rare but important words.

    \subsection{TF-IDF and Term Weighting}
    Term Frequency-Inverse Document Frequency (TF-IDF) is a statistical measure used to evaluate how important a word is to a document relative to a collection of documents (the corpus). It is a common weighting technique used to prioritize important words while downplaying frequent but less informative words \cite{wu2008interpreting}.

    The formula for TF-IDF is as follows:

    \[
    \text{TF-IDF}(t, d) = \text{TF}(t, d) \times \text{IDF}(t)
    \]

    Where:
    \begin{itemize}
        \item \textbf{TF(t, d)}: Term Frequency, the number of times the term \(t\) appears in document \(d\).
        \item \textbf{IDF(t)}: Inverse Document Frequency, calculated as \( \log(\frac{N}{1 + \text{df}(t)}) \), where \(N\) is the total number of documents and \(\text{df}(t)\) is the number of documents containing term \(t\).
    \end{itemize}

    In Python, we can use `TfidfVectorizer` from the `sklearn` library to compute TF-IDF scores:

    \begin{lstlisting}[style=python]
    from sklearn.feature_extraction.text import TfidfVectorizer

    # Example corpus
    corpus = [
        'Python is great for data analysis',
        'I love Python programming',
        'Data analysis is fun'
    ]

    vectorizer = TfidfVectorizer()
    X = vectorizer.fit_transform(corpus)

    # Display the TF-IDF matrix
    print(X.toarray())
    print(vectorizer.get_feature_names_out())
    \end{lstlisting}

    This will output a TF-IDF matrix where each row represents a document, and each column corresponds to a word from the vocabulary. The values represent the TF-IDF scores of the words in the documents.

    \subsection{Cosine Similarity}
    Cosine similarity is a measure used to calculate the similarity between two documents based on their vector representation. It measures the cosine of the angle between two vectors, with values ranging from -1 to 1. If two documents are identical, the cosine similarity will be 1 \cite{rahutomo2012semantic}.

    The formula for cosine similarity is:

    \[
    \text{Cosine Similarity}(A, B) = \frac{A \cdot B}{\|A\| \|B\|}
    \]

    In Python, we can compute cosine similarity using the `cosine\_similarity` function from `sklearn`:

    \begin{lstlisting}[style=python]
    from sklearn.metrics.pairwise import cosine_similarity

    # Compute cosine similarity between documents
    cosine_sim = cosine_similarity(X)
    print(cosine_sim)
    \end{lstlisting}

    This will give us a similarity matrix where each value represents the cosine similarity between two documents.

    \section{Boolean Retrieval Model}
    The Boolean Retrieval Model is one of the simplest forms of information retrieval. It allows users to specify queries using Boolean logic (AND, OR, NOT) to retrieve documents that exactly match the query terms. In this model, each document is represented as a binary vector, where each dimension corresponds to the presence (1) or absence (0) of a term \cite{lee1993evaluation}.

    For example, if we have a document represented by the words \{Python, data, analysis\}, and a query is "Python AND analysis", we retrieve this document because it contains both terms.

    \section{Sentiment Analysis}
    Sentiment analysis is the process of determining the emotional tone behind a body of text. It is used to understand opinions, emotions, and attitudes expressed in text data. Sentiment analysis is widely used in social media monitoring, customer feedback analysis, and product reviews \cite{medhat2014sentiment}.

    \subsection{Lexicon-based Methods}
    Lexicon-based sentiment analysis involves using a predefined list of words, each associated with a specific sentiment score (positive, negative, neutral). A document's overall sentiment is determined by summing the sentiment scores of the words it contains \cite{taboada2011lexicon}.

    One popular lexicon for sentiment analysis is the `VADER` lexicon, which is available in the `nltk` library:

    \begin{lstlisting}[style=python]
    from nltk.sentiment import SentimentIntensityAnalyzer
    import nltk

    nltk.download('vader_lexicon')

    sia = SentimentIntensityAnalyzer()

    text = "Python is amazing for data analysis!"
    sentiment = sia.polarity_scores(text)
    print(sentiment)
    \end{lstlisting}

    This will output a dictionary with sentiment scores for positive, negative, neutral, and compound sentiment.

    \subsection{Machine Learning Approaches for Sentiment Analysis}
    Machine learning approaches for sentiment analysis involve training models on labeled datasets where each text is associated with a sentiment label (positive, negative, or neutral). These models can then predict the sentiment of new, unseen text \cite{ren2024deeplearningmachinelearning}.

    One common approach is to use a classification algorithm such as Naive Bayes, Support Vector Machines (SVM), or Logistic Regression. In Python, we can use the `sklearn` library to train a sentiment classifier.

    Here's an example using a Naive Bayes classifier:

    \begin{lstlisting}[style=python]
    from sklearn.model_selection import train_test_split
    from sklearn.feature_extraction.text import CountVectorizer
    from sklearn.naive_bayes import MultinomialNB
    from sklearn.metrics import accuracy_score

    # Example dataset
    texts = ['I love programming', 'Python is terrible', 'I enjoy learning Python', 'This is awful']
    labels = [1, 0, 1, 0]  # 1 for positive, 0 for negative

    # Text vectorization
    vectorizer = CountVectorizer()
    X = vectorizer.fit_transform(texts)

    # Split data into training and testing sets
    X_train, X_test, y_train, y_test = train_test_split(X, labels, test_size=0.3, random_state=42)

    # Train Naive Bayes classifier
    model = MultinomialNB()
    model.fit(X_train, y_train)

    # Make predictions
    y_pred = model.predict(X_test)

    # Calculate accuracy
    accuracy = accuracy_score(y_test, y_pred)
    print(f'Accuracy: {accuracy}')
    \end{lstlisting}

    This example demonstrates how to train a simple Naive Bayes classifier for sentiment analysis using a small dataset. In practice, larger labeled datasets are used for better accuracy.

\chapter{Model Evaluation and Validation}

\section{Model Performance Metrics}

When we build a machine learning model, evaluating its performance is crucial to ensure it works as expected, especially when dealing with unseen data. There are several metrics to assess model performance, including accuracy, precision, recall, and F1-score. Let's discuss each of them with simple examples \cite{bishop2006pattern}.

\subsection{Accuracy, Precision, Recall, and F1-Score}

\subsubsection{Accuracy}

Accuracy is one of the most straightforward evaluation metrics. It measures how often the model correctly classifies the data. The formula for accuracy is:

\[
\text{Accuracy} = \frac{\text{Number of correct predictions}}{\text{Total number of predictions}}
\]

Let's take a classification problem where our model predicts whether an email is spam or not. If the model classifies 90 emails correctly out of 100, the accuracy would be:

\[
\text{Accuracy} = \frac{90}{100} = 0.9 = 90\%
\]

While accuracy seems like a good measure, it may not always be the best choice when we have imbalanced data. For example, if 95\% of emails are non-spam, a model that predicts "non-spam" for every email would still have high accuracy but would fail to catch spam emails.

\begin{lstlisting}[style=python]
from sklearn.metrics import accuracy_score

# Example in Python
y_true = [0, 1, 0, 1, 0, 1, 0, 0, 1, 0]  # True labels
y_pred = [0, 0, 0, 1, 0, 1, 0, 0, 0, 1]  # Predicted labels

accuracy = accuracy_score(y_true, y_pred)
print(f"Accuracy: {accuracy}")
\end{lstlisting}

\subsubsection{Precision}

Precision measures how many of the predicted positive classes were actually correct. It's important when the cost of false positives is high, such as in spam detection. The formula for precision is:

\[
\text{Precision} = \frac{\text{True Positives}}{\text{True Positives} + \text{False Positives}}
\]

For example, if your model predicts 30 emails as spam, and 20 of them are actually spam, the precision would be:

\[
\text{Precision} = \frac{20}{20 + 10} = 0.67 = 67\%
\]

\begin{lstlisting}[style=python]
from sklearn.metrics import precision_score

precision = precision_score(y_true, y_pred)
print(f"Precision: {precision}")
\end{lstlisting}

\subsubsection{Recall}

Recall, also known as sensitivity or true positive rate, measures how many actual positives were correctly predicted. It's important when missing positive cases (false negatives) is costly. The formula for recall is:

\[
\text{Recall} = \frac{\text{True Positives}}{\text{True Positives} + \text{False Negatives}}
\]

For instance, if there are 25 actual spam emails, and your model correctly predicts 20 of them as spam, the recall would be:

\[
\text{Recall} = \frac{20}{20 + 5} = 0.80 = 80\%
\]

\begin{lstlisting}[style=python]
from sklearn.metrics import recall_score

recall = recall_score(y_true, y_pred)
print(f"Recall: {recall}")
\end{lstlisting}

\subsubsection{F1-Score}

F1-score is the harmonic mean of precision and recall. It balances the two when one metric alone isn't enough to evaluate the model performance. The formula for F1-score is:

\[
\text{F1} = 2 \times \frac{\text{Precision} \times \text{Recall}}{\text{Precision} + \text{Recall}}
\]

The F1-score is particularly useful in situations where we need to find an equilibrium between precision and recall, such as fraud detection or medical diagnosis.

\begin{lstlisting}[style=python]
from sklearn.metrics import f1_score

f1 = f1_score(y_true, y_pred)
print(f"F1-Score: {f1}")
\end{lstlisting}

\subsection{ROC Curves and AUC}

The Receiver Operating Characteristic (ROC) curve is a graphical representation of a classifier's performance. It plots the true positive rate (recall) against the false positive rate (1 - specificity). The Area Under the Curve (AUC) is a single number that summarizes the performance of the classifier. The higher the AUC, the better the model is at distinguishing between classes.

For example, an AUC of 0.5 indicates random performance, while an AUC of 1.0 means perfect classification.

\begin{lstlisting}[style=python]
from sklearn.metrics import roc_curve, auc
import matplotlib.pyplot as plt

# Generate ROC curve
fpr, tpr, thresholds = roc_curve(y_true, y_pred)
roc_auc = auc(fpr, tpr)

# Plot the ROC curve
plt.figure()
plt.plot(fpr, tpr, color='darkorange', lw=2, label=f'ROC curve (area = {roc_auc:.2f})')
plt.plot([0, 1], [0, 1], color='navy', lw=2, linestyle='--')
plt.xlim([0.0, 1.0])
plt.ylim([0.0, 1.05])
plt.xlabel('False Positive Rate')
plt.ylabel('True Positive Rate')
plt.title('Receiver Operating Characteristic')
plt.legend(loc='lower right')
plt.show()
\end{lstlisting}

\section{Confusion Matrix and Cost-sensitive Learning}

\subsection{Confusion Matrix}

A confusion matrix provides a comprehensive view of the model's performance by showing the correct and incorrect predictions for each class. It's a 2x2 table for binary classification problems, but it can be extended to more classes. Here's the structure of a confusion matrix:

\[
\begin{tabular}{|c|c|c|}
\hline
 & Predicted Positive & Predicted Negative \\
\hline
Actual Positive & True Positive (TP) & False Negative (FN) \\
\hline
Actual Negative & False Positive (FP) & True Negative (TN) \\
\hline
\end{tabular}
\]

This helps us evaluate metrics like accuracy, precision, recall, and F1-score by simply reading values from the matrix.

\begin{lstlisting}[style=python]
from sklearn.metrics import confusion_matrix

cm = confusion_matrix(y_true, y_pred)
print(f"Confusion Matrix:\n {cm}")
\end{lstlisting}

\subsection{Cost-sensitive Learning}

In many cases, the cost of making different types of errors (false positives vs. false negatives) can be very different. Cost-sensitive learning involves assigning different weights to these errors during model training. This is especially useful in imbalanced datasets, where one class significantly outweighs the other.

\section{Cross-validation Techniques}

Cross-validation is a statistical method used to estimate the skill of a model on unseen data. The idea is to split the data into multiple parts, train the model on one part, and test it on another. This ensures the model is generalized and not overfitted to the training data.

\subsection{K-fold Cross-validation}

In K-fold cross-validation, the dataset is split into \( K \) subsets (or "folds"). The model is trained on \( K-1 \) folds and tested on the remaining fold. This process is repeated \( K \) times, and the final performance is the average of all folds. A common choice for \( K \) is 5 or 10.

\begin{lstlisting}[style=python]
from sklearn.model_selection import KFold, cross_val_score
from sklearn.ensemble import RandomForestClassifier

# Example: 5-fold cross-validation
kf = KFold(n_splits=5)
model = RandomForestClassifier()

scores = cross_val_score(model, X, y, cv=kf)
print(f"Cross-validation scores: {scores}")
\end{lstlisting}

\subsection{Leave-One-Out Cross-validation}

Leave-One-Out Cross-validation (LOO-CV) is a special case of K-fold cross-validation where \( K \) equals the number of data points in the dataset. Each observation is used as a test set once, and the model is trained on the remaining data. This method is more computationally expensive but useful for smaller datasets.

\begin{lstlisting}[style=python]
from sklearn.model_selection import LeaveOneOut

# Example: Leave-One-Out Cross-validation
loo = LeaveOneOut()

scores = cross_val_score(model, X, y, cv=loo)
print(f"LOO Cross-validation scores: {scores}")
\end{lstlisting}

\section{Bootstrapping Methods for Model Validation}

Bootstrapping is a resampling technique used to estimate the accuracy of a model by generating new datasets by sampling with replacement. It is particularly useful when the dataset is small and cross-validation may not give reliable estimates.

\begin{lstlisting}[style=python]
from sklearn.utils import resample

# Example of Bootstrapping
X_resampled, y_resampled = resample(X, y, n_samples=len(X), replace=True)

model.fit(X_resampled, y_resampled)
\end{lstlisting}

Bootstrapping allows for creating several different samples of the data, and the model's performance can be averaged across these samples for a robust estimate of its ability to generalize.

\chapter{Time Series Analysis and Forecasting}
    
    \section{Introduction to Time Series Data}
    
    A \textbf{time series} is a sequence of data points typically measured at successive times, spaced at uniform time intervals. In real-world scenarios, time series data occurs frequently across various domains such as economics, finance, weather forecasting, and stock market analysis. Examples of time series data include daily stock prices, annual sales figures, or monthly temperature measurements.

    Time series analysis aims to understand the underlying structure and pattern in the data and develop models that can predict future values. Unlike standard regression analysis, which assumes that observations are independent of each other, time series data often exhibits serial dependence, where observations at one point in time are related to observations at other points in time \cite{box2015time, brockwell2002introduction}.

    Some common uses of time series analysis include:
    
    \begin{itemize}
        \item Forecasting future values (e.g., stock prices, sales forecasting)
        \item Identifying trends or seasonal patterns
        \item Decomposing the time series into its components to understand its structure
        \item Evaluating the performance of predictive models using residual analysis
    \end{itemize}
    
    In this chapter, we will explore various techniques to analyze and forecast time series data using Python.
    
    \section{Components of Time Series}
    
    A time series can typically be broken down into several components that help in understanding its behavior. The main components of a time series are:
    
    \subsection{Trend, Seasonal, and Cyclical Components}
    
    \begin{itemize}
        \item \textbf{Trend (T):} A long-term increase or decrease in the data. It represents the general direction in which the data is moving over a long period.
        \item \textbf{Seasonality (S):} A repeating pattern in the data that occurs at regular intervals due to seasonal factors (e.g., quarterly sales, monthly temperature).
        \item \textbf{Cyclical (C):} Long-term fluctuations in the data that are not regular, often related to economic or business cycles.
    \end{itemize}

    The mathematical representation of a time series with these components can be either additive or multiplicative:
    
    \begin{itemize}
        \item \textbf{Additive model:} $Y_t = T_t + S_t + C_t + e_t$
        \item \textbf{Multiplicative model:} $Y_t = T_t \times S_t \times C_t \times e_t$
    \end{itemize}
    
    \section{Smoothing Techniques}
    
    Smoothing techniques help reduce the noise in a time series to better reveal the underlying patterns. We will discuss two popular smoothing methods: Moving Average and Exponential Smoothing.
    
    \subsection{Moving Average}
    
    The moving average is a simple technique that calculates the average of a fixed number of consecutive observations. It helps smooth short-term fluctuations and highlight longer-term trends or cycles.
    
    The formula for a simple moving average is:
    
    \[
    \text{SMA}_t = \frac{Y_t + Y_{t-1} + Y_{t-2} + \dots + Y_{t-(n-1)}}{n}
    \]
    
    where $n$ is the number of observations used in the moving average.

    Example Python code for calculating a moving average:

    \begin{lstlisting}[style=python]
import pandas as pd

# Example time series data
data = {'Month': ['Jan', 'Feb', 'Mar', 'Apr', 'May', 'Jun'],
        'Sales': [200, 220, 250, 230, 270, 290]}
df = pd.DataFrame(data)

# Calculate a 3-month moving average
df['Moving_Avg'] = df['Sales'].rolling(window=3).mean()

print(df)
    \end{lstlisting}
    
    In this code, we use the \texttt{rolling()} function in Python’s pandas library to compute the moving average. This function takes a window size (3 in this case) and calculates the moving average for the sales data.
    
    \subsection{Exponential Smoothing}
    
    Exponential smoothing assigns exponentially decreasing weights to past observations. This means that more recent observations have a higher weight than older ones. The formula for single exponential smoothing is:

    \[
    \hat{Y}_t = \alpha Y_{t-1} + (1 - \alpha)\hat{Y}_{t-1}
    \]

    where $\alpha$ is the smoothing constant ($0 < \alpha < 1$) that controls how much weight is given to the most recent observation.

    Example Python code for exponential smoothing:

    \begin{lstlisting}[style=python]
import pandas as pd
from statsmodels.tsa.holtwinters import SimpleExpSmoothing

# Example time series data
data = {'Month': ['Jan', 'Feb', 'Mar', 'Apr', 'May', 'Jun'],
        'Sales': [200, 220, 250, 230, 270, 290]}
df = pd.DataFrame(data)

# Fit the Exponential Smoothing model
model = SimpleExpSmoothing(df['Sales']).fit(smoothing_level=0.2)
df['Exp_Smoothing'] = model.fittedvalues

print(df)
    \end{lstlisting}
    
    Here, we use the \texttt{SimpleExpSmoothing} class from the \texttt{statsmodels} library to fit the exponential smoothing model. The parameter \texttt{smoothing\_level} is set to 0.2, which controls the weight assigned to the most recent observations.
    
    \section{Time Series Regression Models}
    
    Regression models can be used to model time series data by treating time as an independent variable. The simplest form of time series regression is a linear trend model, where the dependent variable is modeled as a linear function of time:
    
    \[
    Y_t = \beta_0 + \beta_1 t + e_t
    \]
    
    Example Python code for a time series regression model:

    \begin{lstlisting}[style=python]
import numpy as np
import pandas as pd
from sklearn.linear_model import LinearRegression

# Example time series data
data = {'Month': ['Jan', 'Feb', 'Mar', 'Apr', 'May', 'Jun'],
        'Sales': [200, 220, 250, 230, 270, 290]}
df = pd.DataFrame(data)

# Create a time variable
df['Time'] = np.arange(len(df))

# Fit a linear regression model
X = df[['Time']]
y = df['Sales']
model = LinearRegression().fit(X, y)

# Predict future values
df['Sales_Predicted'] = model.predict(X)

print(df)
    \end{lstlisting}
    
    In this example, we use Python’s \texttt{sklearn.linear\_model.LinearRegression} to fit a linear regression model to the sales data, with time as the independent variable.

    \section{Autoregressive (AR) and ARMA Models}
    
    \textbf{Autoregressive (AR) models} model the current value of a time series based on its previous values. The AR model of order $p$ is denoted as AR(p) and is represented as:

    \[
    Y_t = c + \phi_1 Y_{t-1} + \phi_2 Y_{t-2} + \dots + \phi_p Y_{t-p} + e_t
    \]

    where $c$ is a constant, $\phi_i$ are the autoregressive coefficients, and $e_t$ is white noise.
    
    The \textbf{ARMA} (Autoregressive Moving Average) model combines AR and Moving Average (MA) models. The ARMA(p, q) model is given by:

    \[
    Y_t = c + \phi_1 Y_{t-1} + \dots + \phi_p Y_{t-p} + \theta_1 e_{t-1} + \dots + \theta_q e_{t-q} + e_t
    \]

    Example Python code for fitting an ARMA model:

    \begin{lstlisting}[style=python]
from statsmodels.tsa.arima.model import ARIMA

# Fit an ARMA model (AR=1, MA=1)
model = ARIMA(df['Sales'], order=(1, 0, 1)).fit()

# Forecast future values
df['ARMA_Forecast'] = model.fittedvalues

print(df)
    \end{lstlisting}
    
    \section{Residual Analysis and Model Evaluation}
    
    After fitting a model to a time series, it's important to evaluate how well the model captures the underlying patterns in the data. One way to do this is through residual analysis, which involves analyzing the difference between the actual and predicted values:
    
    \[
    \text{Residual} = Y_t - \hat{Y}_t
    \]
    
    A good model should have residuals that are random (i.e., no discernible pattern), with a mean close to zero and no autocorrelation.

    Example Python code for residual analysis:

    \begin{lstlisting}[style=python]
import matplotlib.pyplot as plt

# Calculate residuals
df['Residuals'] = df['Sales'] - df['Sales_Predicted']

# Plot residuals
plt.figure(figsize=(10, 6))
plt.plot(df['Month'], df['Residuals'], marker='o')
plt.title('Residual Analysis')
plt.xlabel('Month')
plt.ylabel('Residuals')
plt.show()
    \end{lstlisting}
    
    In this code, we plot the residuals to visually inspect whether there are any patterns in the residuals. Ideally, the residuals should fluctuate randomly around zero, indicating that the model has successfully captured the structure in the time series data.

\chapter{Recommender Systems}
\section{Introduction to Recommender Systems}

Recommender systems are a subclass of information filtering systems that seek to predict the preferences or ratings a user might give to an item. These systems have become essential components of many online platforms, including e-commerce sites, streaming services, and social media platforms, where personalized recommendations play a critical role in improving user satisfaction and engagement \cite{jannach2010recommender}.

The goal of a recommender system is to filter and present only the most relevant content to a user from a large pool of options. For instance, Netflix recommends movies and TV shows based on your past viewing behavior, while Amazon suggests products you might be interested in purchasing.

There are several types of recommender systems:
\begin{itemize}
    \item \textbf{Collaborative Filtering:} This method uses the behavior of multiple users to make recommendations, assuming that if users agreed on past interactions, they will agree on future preferences.
    \item \textbf{Content-based Filtering:} This technique analyzes the features of items and recommends those with similar characteristics to what the user liked in the past.
    \item \textbf{Hybrid Methods:} These systems combine collaborative and content-based methods to provide more accurate and personalized recommendations.
\end{itemize}

In the following sections, we will explore these methods in more detail, starting with collaborative filtering, then moving on to content-based systems, and finally discussing hybrid approaches.

\section{Collaborative Filtering Methods}

Collaborative filtering (CF) is one of the most widely used techniques in recommender systems. The underlying principle is simple: similar users will like similar items \cite{koren2021advances}. CF is divided into two major types:
\begin{itemize}
    \item \textbf{User-User Collaborative Filtering:} This method focuses on finding similarities between users to make recommendations.
    \item \textbf{Item-Item Collaborative Filtering:} This method looks at the similarities between items and suggests items that are similar to what the user has previously interacted with.
\end{itemize}

We will now explore these methods in more detail.

\subsection{User-User Collaborative Filtering}

In user-user collaborative filtering, the system recommends items to a user by finding other users with similar tastes or preferences. For example, if User A and User B have both liked a set of movies, the system assumes they share similar tastes and can recommend a movie that User B liked to User A \cite{breese2013empirical}.

\textbf{Steps:}
\begin{enumerate}
    \item Compute the similarity between users based on their ratings or interactions with items.
    \item Identify the most similar users (neighbors) to the target user.
    \item Recommend items that the neighbors have liked but the target user has not yet interacted with.
\end{enumerate}

\textbf{Example:}
Consider the following user-item matrix, where each entry represents the rating given by a user to an item:

\begin{center}
\begin{tabular}{ |c|c|c|c|c| }
 \hline
 User & Item A & Item B & Item C & Item D \\
 \hline
 User 1 & 5 & 3 & 0 & 1 \\
 User 2 & 4 & 0 & 4 & 1 \\
 User 3 & 2 & 3 & 5 & 0 \\
 User 4 & 0 & 4 & 4 & 0 \\
 \hline
\end{tabular}
\end{center}

In this case, we want to recommend an item for User 1. Based on the ratings of the other users, we calculate the similarity between User 1 and the others, and recommend items that the most similar user (say User 2) has rated highly but User 1 has not yet rated.

A common method to calculate similarity is the cosine similarity or Pearson correlation.

\textbf{Python Code Example:}
\begin{lstlisting}[style=python]
from sklearn.metrics.pairwise import cosine_similarity
import numpy as np

# User-Item matrix
user_item_matrix = np.array([
    [5, 3, 0, 1],
    [4, 0, 4, 1],
    [2, 3, 5, 0],
    [0, 4, 4, 0]
])

# Calculate cosine similarity between users
user_similarity = cosine_similarity(user_item_matrix)

# Output similarity matrix
print(user_similarity)
\end{lstlisting}

\subsection{Item-Item Collaborative Filtering}

Item-item collaborative filtering is similar to user-user collaborative filtering, but instead of finding similar users, it finds similar items. If a user has rated an item highly, the system recommends items that are similar to it \cite{breese2013empirical}.

\textbf{Steps:}
\begin{enumerate}
    \item Compute the similarity between items based on user ratings.
    \item For a given user, find items similar to those the user has already rated highly.
    \item Recommend those similar items to the user.
\end{enumerate}

\textbf{Example:}
Using the same user-item matrix from the previous example, we can calculate the similarity between items instead of users and make recommendations based on the items the user has liked.

\textbf{Python Code Example:}
\begin{lstlisting}[style=python]
# Transpose user-item matrix to get item-user matrix
item_user_matrix = user_item_matrix.T

# Calculate cosine similarity between items
item_similarity = cosine_similarity(item_user_matrix)

# Output similarity matrix
print(item_similarity)
\end{lstlisting}

\section{Content-based Recommender Systems}

Content-based recommender systems focus on analyzing the characteristics (features) of items and making recommendations based on a user's past preferences. The system builds a profile of each user based on the features of items they have interacted with \cite{lops2011content}.

\subsection{Item Profiles and Feature Extraction}

In content-based filtering, items are described by a set of attributes or features. For example, in a movie recommender system, features could include genre, director, cast, and keywords \cite{lops2011content}.

The recommender system needs to extract these features from the items and build a profile of each item.

\textbf{Example:}
Consider a movie recommender system where the features of a movie could include:
\begin{itemize}
    \item Genre: Action, Comedy, Drama, etc.
    \item Director: Steven Spielberg, Christopher Nolan, etc.
    \item Cast: Actor names.
    \item Keywords: Specific terms associated with the movie (e.g., "space," "adventure," "robot").
\end{itemize}

\textbf{Python Code Example:}
\begin{lstlisting}[style=python]
from sklearn.feature_extraction.text import TfidfVectorizer

# Example item descriptions (could be movie descriptions)
item_descriptions = [
    "Action movie with robots and spaceships",
    "Romantic comedy with lots of humor",
    "Drama about family and relationships"
]

# Convert the item descriptions into TF-IDF feature vectors
vectorizer = TfidfVectorizer()
item_profiles = vectorizer.fit_transform(item_descriptions)

# Output the feature vectors
print(item_profiles.toarray())
\end{lstlisting}

\subsection{User Profiles and Preference Learning}

Once the item profiles are built, the system creates a user profile by analyzing the items that the user has liked or interacted with. The user profile is a weighted combination of the features of those items.

\textbf{Example:}
If a user has watched two movies, one action movie with robots and another science fiction movie with spaceships, the user's profile would indicate a preference for action and science fiction genres, along with keywords like "robots" and "spaceships."

\textbf{Python Code Example:}
\begin{lstlisting}[style=python]
# Example user interaction with items (1 if user liked, 0 if not)
user_interactions = np.array([1, 0, 1])

# Calculate user profile as the weighted sum of item profiles
user_profile = np.dot(user_interactions, item_profiles.toarray())

# Output the user profile
print(user_profile)
\end{lstlisting}

\section{Hybrid Recommender Systems}

Hybrid recommender systems combine collaborative filtering and content-based methods to leverage the strengths of both approaches. These systems can provide more accurate and personalized recommendations by using both user behavior and item features \cite{burke2002hybrid}.

\subsection{Combining Collaborative and Content-based Approaches}

There are several ways to combine collaborative filtering and content-based methods:
\begin{itemize}
    \item \textbf{Weighted hybrid:} Combine the recommendations from both systems by assigning different weights to each method.
    \item \textbf{Switching hybrid:} Switch between methods depending on the situation, such as using content-based filtering for new users and collaborative filtering for experienced users.
    \item \textbf{Feature augmentation:} Use one method to enhance the input to the other method, such as using content features to improve the collaborative filtering process.
\end{itemize}

\section{Evaluation of Recommender Systems}

To evaluate the performance of recommender systems, we use several metrics that assess their accuracy and effectiveness.

\subsection{Precision, Recall, and F-Measure}

\textbf{Precision} measures the proportion of relevant items in the recommended set. \textbf{Recall} measures the proportion of relevant items that were successfully recommended. The \textbf{F-Measure} is the harmonic mean of precision and recall.

\subsection{ROC Curve and Ranking Metrics}

The \textbf{ROC Curve} is used to visualize the trade-off between true positive rate (recall) and false positive rate. Ranking metrics like \textbf{Mean Average Precision (MAP)} and \textbf{Normalized Discounted Cumulative Gain (NDCG)} measure how well the system ranks the relevant items higher in the recommendation list.

\chapter{Advanced Techniques in Big Data Analytics}

\section{Introduction to Deep Learning}
Deep Learning is a subset of machine learning that deals with algorithms inspired by the structure and function of the brain called artificial neural networks. The concept of deep learning revolves around building and training neural networks that consist of many layers (hence "deep"). These neural networks are used to solve complex problems such as image recognition, speech processing, and natural language understanding. In the context of big data analytics, deep learning techniques can analyze large datasets in an efficient manner, automatically extracting useful features and patterns from the data \cite{chen2024deeplearningmachinelearning, feng2024deeplearningmachinelearning}.

\subsection{What is a Neural Network?}
A neural network consists of a collection of connected nodes or neurons organized in layers: input layer, hidden layers, and output layer. Each connection between neurons is assigned a weight, and each neuron has an activation function. During training, the network adjusts these weights in order to reduce the error in predictions.

\textbf{Example:}
Suppose you want to classify images of handwritten digits (0–9) from the MNIST dataset using a deep neural network. Each image is 28x28 pixels, resulting in 784 input features (one for each pixel). A simple neural network would consist of:

\begin{center}
\begin{tikzpicture}[
    node distance=3cm and 2cm, 
    every node/.style={align=center} 
]
  \node (input) at (0,0) {Input Layer\\(784 units)};
  \node (hidden1) at (4,-2) {Hidden Layer 1\\(256 units)};
  \node (hidden2) at (8,-2) {Hidden Layer 2\\(128 units)};
  \node (output) at (12,0) {Output Layer\\(10 units)};
  
  \draw[->] (input) -- (hidden1);
  \draw[->] (hidden1) -- (hidden2);
  \draw[->] (hidden2) -- (output);
\end{tikzpicture}
\end{center}

This is a simple neural network architecture where each layer is fully connected to the next. The final output layer has 10 units representing the probability for each of the 10 classes (digits 0–9).

In Python, neural networks can be implemented using the popular library \texttt{Keras}:

\begin{lstlisting}[style=python]
from keras.models import Sequential
from keras.layers import Dense

# Initialize the model
model = Sequential()

# Input layer (784 units) and first hidden layer (256 units)
model.add(Dense(256, input_dim=784, activation='relu'))

# Second hidden layer (128 units)
model.add(Dense(128, activation='relu'))

# Output layer (10 units for 10 classes)
model.add(Dense(10, activation='softmax'))

# Compile the model
model.compile(loss='categorical_crossentropy', optimizer='adam', metrics=['accuracy'])

# View model summary
model.summary()
\end{lstlisting}

This code snippet defines a neural network with two hidden layers and an output layer using the \texttt{Keras} library \cite{li2024deeplearningmachinelearning}.

\section{Convolutional Neural Networks (CNNs)}
Convolutional Neural Networks (CNNs) are a specialized kind of neural network designed for processing structured grid data, such as images. CNNs are widely used for image classification tasks, as they are very effective at capturing spatial features (such as edges and textures) through a series of filters or kernels \cite{krizhevsky2012imagenet}.

\subsection{How CNNs Work}
A CNN typically consists of three types of layers:
\begin{itemize}
  \item \textbf{Convolutional Layer:} This layer applies filters to the input image, detecting features such as edges, corners, and textures.
  \item \textbf{Pooling Layer:} This layer reduces the spatial dimensions of the image, making the computation more efficient while retaining important features.
  \item \textbf{Fully Connected Layer:} This layer is similar to the layers in a regular neural network and is used for the final classification.
\end{itemize}

\textbf{Example:}
Consider a CNN for classifying handwritten digits (0–9). The CNN architecture may look like this:

\begin{center}
\begin{tikzpicture}[
    node distance=2cm and 1cm, 
    every node/.style={align=center}
]
  \node (input) at (0,0) {Input Image\\(28x28x1)};
  \node (conv1) at (0,-2) {Convolutional Layer\\(28x28x32)};
  \node (pool1) at (0,-4) {Pooling Layer\\(14x14x32)};
  \node (conv2) at (0,-6) {Convolutional Layer\\(14x14x64)};
  \node (pool2) at (0,-8) {Pooling Layer\\(7x7x64)};
  \node (fc) at (0,-10) {Fully Connected Layer\\(10 units)};
  
  \draw[->] (input) -- (conv1);
  \draw[->] (conv1) -- (pool1);
  \draw[->] (pool1) -- (conv2);
  \draw[->] (conv2) -- (pool2);
  \draw[->] (pool2) -- (fc);
\end{tikzpicture}
\end{center}

In Python, CNNs can also be implemented using \texttt{Keras}:

\begin{lstlisting}[style=python]
from keras.models import Sequential
from keras.layers import Conv2D, MaxPooling2D, Flatten, Dense

# Initialize the model
model = Sequential()

# First convolutional layer with 32 filters, 3x3 kernel, and ReLU activation
model.add(Conv2D(32, kernel_size=(3, 3), activation='relu', input_shape=(28, 28, 1)))

# First pooling layer
model.add(MaxPooling2D(pool_size=(2, 2)))

# Second convolutional layer with 64 filters
model.add(Conv2D(64, kernel_size=(3, 3), activation='relu'))

# Second pooling layer
model.add(MaxPooling2D(pool_size=(2, 2)))

# Flatten the results before the fully connected layer
model.add(Flatten())

# Fully connected layer for output
model.add(Dense(10, activation='softmax'))

# Compile the model
model.compile(loss='categorical_crossentropy', optimizer='adam', metrics=['accuracy'])

# View model summary
model.summary()
\end{lstlisting}

\section{Recurrent Neural Networks (RNNs)}
Recurrent Neural Networks (RNNs) are a type of neural network designed to work with sequential data, such as time series or text. Unlike traditional neural networks, RNNs have connections that loop back, allowing them to maintain a memory of previous inputs. This makes them particularly useful for tasks like speech recognition, language modeling, and stock price prediction \cite{medsker2001recurrent}.

\subsection{How RNNs Work}
RNNs process sequences one element at a time, maintaining a hidden state that is updated at each step. This hidden state allows the network to capture information from previous steps and use it to make better predictions.

\textbf{Example:}
Consider a simple RNN that processes a sequence of numbers. The hidden state is updated after each number is processed, and the final output depends on both the current input and the accumulated hidden state.

\begin{center}
\begin{tikzpicture}
  \node (input1) at (0,0) {Input 1};
  \node (hidden1) at (2,-2) {Hidden State 1};
  \node (input2) at (4,0) {Input 2};
  \node (hidden2) at (6,-2) {Hidden State 2};
  \node (input3) at (8,0) {Input 3};
  \node (hidden3) at (10,-2) {Hidden State 3};
  \draw[->] (input1) -- (hidden1);
  \draw[->] (hidden1) -- (hidden2);
  \draw[->] (input2) -- (hidden2);
  \draw[->] (hidden2) -- (hidden3);
  \draw[->] (input3) -- (hidden3);
\end{tikzpicture}
\end{center}

In Python, RNNs can be implemented using the \texttt{Keras} library:

\begin{lstlisting}[style=python]
from keras.models import Sequential
from keras.layers import SimpleRNN, Dense

# Initialize the model
model = Sequential()

# Add a SimpleRNN layer with 50 units
model.add(SimpleRNN(50, input_shape=(10, 1)))

# Add a fully connected output layer
model.add(Dense(1))

# Compile the model
model.compile(loss='mean_squared_error', optimizer='adam')

# View model summary
model.summary()
\end{lstlisting}

\section{Natural Language Processing (NLP)}
Natural Language Processing (NLP) refers to the application of computational techniques to the analysis and synthesis of natural language and speech. NLP encompasses tasks such as sentiment analysis, language translation, and text summarization \cite{li2024surveyingmllmlandscapemetareview}.

\subsection{Basic NLP Techniques}
Some common NLP techniques include:
\begin{itemize}
  \item \textbf{Tokenization:} Splitting a text into individual words or tokens.
  \item \textbf{Part-of-speech Tagging:} Assigning grammatical categories to each word.
  \item \textbf{Named Entity Recognition (NER):} Identifying names of people, organizations, locations, etc.
  \item \textbf{Sentiment Analysis:} Determining the emotional tone of a text.
\end{itemize}

For basic text processing, Python's \texttt{nltk} and \texttt{spaCy} libraries can be used:

\begin{lstlisting}[style=python]
import spacy

# Load the English NLP model
nlp = spacy.load('en_core_web_sm')

# Process a text
doc = nlp("Apple is looking at buying U.K. startup for $1 billion")

# Tokenization
for token in doc:
    print(token.text, token.pos_, token.dep_)
\end{lstlisting}

\section{MapReduce and Distributed Computing}
MapReduce is a programming model used to process large datasets across distributed computing environments \cite{dean2008mapreduce, li2024deeplearningmachinelearning}. The model breaks down tasks into two functions:
\begin{itemize}
  \item \textbf{Map:} This function processes input data and generates intermediate key-value pairs.
  \item \textbf{Reduce:} This function takes the intermediate key-value pairs and merges them to produce the final output.
\end{itemize}

\textbf{Example:} Consider processing a large dataset of text files to count the occurrence of each word. Using the MapReduce model:
\begin{itemize}
  \item The \textbf{Map} function reads each file, splits it into words, and emits each word along with a count of 1.
  \item The \textbf{Reduce} function takes all emitted word counts, sums them up, and produces the total count for each word.
\end{itemize}

\begin{lstlisting}[style=cmd]
# Map function
def mapper(file):
    for line in file:
        for word in line.split():
            print(f"{word}\t1")

# Reduce function
def reducer(word, counts):
    total = sum(counts)
    print(f"{word}\t{total}")
\end{lstlisting}

\section{Big Data Analytics in the Cloud}
Cloud computing provides an ideal environment for big data analytics by offering scalable storage and processing power on demand. Some of the most common cloud platforms for big data analytics include:
\begin{itemize}
  \item \textbf{Amazon Web Services (AWS):} Provides services like Amazon S3 for storage and Amazon EMR for running big data frameworks like Hadoop.
  \item \textbf{Google Cloud Platform (GCP):} Offers services like Google BigQuery for analyzing large datasets and Google Cloud Dataproc for running Hadoop and Spark jobs.
  \item \textbf{Microsoft Azure:} Provides services like Azure Data Lake and Azure HDInsight for big data processing.
\end{itemize}

Cloud platforms help you run analytics on big data without having to manage physical hardware, allowing for scalability and flexibility.

\chapter{Case Studies and Applications of Big Data}

\section{Big Data in Healthcare}
The healthcare industry has been transformed by the integration of Big Data analytics. In the past, medical data was largely unstructured, fragmented, and stored across different systems, making it challenging to use efficiently. With the advent of Big Data tools and technologies, healthcare providers can now aggregate, process, and analyze vast amounts of patient data, leading to better patient outcomes, more accurate diagnoses, and the identification of emerging health trends \cite{dash2019big, niu2024textmultimodalityexploringevolution}.

\subsection{Predictive Analytics for Patient Care}
One of the most important applications of Big Data in healthcare is predictive analytics. By analyzing historical patient data, doctors and hospitals can predict the likelihood of a patient developing certain conditions such as diabetes or heart disease \cite{niu2024textmultimodalityexploringevolution}.

For example, consider a healthcare dataset containing data points such as patient age, weight, blood pressure, and cholesterol levels. We can use Python and libraries like \texttt{pandas} and \texttt{scikit-learn} to build a machine learning model that predicts the likelihood of heart disease.

\begin{lstlisting}[style=python]
import pandas as pd
from sklearn.model_selection import train_test_split
from sklearn.ensemble import RandomForestClassifier
from sklearn.metrics import accuracy_score

# Load the dataset
data = pd.read_csv('health_data.csv')

# Selecting features and the target variable
X = data[['age', 'weight', 'blood_pressure', 'cholesterol']]
y = data['heart_disease']

# Split the dataset into training and testing sets
X_train, X_test, y_train, y_test = train_test_split(X, y, test_size=0.2)

# Train a Random Forest model
model = RandomForestClassifier()
model.fit(X_train, y_train)

# Make predictions
y_pred = model.predict(X_test)

# Calculate accuracy
accuracy = accuracy_score(y_test, y_pred)
print(f"Model accuracy: {accuracy * 100:.2f}%")
\end{lstlisting}

\subsection{Personalized Medicine}
Big Data allows for more personalized treatment plans. By analyzing patient data in real-time, doctors can tailor treatments based on the individual’s genetic makeup, lifestyle, and medical history. This reduces the risk of ineffective treatment and can improve patient satisfaction \cite{jain2002personalized}.

\section{Big Data in Finance}
The finance industry is another sector that has greatly benefited from Big Data analytics. From risk assessment to fraud detection, Big Data allows financial institutions to make more informed decisions, detect patterns, and respond to market changes faster \cite{subrahmanyam2019big}.

\subsection{Fraud Detection}
Fraud detection is one of the most critical applications of Big Data in the financial sector. By analyzing transaction data in real-time, financial institutions can detect suspicious activities, such as unusual withdrawal patterns or login attempts from unexpected locations \cite{bolton2002statistical}.

Here's an example of how Python can be used for fraud detection using a logistic regression model:

\begin{lstlisting}[style=python]
import pandas as pd
from sklearn.model_selection import train_test_split
from sklearn.linear_model import LogisticRegression
from sklearn.metrics import confusion_matrix

# Load the transaction dataset
data = pd.read_csv('transactions.csv')

# Selecting features and target
X = data[['transaction_amount', 'time_of_day', 'location']]
y = data['fraudulent']

# Split the dataset
X_train, X_test, y_train, y_test = train_test_split(X, y, test_size=0.2)

# Train a logistic regression model
model = LogisticRegression()
model.fit(X_train, y_train)

# Make predictions
y_pred = model.predict(X_test)

# Confusion matrix to check performance
conf_matrix = confusion_matrix(y_test, y_pred)
print(conf_matrix)
\end{lstlisting}

\subsection{Algorithmic Trading}
Big Data also plays a significant role in algorithmic trading, where trading decisions are made by algorithms that can analyze vast amounts of data at incredible speeds. These algorithms rely on historical and real-time market data to make trades with minimal human intervention \cite{nuti2011algorithmic}.

\section{Big Data in Marketing and Consumer Analytics}
Big Data has revolutionized the way businesses understand and interact with their customers. By analyzing customer behavior and preferences, companies can develop more effective marketing strategies and improve customer satisfaction \cite{provost2013data}.

\subsection{Customer Segmentation}
Companies use Big Data to perform customer segmentation, dividing their customer base into groups based on demographics, purchasing behavior, and preferences. This allows businesses to tailor their marketing efforts to different segments for maximum impact \cite{kim2006customer}.

For instance, Python can be used to group customers based on their past purchases:

\begin{lstlisting}[style=python]
import pandas as pd
from sklearn.cluster import KMeans

# Load customer purchase data
data = pd.read_csv('customer_data.csv')

# Select features for clustering
X = data[['age', 'annual_income', 'spending_score']]

# Applying KMeans clustering
kmeans = KMeans(n_clusters=3)
data['cluster'] = kmeans.fit_predict(X)

# View the first few rows of the dataset with cluster labels
print(data.head())
\end{lstlisting}

\subsection{Recommendation Systems}
Recommendation systems, like those used by e-commerce sites, use Big Data to recommend products to users based on their browsing history, past purchases, and preferences. These systems use collaborative filtering or content-based filtering to predict what the user might want to purchase next \cite{jannach2010recommender}.

\section{Big Data for Government and Policy Making}
Governments worldwide are leveraging Big Data analytics to improve decision-making and service delivery. Big Data can help in areas like traffic management, public health, and resource allocation \cite{chen2014review}.

\subsection{Traffic Management}
By analyzing traffic data from sensors and cameras, governments can manage traffic more efficiently and reduce congestion. This helps cities design better infrastructure and plan road networks that cater to growing populations \cite{rizwan2016real}.

\subsection{Public Health Policy}
In public health, Big Data analytics can be used to monitor and predict the spread of diseases, identify at-risk populations, and allocate resources more effectively. During the COVID-19 pandemic, Big Data played a critical role in tracking infection rates and determining the impact of social distancing measures \cite{dash2019big}.

\section{Future Trends in Big Data Analytics}
The future of Big Data analytics is promising, with innovations such as machine learning, artificial intelligence, and the Internet of Things (IoT) expected to drive further advancements \cite{bahga2014internet}.

\subsection{AI and Machine Learning Integration}
The integration of AI and machine learning with Big Data analytics will make it easier to process and analyze vast datasets in real-time. These technologies will enable predictive analytics, anomaly detection, and automation of data processing tasks \cite{peng2024deeplearningmachinelearning}.

\subsection{Edge Computing}
Edge computing is another emerging trend in Big Data. Instead of sending all data to the cloud for processing, edge computing processes data closer to where it is generated, reducing latency and bandwidth requirements. This is especially useful for IoT devices that generate massive amounts of data \cite{qiu2020edge}.

\subsection{Ethics and Data Privacy}
As the use of Big Data grows, so do concerns about privacy and data security. It will become increasingly important for companies and governments to adopt ethical data practices and comply with regulations such as the GDPR and CCPA \cite{van2019does}.


\bibliographystyle{ieeetr}
\bibliography{sample}

\end{document}